\documentclass[twoside,11pt]{article}

\usepackage{blindtext}
\usepackage{jmlr2e}

\usepackage{lastpage}
\jmlrheading{25}{2024}{1-\pageref{LastPage}}{10/23; Revised
11/24}{12/24}{23-1368}{Huayi Tang and Yong Liu}
\usepackage{amsmath}
\usepackage{mathrsfs}
\usepackage{dsfont}
\usepackage{bm,bbm}
\usepackage{booktabs}
\usepackage{subfigure}
\usepackage{siunitx}
\def\dif{\mathop{}\hphantom{\mskip-\thinmuskip}\mathrm{d}}
\let\daccent\d
\let\d\relax
\newcommand\d{\ifmmode\dif\else\expandafter\daccent\fi}

\ShortHeadings{Information-Theoretic Generalization Bounds for Transductive Learning}{Tang and Liu}
\firstpageno{1}

\begin{document}

\title{Information-Theoretic Generalization Bounds for Transductive Learning and its Applications}

\author{\name Huayi Tang \email huayitang@ruc.edu.cn \\
       \addr Gaoling School of Artificial Intelligence\\
       Renmin University of China\\
       Beijing, 100872, China
       \AND
       \name Yong Liu\thanks{Yong Liu is the corresponding author.} \email liuyonggsai@ruc.edu.cn \\
       \addr Gaoling School of Artificial Intelligence\\
       Renmin University of China\\
       Beijing, 100872, China}

\editor{Pierre Alquier}

\maketitle

\begin{abstract}%
In this paper, we establish generalization bounds for transductive learning algorithms in the context of information theory and PAC-Bayes, covering both the random sampling and the random splitting setting. First, we show that the transductive generalization gap can be controlled by the mutual information between training label selection and the hypothesis. Next, we propose the concept of transductive supersample and use it to derive transductive information-theoretic bounds involving conditional mutual information and different information measures. We further establish transductive PAC-Bayesian bounds with weaker assumptions on the type of loss function and the number of training and test data points. Lastly, we use the theoretical results to derive upper bounds for adaptive optimization algorithms under the transductive learning setting. We also apply them to semi-supervised learning and transductive graph learning scenarios, meanwhile validating the derived bounds by experiments on synthetic and real-world datasets.
\end{abstract}


\section{Introduction}

Supervised learning is one of the most common paradigms in the field of machine learning \citep{Shalev2014,Mehryar2018}, where a limited number of data points coming from an unknown distribution are provided and each consisting of some features and a unique associated label. Our task is to build a model with these data points and use it to predict the labels of new coming data points based solely on their features. To this end, we first need to specify the learning algorithm and the model class. Next, we feed the collected data to the learning algorithm, which selects a model from the model class and returns it to us. Before applying this model to real-world scenarios, we need to evaluate its capability from diverse perspectives. 
Generalization ability, the prediction performance of the model on unseen data, is one of the most important capabilities that we focus on. Over the past decades, researchers have been developing theories to analyze and explain the generalization ability of machine learning algorithms. Early findings in this field suggest that generalization can be connected to the complexity of hypothesis space \citep{Koltchinskii2000,Koltchinskii2001,Bartlett2002,Bartlett2005} and the stability of learning algorithms \citep{Rogers1978,Bousquet2002,Kutin2002,Shwartz2010}. Most recently, information theory is shown to be a promising theoretical framework to analyze the generalization ability of learning algorithms \citep{Zhang2006,Russo2016,Russo2020,Xu2017,Negrea2019,Haghifam2020,Steinke2020,Haghifam2021,Sefidgaran2022,Wang2023}. Theoretical results derived under this framework convey a key insight. Specifically, a hypothesis that reveals less information about the training data tends to have better generalization ability. Moreover, \cite{Banerjee2021,Grunwald2021} have revealed that studying generalization from the perspective of information theory is closely related to PAC-Bayes, an earlier research route that uses the divergence between two probability measures on the hypothesis space to depict the generalization ability of learning algorithms \citep{Shawe1997,McAllester1998,McAllester1999,Catoni2007}. Notably, results derived from both information theory and PAC-Bayes are data-dependent and algorithm-dependent, thereby reflecting the impact of training data and learning algorithms on generalization.

So far, the theoretical analysis of the generalization ability of supervised learning algorithms is well-developed and productive. However, the supervised learning paradigm is not sufficient to cover all real-world application scenarios. First, real-world data may not be identically distributed. For example, a global server will exchange data with diverse clients in computer networks or distributed computing systems. Data received from different clients may not follow the same distribution. Second, real-world data may lack labels due to the expensive annotation cost or privacy protection requirements. Third, real-world data itself could be of low quality. Due to environmental interference or equipment failures, some data points could have noisy labels or incomplete features. These issues give rise to new learning paradigms and learning algorithms, prompting researchers to further develop new generalization theories for them. In this paper, we focus on the transductive learning paradigm \citep[Chapter~8]{Vapnik1998}, where both labeled and unlabeled data points are provided. Our task is to build a model based on them and use it to make predictions for these unlabeled ones. Notably, in the transductive learning setting, the features of the unlabeled data points intended for prediction are accessible to the model. Further, two settings for transductive learning are proposed in Chapter~8 of \cite{Vapnik1998} and they are respectively referred to as Setting~1 and Setting~2. In Setting~1, we sample partial data points from the collection of full data points without replacement and reveal their labels to the model. The learning goal is to minimize the risk of the model on the rest data points. Since we only concern the predictions of those unlabeled data points, \cite{Vapnik1982,Vapnik1998} terms this setting \emph{the problem of estimating the values of a function at given points}, and we term it \emph{the random splitting setting} in this paper. In Setting~2, however, a sequence of data points are drawn from an unknown distribution and we could only observe the labels of the data points at the front of this sequence. Notice that the data points in this sequence could come from different distributions, or there exists dependency between theirs. The learning goal is to choose a model so as to minimize the expected risk on the remaining data points in this sequence whose labels are not visible. Particularly, if the data points in this sequence are independent and identically distributed, the expected risk on the rest data points is exactly the expected risk in supervised learning, which depicts how well the hypothesis generalizes on unseen data points. Consequently, \cite{Vapnik1982,Vapnik1998} terms this setting \emph{the problem of estimating a function}, and we term it \emph{the random sampling setting} in this paper. In contrast, within the supervised learning paradigm, the features of test data points are not accessible to the model during training. As a result, this scenario, where models must make predictions on previously unseen data points, is referred to by \cite{Vapnik1982,Vapnik1998} as the inductive learning setting or inductive inference.

To understand these two kinds of transductive learning settings more intuitively, let us consider a scenario where we have a set of images and each of them is annotated by experts. We then randomly select some images without replacement from this set and reveal their labels together with all images to the model, and require it to make predictions for those images whose labels are not revealed. This process exemplifies the random splitting setting. Now suppose that we acquire new unlabeled images after a few days. By feeding both the previously collected labeled images and the new unlabeled images into the model, we ask it to predict the labels for the newly acquired images. This process exemplifies the random sampling setting. Compared with the random splitting setting, the random sampling setting is closer to real-world scenarios. However, we emphasize that the random splitting setting has its own unique value. Specifically, all the randomness in the random splitting setting is due to the partition of the full sample into training and test data. Consequently, we do not need to know the underlying distribution of the data or make assumptions about it, even in the presence of independence among the data points. For example, a fundamental task in graph learning is node classification, which involves a single graph composed of nodes and edges. Each node has associated features and a label, and our goal is to construct a model that can predict the labels of nodes in this graph. In this task, we assume that the graph structure is static. That is, one that does not add new nodes over time. Then, we can apply the random splitting setting by randomly selecting some nodes without replacement from the entire set of nodes and providing their labels along with all node features to the model. Indeed, this approach is widely used to train graph neural networks (GNNs) for node classification \citep{Justin2017,Kipf2017semisupervised,Petar2018}.

Existing results of the generalization upper bounds for transductive learning algorithms include complexity-based bounds derived from VC-dimension \citep{Cortes2006}, transductive Rademacher complexity \citep{Yaniv2007,Yang2023Sharp}, permutational Rademacher complexity \citep{Tolstikhin2015} and transductive local Rademacher complexity \citep{Yang2023Sharp}, stability-based bounds for transductive classification \citep{Yaniv2006} and transductive regression \citep{Cortes2008}, and PAC-Bayesian bounds \citep[Chapter~3]{Audibert2007,Derbeko2004,Begin2014,Catoni2007}. However, these findings still face certain limitations. Under the random splitting setting, complexity-based bounds \citep{Cortes2006,Yaniv2007,Tolstikhin2015} and stability-based bounds \citep{Cortes2008} are independent of the learning algorithm and training data selection, respectively. Consequently, these results fail to simultaneously reflect the influence of both the learning algorithm and the selection of training data on generalization performance. Moreover, applying these results to deep transductive models like GNNs presents additional challenges. Specifically, bounds derived from VC-dimension could become trivial \citep{Esser2021}, while stability based bounds \citep{Cong2021} involve Lipschitz or smoothness constants of the loss function that are hard to estimate or even be bounded \citep{Neu2021}. Additionally, for PAC-Bayesian bounds, results established by \cite{Derbeko2004} are of slow order, and results of \cite{Begin2014} require strong assumptions about the loss function and the number of training and test data points. Furthermore, under the random sampling setting, existing results require that the number of test data points is a multiple of the number of training data points \citep[Chapter~3]{Catoni2007} or equals the number of training data points \citep{Catoni2003,Audibert2007}.

In this paper, we delve into the generalization ability analysis of transductive learning algorithms within the framework of information theory and PAC-Bayesian theory. First, under the random splitting setting, we establish average and single-draw bounds for the transductive generalization gap using tools from information theory. These results contain the mutual information between the hypothesis and the training label selection, which indicates that transductive learning algorithms generalize better when their output hypothesis is less dependent on the specific choice of training label. We then introduce the concept of transductive supersample and extend the framework of \cite{Steinke2020} to the transductive learning setting, which allows us to transport results based on various information measures derived under the inductive learning setting to the transductive learning setting. Second, we establish PAC-Bayesian bounds for the transductive generalization gap under both the random splitting setting and the random sampling setting. These results allow the number of training and test data points to be arbitrary integers and only require the loss function to be bounded. Moreover, we demonstrate that a previous result, which states that a flatter loss landscape implies better generalization performance under the inductive learning setting, remains valid under the transductive learning setting. This result provides theoretical support to recent empirical observation of \cite{Chen2023}. Third, we apply the above results to establish generalization upper bounds for adaptive optimization algorithms under the transductive learning setting. We also demonstrate the applications of our theoretical results on semi-supervised learning and transductive graph learning scenarios and validate our results with numerical experiments on both synthetic and real-world datasets. The key contributions of this work can be summarized as follows.
\begin{itemize}
    \item We establish information-theoretic bounds for the transductive generalization gap and the largest eigenvalue of its Hessian under the random splitting setting. 
    \item We propose the concept of transductive supersamples for the first time and derive new information-theoretic and PAC-Bayesian bounds under the random splitting setting.
    \item We establish PAC-Bayesian bounds with weaker assumptions on the number of data points and the type of loss function under the random sampling setting.
\end{itemize}

In the remainder of this paper, we begin with an overview of the literature related to our work in Section~\ref{sec2}. We then introduce the mathematical notations used throughout this paper along with the random splitting setting and the sampling setting of transductive learning in Section~\ref{preli}. The main theoretical results are presented in Section~\ref{main_sec}, followed by their applications in Section~\ref{appli}. The experimental setting and results are detailed in Section~\ref{experi}. Finally, we conclude the paper in Section~\ref{conclu}. Complete proofs of all theoretical results in the main text are provided in the appendix. 

\section{Related Work}\label{sec2}

\subsection{Information-theoretic Generalization Theory}
\cite{Russo2016,Russo2020} and \cite{Xu2017} link the expected generalization error with the mutual information between training examples and algorithm output. Subsequent studies can be categorized into four main groups: (\romannumeral1) deriving tighter upper bounds by introducing novel information measures \citep{Harutyunyan2021,Hellstrom2022,Wang2023}, problem settings \citep{Steinke2020,Mohamad2022,Haghifam2022} or proof techniques \citep{Asadi2018,Bu2020,Hafez2020,Borja2021,Zhou2022,Clerico2022}; (\romannumeral2) establishing bounds characterized by various divergences \citep{Lopez2018,Wang2019wasser,Esposito2021,Aminian2021Jensen,Aminian2021Moments}; (\romannumeral3) applying existing results to derive upper bounds for optimization algorithms like stochastic gradient descent (SGD) \citep{Neu2021,Wang2022} or stochastic gradient langevin dynamics (SGLD) \citep{Pensia2018,Negrea2019,Wang2021analyzing}; and (\romannumeral4) extending above theoretical results to different scenarios such as meta-learning \citep{Jose2021,Rezazadeh2021,Chen2021generalization,Jose2022}, transfer learning \citep{Wu2020,Jose2021transfer,Masiha2021,Bu2022}, semi-supervised learning \citep{Aminian2022,He2022}, self-supervised learning \citep{Yige2022}, and domain adaption \citep{Wang2023Domain}. However,
these studies fail to account for the transductive learning setting. Another related area is the information bottleneck theory \citep{Tishby2000} and its applications in explaining representation \citep{Tishby2015,Tishby2017} and generalization \citep{Hassan2020,Wang2022pacbayes,Kawaguchi2023} of deep neural networks, which is independent of our works. For a thorough review of information-theoretic generalization theory, we refer readers to the recent monograph of \cite{Hellstrom2023}.

\subsection{PAC-Bayesian Generalization Theory} 
The classical results in PAC-Bayesian generalization theory include McAllester's bound \citep{McAllester1999}, Maurer-Langford-Seeger's bound \citep{Langford2001book,Seeger2002,Maurer2004} and Catoni’s bound \citep[Chapter~1]{Catoni2007}. Building on these foundational works, a substantial body of research has emerged that applies or extends these results to analyze various models or algorithms, including computing non-vacuous bounds for deep neural networks \citep{Dziugaite2017,Zhou2019nonvacuous,Ortiz2021,Dziugaite2021,Lotfi2022pacbayes} and establishing upper bounds for optimization algorithms \citep{London2017,Rivasplata2018,Arora2018,Mou2018,Yang2019,Li2020On,Luo2022generalization} or specific neural network architectures \citep{Neyshabur2018,Liao2021,Mbacke2023}. For more details, see monographs of \cite{Benjamin2019,Pierre2021} along with their references. However, the above studies do not yet encompass the random splitting setting of transductive learning.

\subsection{Generalization Theory of Transductive Learning}
The concept of transductive learning along with the earliest generalization bounds are introduced by \cite{Vapnik1982}. Under the random splitting setting, generalization bounds are established by transductive algorithm stability \citep{Yaniv2006} and transductive Rademacher complexity \citep{Yaniv2007}, respectively. Later, \cite{Tolstikhin2015} introduce permutational Rademacher complexity and demonstrate that it is more suitable for the transductive learning setting compared to transductive Rademacher complexity. Their difference is that the expectation is taken over the transductive Rademacher variables in transductive Rademacher Complexity, while in permutational Rademacher complexity, the expectation is taken over the selection of training labels. By incorporating the variance of functions, \cite{Tolstikhin2014} develop new concentration inequalities and derive tighter bounds. A novel complexity metric, termed transductive local Rademacher complexity, is introduced by \cite{Yang2023Sharp} recently. The expectation is taken over both standard Rademacher variables and the selection of training labels in transductive local Rademacher complexity, which allows us to obtain tighter bounds in some scenarios. In contrast, we establish upper bounds based on information theory. Transductive PAC-Bayesian bounds under the random splitting setting are initially developed by \cite{Derbeko2004}, which is subsequently improved by \cite{Begin2014}. We further improve their results and apply them to reveal the impact of loss landscape flatness on generalization. Under the random sampling setting, Catoni firstly establishes the generalization bound in the context of PAC-Bayesian theory \citep[Chapter~3]{Catoni2007}, which is further improved by \cite{Audibert2007} via the generic chaining technique. Different from these studies, we establish transductive PAC-Bayesian bounds through distinct techniques, and the results have weaker assumptions on the number of data points and loss function type. Besides, the above theoretical results have been applied to areas such as transductive graph learning \citep{Shivanna2014,Shivanna2015,De2018,Oono2020,Esser2021,Cong2021,Tang2023}, semi-supervised learning \citep{Yury2018,Chen2018,Xu2023}, matrix completion \citep{Febrer2020,Shamir2014}, distributed optimization \citep{Shamir2016} and collaborative filtering \citep{Xu2021Rethink,Deng2022graph}. We select semi-supervised learning and transductive graph learning as examples to demonstrate our theoretical results. Extensions to other domains are left for future work.

\section{Preliminaries}\label{preli}

\subsection{Notations}
We stipulate that random variables and their realizations are denoted by uppercase and lowercase letters, respectively. For a given random variable $X$, we denote its distribution measure by $P_X$. The conditional distribution measure of $X$ given $Y$ is denoted by $P_{X|Y}$. We use $\mathrm{D_{KL}}(P||Q)$ to denote the Kullback–Leibler (KL) divergence between two probability measures $P$ and $Q$ from the same probability space. Notice that $\mathrm{D_{KL}}(P||Q)$ is well defined if $P$ is absolutely continuous with respect to (w.r.t.) $Q$, which we denote by $P \ll Q$. The mutual information between $X$ and $Y$ is represented as $I(X;Y) = \mathrm{D_{KL}}(P_{X,Y}||P_X P_Y)$. Furthermore, we use $I^z(X;Y) = \mathrm{D_{KL}}(P_{X,Y|Z=z}||P_{X|Z=z} P_{Y|Z=z})$ to represent the disintegrated mutual information, whose expectation taken over $Z\sim P_Z$ is the conditional mutual information $I(X;Y|Z) = \mathbb{E}_{Z} [I^Z(X;Y)]$. Besides, we use $\{ \cdot\}$ and $(\cdot)$ to denote sets and sequences, respective. Particularly, $[n]$ represents the set $\{1,\ldots,n\}$. $\mathbb{R}$ and $\mathbb{N}$ are the set of real numbers and natural numbers, respectively. $\mathbb{R}_{\geq 0}$ and $\mathbb{N}_{+}$ are the set of non-negative real numbers and positive integers, respectively. The Hadamard Product and Kronecker Product are denoted by $\odot$ and $\otimes$, respectively. The spectral norm and $L_2$ norm are denoted by $\Vert \cdot \Vert$ and $\Vert \cdot \Vert_2$, respectively. We use $\mathrm{Perm}((\cdot))$ to denote the set containing the full permutations of a given sequence $(\cdot)$. For example, we have $\mathrm{Perm}((1,2,3)) = \{ (1,2,3), (1,3,2), (2,1,3), (2,3,1), (3,1,2), (3,2,1) \}$. Besides, ${\rm Unif}(\{\cdot\})$ means the uniform distribution over a fixed support set $\{\cdot\}$. For any fixed integers $m,u \in \mathbb{N}_+$, we define $C_{m,u} \triangleq \frac{2(m+u)\max(m,u)}{(m+u-1/2)(2\max(m,u)-1)}$ as a constant with regard to $m$ and $u$. For $0 < p < 1$, we define $\Phi_a(p) \triangleq -a^{-1}\log \left(1-[1-e^{-a}]p \right)$ as a function of $p$, where $a\in \mathbb{R}$ is the parameter.

\subsection{Random Splitting Setting for Transductive Learning}\label{trc_pre}
Let $\{ s_i \}_{i=1}^n$ be a given set with finite cardinality. Each element $s \triangleq (x,y) \in \mathcal{X} \times \mathcal{Y}$ in this set is composed of feature $x \in \mathcal{X}$ and label $y \in \mathcal{Y}$, where $\mathcal{X}$ and $\mathcal{Y}$ are the feature domain and finite label domain, respectively. Denote by $m \in \{1,\ldots,n-1\}$ and $u \triangleq n-m$ the number of training data points with labels and test data points, respectively. The training data points are obtained by randomly sampled $m$ elements without replacement from $\{ s_i \}_{i=1}^n$. To this end, we continuously sample from $\{1,\ldots,n\}$ without replacement and sequentially place the obtained elements into a sequence $Z \triangleq (Z_1,\ldots,Z_n)$. In other words, $Z_i$ is the element obtained from the $i$-th sampling process. Once the sampling process is finished and $Z$ is obtained, the training labels are given by $\{y_{Z_i}\}_{i=1}^{m}$. That is, only partial selected labels $\{y_{Z_i}\}_{i=1}^{m}$ together with all features $\{x_i\}_{i=1}^{m}$ are revealed to the transductive model. Notice that the randomness comes from the selection of training labels and it is essentially contained in the sampling sequence $Z$. Now let us see a simple example to understand the above process. For simplicity, we let $n=3$ and $m=1$. Assuming that we have completed a sampling of $\{1,2,3\}$. The obtained elements from the first, second, and third times were $2$, $3$, and $1$, respectively. Then the sampling sequence is given by $z=(2,3,1)$, which is a realization of $Z$. In this example, the training set provided to the transductive model is $\{(x_{2},y_{2})\} \bigcup \{x_1,x_3\}$, and it is required to predict the labels of $s_1$ and $s_3$.

Formally, a transductive model is defined as a mapping $f: \mathcal{X} \to \mathcal{Y}$. In this paper, we stipulate that each model is parameterized. That is, each model (hypothesis) is associated with a unique parameter $w \in \mathcal{W}$, where $\mathcal{W}$ is the parameter space. We denote by $\mathcal{P}(\mathcal{W})$ the set of distribution over $\mathcal{W}$. The model class (hypothesis space) is given by $\mathcal{F}_{\mathcal{W}} \triangleq \{f_w: \mathcal{X} \to \mathcal{Y} | w \in \mathcal{W}\}$.  A transductive learning algorithm will take all features together with training labels determined by $Z$ as input and return a hypothesis $f_W \in \mathcal{F}_{\mathcal{W}}$, whose randomness is depicted by a Markov kernel $P_{W|Z}$. Let $\ell: \mathcal{W} \times (\mathcal{X} \times \mathcal{Y}) \to \mathbb{R}_{\geq 0}$ be the loss function, the transductive training error and test error of a hypothesis $f_W$ are defined as $R_{\rm train}(W,Z) \triangleq \frac{1}{m} \sum_{i=1}^m \ell(W,s_{Z_i})$ and $R_{\rm test}(W,Z) = \frac{1}{u} \sum_{i=m+1}^{m+u} \ell(W,s_{Z_i})$, respectively. The transductive generalization gap is then defined as $\mathcal{E}(W,Z) \triangleq R_{\rm test}(W,Z) - R_{\rm train}(W,Z)$. Furthermore, we use $\mathbb{E}_{W,Z} [\mathcal{E}(W,Z)]$ to denote the expectation of $\mathcal{E}(W,Z)$ taken over $P_{W,Z}=P_Z \otimes P_{W|Z}$, which represents the transductive generalization gap of the hypothesis $f_W$ over the randomness of training labels selection and the learning algorithm. In certain cases, the loss function $\ell(w,s)$ can also be represented as $r(f_w(x),y)$, where $f_w(x)$ is the prediction of the model on $x$, and $r: \widehat{\mathcal{Y}} \times \mathcal{Y} \to \mathbb{R}_{\geq 0}$ is the criterion. For example, we have $r(\hat{y},y) = \mathds{1}\{\hat{y} \ne y\}$ when the criterion is zero-one loss, where $\mathds{1}\{\cdot\}$ is the indicator function. 

\subsection{Random Sampling Setting for Transductive Learning}\label{trc2}
Denote by $S=(S_1,\ldots,S_n) \sim P_{S_1,\ldots,S_n}$ a sequence of data points, where $S_i \triangleq (X_i,Y_i),i\in [n]$ is composed of feature $X_i \in \mathcal{X}$ and label $Y_i \in \mathcal{Y}$. Notice that $S$ is a sequence of random variables rather than constants. After the data sequence $S$ is obtained, partial labels $(Y_1,\ldots,Y_m)$ together with all features $(X_1,\ldots,X_n)$ are revealed to a transductive model, whose task is predicting the labels of $(X_{m+1},\ldots,X_n)$. Recall that $m \in \{1,\ldots,n-1\}$ and $u \triangleq n-m$ are the number of training data points with labels and test data points, respectively. Similar to the notations used in the random splitting, the hypothesis space is denoted by $\mathcal{F}_{\mathcal{W}} \triangleq \{f_w:\mathcal{X} \to \mathcal{Y}| w\in \mathcal{W}\}$. For a hypothesis $f_W \in \mathcal{F}_{\mathcal{W}}$, its transductive training and test error are defined as $R_{\text{train}}(W,S) \triangleq \frac{1}{m} \sum_{i=1}^m \ell(W,S_i)$ and $R_{\text{test}}(W,S) = \frac{1}{u} \sum_{i=m+1}^{m+u} \ell(W,S_i)$ respectively. Particularly, under the case that $u=km,k\in \mathbb{N}_+$, the transductive training and test errors are reformulated as $R_{\text{train}}(W,S) \triangleq \frac{1}{m}\sum_{i=1}^m \ell \left(W,S_{i} \right)$ and $R_{\text{test}}(W,S) \triangleq \frac{1}{km}\sum_{i=m+1}^{(k+1)m} \ell \left(W,S_i\right)$, respectively. The randomness of learning algorithms is depicted by the Markov kernel $P_{W|S}$. 

\section{Main Results}\label{main_sec}

\subsection{Mutual Information Bounds for Transductive Learning}\label{mi_bounds}
In this section, we establish generalization bounds for the transductive generalization gap $\mathcal{E}(W,Z)$ in the context of information theory. To this end, there exist two technical challenges. The first challenge is that the parameter $W$ returned by a transductive learning algorithm is dependent on the sampling sequence $Z$. The reason is that the algorithm is run after the training labels are selected, which is determined by the sampling sequence $Z$. Under the framework of information theory \citep{Xu2017,Hellstrom2023}, the approach for this issue is applying the change of measure technique and shifting the distribution from $P_{W,Z}$ to $P_{W \otimes Z}$, by which we can decouple the dependence between $W$ and $Z$. The second challenge is that the training and test data points are dependent since the labels of training data points are obtained by sampling without replacement. In other words, once the training labels are specified (or the first $m$ entries of $Z$ are observed), both $R_{\rm train}(W,Z)$ and $R_{\rm test}(W,Z)$ are uniquely determined. In this work, we adopt the martingale technique to address this issue, which is a widely used tool in statistics. The process is constructing a Doob's martingale and then showing that the corresponding martingale differences are bounded. Integrating the above steps, we obtain the following results.
\begin{theorem}\label{thm1}
Suppose that $\ell(w,s) \in [0,B]$ holds for all $w \in \mathcal{W}, s \in \{s_i\}_{i=1}^n$, where $B > 0$ is a constant. Also, suppose that $P_{W,Z} \ll P_W P_Z$. Then we have
\begin{align}
    & \left\vert \mathbb{E}_{W,Z} \left[ R_{\rm test}(W, Z) - R_{\rm train}(W, Z) \right] \right\vert \leq \sqrt{\frac{B^2C_{m,u}I(W;Z)(m+u)}{2mu}} \label{first}, \\
    & \mathbb{E}_{W,Z} \left[ (R_{\rm test}(W, Z) - R_{\rm train}(W, Z))^2 \right] \leq \frac{B^2C_{m,u}(I(W;Z)+\log 3)(m+u)}{mu} \label{sqexp}.
\end{align}
\end{theorem}
Notice that $C_{m,u} \approx 1$ holds when the values of $m,u$ are sufficiently large. Theorem~\ref{thm1} shows that the expectation of transductive generalization gap is upper bounded by the mutual information between the sampling sequence $Z$ and the hypothesis $W$ returned by the learning algorithm. Since $Z$ is obtained by sampling without replacement from $[n]$ and its first $m$ entries are indices of training data points, it essentially depicts the randomness of training label selection. Therefore, Theorem~\ref{thm1} demonstrates that for the hypothesis returned by a transductive learning algorithm, the less dependence it has on the selection of training labels, the smaller the generalization gap it will have. Intuitively, if the model only ``memorizes'' the obtained training labels (or strongly relies on training labels to predict those unlabeled data points), it fails to capture the underlying relation between data points and their labels, making it difficult to perform well on data points whose labels are not revealed. As a comparison, the result under the inductive learning setting \citep[Theorem~1]{Xu2017} says that the generalization error is upper bounded by the mutual information between the training set $S=\{S_i\}_{i=1}^n$ and the hypothesis $W$, where data points $S_i = (X_i,Y_i) \sim P_{X,Y}, i\in [n]$ are random variables. Since all features are available for the model, the randomness only comes from training label selection in the transductive learning setting, and the training set $S$ is accordingly replaced by the sampling sequence $Z$. Moreover, the assumption of our result is slightly stronger than that used in the inductive learning setting. To be specific, Theorem~1 of \cite{Xu2017} only requires the loss function is sub-Gaussian (that is, supposing that there exists a constant $\sigma > 0$ such that $\log \mathbb{E}_{P_{X,Y}}\left[e^{\lambda \ell(w,(X,Y))}\right] \leq \frac{\lambda^2 \sigma^2}{2}$ for all $w \in \mathcal{W}$), while Theorem~\ref{thm1} requires the loss function to be bounded (that is, supposing that $\ell(w,s)$ is bounded for all $w \in \mathcal{W}$ and $s \in \{s_i\}_{i=1}^n$). The reason why we make a stronger reason is to ensure that martingale differences constructed in the proof have bounded differences. However, we point out that the bounded assumption of the loss function in Theorem~\ref{thm1} can be further relaxed, inspired by the work of \cite{Steinke2020}. Concretely, by replacing the loss bounded assumption in Theorem~\ref{thm1} with the following one: suppose that there exists a function $\widetilde{\Delta}: (\mathcal{X}\times \mathcal{Y})^2 \to \mathbb{R}_{\geq 0}$ such that $\vert \ell(w,s_1) - \ell(w,s_2) \vert \leq \widetilde{\Delta}(s_1,s_2)$ holds for all $s_1,s_2 \in \{s_i\}_{i=1}^n$ and $w\in \mathcal{W}$, we can obtain results analogous to Eqs.~(\ref{first},\ref{sqexp}) where $B$ is accordingly replaced by $\mathop{\rm sup}_{s_1, s_2 \in \{s_i\}_{i=1}^n} \widetilde{\Delta}(s_1,s_2)$. If further assuming that $\mathop{\rm sup}_{s_1, s_2 \in \{s_i\}_{i=1}^n} \widetilde{\Delta}(s_1,s_2) \leq B$ holds for a constant $B > 0$, we exactly recover the results presented in Theorem~\ref{thm1}.

Eq.~(\ref{first}) in Theorem~\ref{thm1} is an upper bound for the average transductive generalization error, where the expectation is taken over the distribution $P_{W,Z}$. In real-world applications, particularly deep learning scenarios, only a few train-test splits $Z$ are sampled to evaluate the performance of a transductive learning algorithm. And, restricted by computational costs, we will only repeat the algorithm several times. Therefore, a single-draw bound that holds with probability at least $1-\delta$ under the distribution $P_{W,Z}$ can better depict the generalization ability of learning algorithms under this circumstance, which is derived by adopting the monitor technique proposed by \cite{Bassily2016}.
\begin{theorem}\label{thm2}
Under the assumptions of Theorem~\ref{thm1}, for any $0 < \delta < 1$, with probability at least $1-\delta$ over the randomness of $Z$ and $W$ we have
\begin{equation}\label{highpro}
    \left\vert R_{\rm test}(W, Z) - R_{\rm train}(W, Z) \right\vert \leq \sqrt{\frac{2B^2C_{m,u}(m+u)}{mu} \left(\log \left(\frac{2}{\delta}\right) +  \frac{I(W;Z)}{\delta} \right)}.
\end{equation}
Moreover, we have
\begin{equation}\label{absexp}
    \mathbb{E}_{W,Z} [\left\vert R_{\rm test}(W, Z) - R_{\rm train}(W, Z) \right\vert] \leq \sqrt{\frac{B^2C_{m,u}(m+u)(I(W;Z)+\log 2)}{2mu}}.
\end{equation}
\end{theorem}

In addition to the single-draw bound, we obtain another result from the monitor technique, which is an expectation bound on the absolute value of the transductive generalization error and serves as a supplement of Theorem~\ref{thm1}. Now we are in a place to compare Theorem~\ref{thm2} with a previous transductive generalization bound based on the complexity of hypothesis space \citep[Theorem~1]{Yaniv2007}, which is restated as follows.
\begin{theorem}[\citealp{Yaniv2007}, Theorem~1]\label{thm_yaniv}
    Let $B > 0$ be a constant. Suppose that $\ell(w,s) \in [0,B]$ holds for all $w \in \mathcal{W}, s \in \{s_i\}_{i=1}^n$. For any $0 < \delta < 1$, with probability $1-\delta$ over the randomness of $Z$ and $W$,
    \begin{small}
    \begin{equation*}
    R_{\rm test}(W, Z) \leq R_{\rm train}(W, Z) + \mathfrak{R}_{m+u}(\mathcal{W}) +  \frac{c_0 B(m+u)\sqrt{\min(m,u)}}{mu} + \sqrt{\frac{C_{m,u}(m+u)\log(1/\delta)}{2mu}},
    \end{equation*}
    \end{small}%
    where $c_0 \triangleq \sqrt{\frac{32\log(4e)}{3}}$ and $\mathfrak{R}_{m+u}(\mathcal{W})$ is the transductive Rademacher complexity.
\end{theorem}
The single-draw bound in Eq.~(\ref{highpro}) of Theorem~\ref{thm2} and Theorem~\ref{thm_yaniv} is of order $\sqrt{(m+u)/mu}$ and $\sqrt{\min(m,u)}(m+u)/mu$, respectively. Since $\sqrt{\min(m,u)} > \sqrt{mu/(m+u)}$, the bound in Eq.~(\ref{highpro}) is sharper than that in Theorem~\ref{thm_yaniv} regarding the dependency of $m$ and $u$, if
\begin{equation}
    I(W;Z) \lesssim \frac{mu \cdot \min(m,u)}{m+u} \cdot \log(1/\delta) \cdot \delta.
\end{equation}
Although the mutual information term $I(W;Z)$ could not be easily computed, we show in Section~\ref{opti_sec} that it has a unique advantage for analyzing iterative learning algorithms, such as SGD and its variants. Besides, a result similar to Eq.~(\ref{absexp}) can be derived by taking the square root on both sides of Eq.~(\ref{sqexp}) and applying Jensen's inequality. But, the constant factor of the derived result is slightly larger than that of Eq.~(\ref{absexp}). We remark that Eq.~(\ref{absexp}) is not a deteriorated version of Eq.~(\ref{first}), as it is obtained by the monitor technique rather than directly derived from Eq.~(\ref{absexp}). Besides, one may notice a limitation of Eq.~(\ref{highpro}) that its dependency of $\delta$ deteriorates from $\log(1/\delta)$ to $1/\delta$ compared with Theorem~\ref{thm_yaniv}. This issue can not be easily addressed without extra cost, that is, improving the dependency of $\delta$ may deteriorate the dependency of $m$ and $u$. To see this, we adopt the technique proposed by \cite{Fredrik2020} to derive another single draw bound analogous to Eq.~(\ref{highpro}), where the cost of performing a change of measure from $P_{W,Z}$ to $P_{W \otimes Z}$ is depicted by the information density $\log \frac{\dif P_{W,Z}}{\dif P_W P_Z}$ instead of mutual information $I(W;Z)$. The derived result is given in Proposition~\ref{pro_density}. As can be seen, the confidence parameter $\delta$ in Eq.~(\ref{infor_density}) is of order $\log(1/\delta)$, yet the dependency of $m,u$ is worse than that of Eq.~(\ref{highpro}).
\begin{proposition}\label{pro_density}
    Under the assumptions of Theorem~\ref{thm1} and suppose that $m \geq 2, u \geq 3$ or $m \geq 3, u \geq 2$, with probability at least $1-\delta$ over the randomness of $Z$ and $W$, we have
    \begin{equation}\label{infor_density}
        \left\vert R_{\rm test}(W, Z) - R_{\rm train}(W, Z) \right\vert \leq \sqrt{\frac{B^2C_{m,u}(m+u)}{2(mu-m-u)} \left(\log \left(\frac{1}{\delta}\sqrt{\frac{mu}{m+u}}\right) +  \log \frac{\dif P_{W,Z}}{\dif P_W P_Z} \right)}.
    \end{equation}
\end{proposition}

\subsection{Conditional Mutual Information Bounds for Transductive Learning}\label{cmi_sec}

The established bounds in Section~\ref{mi_bounds} contain either the mutual information or the information density. Both of them are unbounded and may result in the bounds being unable to provide meaningful learning guarantees. Also, both $W$ and $Z$ are high dimensional random variables in real-world scenarios, which makes it challenging to compute the numerical value of $I(W;Z)$ with finite samples. To address this issue, under the inductive learning setting, \cite{Steinke2020} propose the conditional mutual information (CMI) framework to disentangle the randomness of sampling data points and splitting them into training and test sets. Concretely, the randomness of sampling data points is depicted by the supersample $\widetilde{Z} \in (\mathcal{X}\times \mathcal{Y})^{m\times 2}$, which is an array containing $2m$ data points drawn independently from $P_{X,Y}$. The randomness of splitting training and test set is depicted by the selector sequence $U = (U_1,\ldots,U_m) \sim \mathrm{Unif}(\{0,1\})^m$, which is independent of $\widetilde{Z}$. In this way, we can use $I(W;U|\widetilde{Z})$ instead of $I(W;Z)$ to measure the information on training data carried by the hypothesis returned by learning algorithms. Now let us come back to the transductive learning setting. Notice that the training and test data points are \emph{independent} in the inductive setting yet \emph{dependent} in the transductive learning setting. Thus, the supersample setting of \cite{Steinke2020} is not applicable to the transductive learning setting. To bridge this gap, we propose the concept of transductive supersample under the assumption that the number of test data points is a multiple of the number of training data points, namely $u=km, k \in \mathbb{N}_+$. As a warm-up example, we first discuss the case that the number of training examples is equal to that of test examples, namely $k=1$.
\begin{definition}[Transductive Supersample]\label{def}
Let $m=u=\frac{n}{2}$. Performing $m$ rounds of sampling without replacement from the set $\{1,\ldots, n\}$ and each time two elements are chosen. For each $i\in [m]$, arrange the elements from the $i$-th sampling into a sequence $\widetilde{Z}_i \triangleq (\widetilde{Z}_{i,0},\widetilde{Z}_{i,1})$ in ascending order, that is, we have $\widetilde{Z}_{i,0} < \widetilde{Z}_{i,1}$. The transductive supersample is defined as a sequence $\widetilde{Z} \triangleq (\widetilde{Z}_1, \ldots, \widetilde{Z}_m)$.
\end{definition}
 
Notice that for each $i\in [m]$, the first and second elements of $\widetilde{Z}_i$ are $\widetilde{Z}_{i,0}$ and $\widetilde{Z}_{i,1}$, respectively. Now let us elaborate on Definition~\ref{def} through a simple example. For simplicity, we let $m=u=2$. Assuming that we have completed a sampling of $[4]$, and the elements obtained from the first and second rounds are $\{3,2\}$ and $\{1,4\}$. By arranging the elements $\{3,2\}$ into a sequence in ascending order, we have $\widetilde{z}_{1,0}=2$ and $\widetilde{z}_{1,1}=3$ and thus $\widetilde{z}_1=(2,3)$. Similarly, we have $\widetilde{z}_2=(1,4)$. Then the transductive supersample is denoted by $\widetilde{z} = (\widetilde{z}_1,\widetilde{z}_2) = ((2,3), (1,4))$. Clearly, each entry in $\widetilde{z}$ is a sequence of length $2$, and we remark that $((2,3), (1,4)) \ne ((1,4),(2,3))$. Recall that the sequence $Z$ defined in Section~\ref{trc_pre} is obtained by each time sampling one element from $[n]$ without replacement. By introducing the selector sequence $U$ to permute elements in the transductive supersample $\widetilde{Z}$, we show that the sampling sequence $Z$ can be recovered from $\widetilde{Z}$ and $U$. 
\begin{proposition}\label{prop1}
Let $U\triangleq(U_1,\ldots,U_m) \sim \mathrm{Unif}(\{0,1\})^m$ be the sequence of random variables that is independent of a transductive supersamples $\widetilde{Z}=(\widetilde{Z}_{1},\ldots,\widetilde{Z}_m)$. Define 
\begin{equation}\label{trc_2}
    \mathscr{Z}(\widetilde{Z},U) \triangleq (\widetilde{Z}_{1,U_1}, \ldots, \widetilde{Z}_{m,U_m}, \widetilde{Z}_{1,1-U_1}, \ldots, \widetilde{Z}_{m,1-U_m})
\end{equation}
as the sequence induced by $\widetilde{Z}$ and $U$. Define the transductive training and test error as
\begin{equation}
\begin{aligned}
    & R_{\rm train}(W,\widetilde{Z},U) \triangleq \frac{1}{m} \sum_{i=1}^m \ell \Big( W,s_{\mathscr{Z}_i(\widetilde{Z},U)} \Big) = \frac{1}{m} \sum_{i=1}^m \ell \Big( W,s_{\widetilde{Z}_{i,U_i}} \Big), \\
    & R_{\rm test}(W,\widetilde{Z},U) \triangleq \frac{1}{m} \sum_{i=m+1}^{2m} \ell \Big( W,s_{\mathscr{Z}_i(\widetilde{Z},U)} \Big) = \frac{1}{m} \sum_{i=1}^m \ell \Big( W,s_{\widetilde{Z}_{i,1-U_i}} \Big),
\end{aligned}
\end{equation}
where $\mathscr{Z}_i(\widetilde{Z},U), i\in [2m]$ is the $i$-th entry of the sequence $\mathscr{Z}(\widetilde{Z},U)$. We have 
\begin{equation}
\begin{aligned}
    & \mathbb{E}_{W,\widetilde{Z},U} [R_{\rm test}(W,\widetilde{Z},U) - R_{\rm train}(W,\widetilde{Z},U) ] \\
    = &\mathbb{E}_{W,Z} [R_{\rm test}(W,Z) - R_{\rm train}(W,Z) ] = \mathbb{E}_{W,Z} [ \mathcal{E}(W, Z)].
\end{aligned}
\end{equation}
\end{proposition}

Again, we use a simple example to explain our thought. As before, we let $m=u=2$.
Assume that the realization of $\widetilde{Z}$ and $U$ are $\widetilde{z} = ((2,3),(1,4))$ and $u=(0,1)$ respectively. In this case, we have $\widetilde{z}_1 = (2,3)$, $\widetilde{z}_2=(1,4)$, $u_1=0$ and $u_2=1$. Then the sequence $Z$ they induce is given by $\mathscr{Z}(\widetilde{z},u) = (\widetilde{z}_{1,u_1},\widetilde{z}_{2,u_2},\widetilde{z}_{1,1-u_1},\widetilde{z}_{2,1-u_2}) = (2,4,3,1)$. 
Proposition~\ref{prop1} shows that if $m=u$, we can replace $Z$ by $\widetilde{Z}$ and $U$. Concretely, instead of directly sampling $Z$, we can firstly sample $\widetilde{Z}$ and $U$ respectively and then permute the entries in $\widetilde{Z}$ according to Eq.~(\ref{trc_2}). In this way, we can disentangle the randomness of $Z$ into two parts, one contained in $\widetilde{Z}$ and the other contained in $U$. Notably, $U$ is dependent to $\widetilde{Z}$. This enables us to tackle the dependence introduced by sampling without replacement and establish CMI bounds for transductive learning algorithms. Moreover, if $\ell(\cdot)$ is the zero-one loss, we can further derive PAC-Bayesian bounds under the CMI framework by using Catoni's technique \citep[Chapter~3.1]{Catoni2007}. The aforementioned results are summarized in Theorem~\ref{thm3}.
\begin{theorem}\label{thm3}
Suppose that $\ell(w,s) \in [0,B]$ holds for all $w \in \mathcal{W}, s \in \{s_i\}_{i=1}^n$. We have
\begin{align}
    & \left\vert \mathbb{E}_{W,Z} \left[ R_{\rm test}(W, Z) - R_{\rm train}(W, Z) \right] \right\vert \leq \mathbb{E}_{\widetilde{Z}} \sqrt{\frac{2B^2}{m}I^{\widetilde{Z}}(W;U)}, \label{cmi} \\
    & \mathbb{E}_{W,Z} \left[ (R_{\rm test}(W,Z) - R_{\rm train}(W,Z))^2 \right] \leq \frac{4B^2}{m}(I(W;U|\widetilde{Z})+\log 3).
\end{align}
Moreover, suppose that $\ell(\cdot)$ is the zero-one loss. Let $P \in \mathcal{P}(\mathcal{W})$ be a prior distribution that is independent of $U$. For any $0 < \delta < 1$, $\lambda > 0$, and $Q \in \mathcal{P}(\mathcal{W})$ such that $P \ll Q$, with probability at least $1-\delta$ over the randomness of $\widetilde{Z}$ and $U$,
\begin{equation}\label{cmi_catoni}
    \mathop{\rm sup}_{Q \in \mathcal{P}(\mathcal{W})} \mathbb{E}_{W\sim Q} \left[ \Phi_{\lambda/m}(R(W)) - R_{\rm train}(W,\widetilde{Z},U) \right] \leq \frac{{\rm D}_{\rm KL}(Q || P) + \log(1/\delta)}{\lambda},
\end{equation}
where $R(W) \triangleq \frac{1}{2}[R_{\rm train}(W,\widetilde{Z},U)+R_{\rm test}(W,\widetilde{Z},U)]=\frac{1}{2m}\sum_{i=1}^{2m} \ell(W,s_i)$.
\end{theorem}
By the property of mutual information that conditioning reduces uncertainty, we have $I(W;U|\widetilde{Z}) \leq I(W;U) \leq m\log 2$ holds, suggesting that the conditional mutual information has a finite upper bound. Eq.~(\ref{cmi}) is consistent with the results of \cite{Steinke2020} in formulation, and the only difference is that $\widetilde{Z}$ should be interpreted as the transductive supersamples. Eq.~(\ref{cmi_catoni}) is the extension of Theorem~3.1.2 given by \cite{Catoni2007} to the random splitting setting. We remark that the prior $P$ is independent of $U$, but may depend on $\widetilde{Z}$. Notice that the randomness of selecting training labels is determined by both $U$ and $\widetilde{Z}$. Thus, the prior $P$ can obtain some but not all information on the selection of training labels, which is similar to Catoni's result where $P$ depends on the dataset $S$. 

Although the mutual information term $I(W;U|\widetilde{Z})$ is bounded, computing its numerical value is still challenging since $W$ may be high-dimensional in real-world application scenarios, particularly deep learning scenarios. To address this issue, various new information measures are proposed \citep{Harutyunyan2021,Hellstrom2022,Wang2023}, which are theoretically smaller than $I(W;U|\widetilde{Z})$ and easier to estimate. Using the concept of transductive supersample, these results can be transported to the transductive learning setting.
\begin{corollary}\label{cor1}
Suppose that $r(\hat{y},y) \in [0,B]$ holds for all $\hat{y} \in \widehat{\mathcal{Y}}$ and $y \in \mathcal{Y}$, where $B > 0$ is a constant. Denote by $f_w(x) \in \mathbb{R}^K$ the prediction of the model and $F_{i} \triangleq (f_W(x_{\widetilde{Z}_{i,0}}), f_W(x_{\widetilde{Z}_{i,1}}))$ the sequence of predictions. Denote by $L_{i} \triangleq (\ell(W,s_{\widetilde{Z}_{i,0}}), \ell(W,s_{\widetilde{Z}_{i,1}}))$ the sequence of loss values and $\Delta_{i} \triangleq \ell(W,s_{\widetilde{Z}_{i,1}}) - \ell(W,s_{\widetilde{Z}_{i,0}})$ the difference of loss value. We have
\begin{align}
    & \left\vert \mathbb{E}_{W,Z} \left[ R_{\rm test}(W, Z) - R_{\rm train}(W, Z) \right] \right\vert \leq \frac{B}{m} \sum_{i=1}^m \mathbb{E}_{\widetilde{Z}} \sqrt{2I^{\widetilde{Z}}(F_i;U_i)}, \label{fmi} \\
    & \left\vert \mathbb{E}_{W,Z} \left[ R_{\rm test}(W,Z) - R_{\rm train}(W,Z) \right] \right\vert \leq \frac{B}{m} \sum_{i=1}^m \mathbb{E}_{\widetilde{Z}} \sqrt{2I^{\widetilde{Z}}(L_i;U_i)} \label{ecmi} , \\
    & \left\vert \mathbb{E}_{W,Z} \left[ R_{\rm test}(W,Z) - R_{\rm train}(W,Z) \right] \right\vert \leq \frac{B}{m} \sum_{i=1}^m \mathbb{E}_{\widetilde{Z}} \sqrt{2I^{\widetilde{Z}}(\Delta_i;U_i)} \label{idcmi}.
\end{align}
\end{corollary}

According to the type of conditional mutual information they contain, the bounds in Eqs.~(\ref{fmi}), (\ref{ecmi}), and (\ref{idcmi}) are termed as transductive $f$-CMI bound \citep{Harutyunyan2021}, transductive e-CMI bound \citep{Hellstrom2022}, and transductive Id-CMI bound \citep{Wang2023}, respectively. The only difference between these results and the previous one is that here $\widetilde{Z}$ represents the \emph{transductive supersamples}. In applications, the prediction of the model is a low-dimension vector and thus reduces the difficulty of computing the conditional mutual information $I(W;U|\widetilde{Z})$. Note that $L_i$ in Eq.~(\ref{ecmi}) and $\Delta_i$ in Eq.~(\ref{idcmi}) are two-dimensional and one-dimensional random variables, yielding more computationally convenient and sharper bounds. However, these bounds can not explicitly depict some factors that affect generalization such as the norm of weights or the sharpness of the loss landscape, and they are accordingly called black-box algorithms generalization bounds \citep{Harutyunyan2021}. As a comparison, the bounds in Theorem~\ref{thm3} and Section~\ref{mi_bounds} are more informative to understanding generalization (see Section~\ref{opti_sec} for more details), yet their numerical values are difficult to compute. To summarize, different types of results provide different perspectives for us to understand the generalization abilities of transductive learning algorithms. Now we are in a place to discuss more general cases that $u=km,k\in \mathbb{N}_+$ by extending the definition of transductive supersample. 
\begin{definition}[$k$-Transductive Supersample]\label{def2}
For $k\in \mathbb{N}_{+}$, let $m=\frac{n}{k+1}$ and $u=km$. Performing $m$ rounds of sampling without replacement from the set $\{1,\ldots,n\}$ and each time $(k+1)$ elements are chosen. For each $i\in [m]$, arrange the elements from the $i$-th sampling into a sequence $\widetilde{Z}_i \triangleq (\widetilde{Z}_{i,0},\ldots,\widetilde{Z}_{i,k})$ in ascending order, that is, $\widetilde{Z}_{i,0} < \cdots < \widetilde{Z}_{i,k}$. The transductive supersample is defined as a sequence $\widetilde{Z} \triangleq (\widetilde{Z}_1, \ldots, \widetilde{Z}_m)$.
\end{definition}
Clearly, Definition~\ref{def} is a special case of Definition~\ref{def2} with $k=1$. Similarly, we extend the definition of the selector sequence as $U\triangleq(U_1,\ldots,U_m) \sim \mathrm{Unif}(\mathrm{Perm}((0,\ldots,k)))^m$, where $U_i \triangleq (U_{i,0},\ldots,U_{i,k}), i\in [m]$ is a sequence of length $(k+1)$. With these notations, we define 
\begin{equation}
    \mathscr{Z}(\widetilde{Z},U) = (\widetilde{Z}_{1,U_{1,0}}, \ldots, \widetilde{Z}_{m,U_{m,0}}, \widetilde{Z}_{1,U_{1,1}}, \ldots, \widetilde{Z}_{m,U_{m,1}}, \ldots, \widetilde{Z}_{1,U_{m,k}}, \ldots, \widetilde{Z}_{m,U_{m,k}})
\end{equation}
as the sequence induced by $\widetilde{Z}$ and $U$. Then the first $m$ elements in $\mathscr{Z}(\widetilde{Z},U)$ are the indices of training data points, and others are the indices of test data points. Now let us see a short example of this definition. For simplicity, we let $m=2$ and $k=3$. Assuming that we have completed a sampling of $[6]$. The obtained elements from the first and second times are $\{3,5,2\}$ and $\{1,6,4\}$, respectively. For the elements $\{3,5,2\}$, we put it into a sequence and get $\widetilde{z}_1 = (2, 3, 5)$. Similarly we have $\widetilde{z}_2=(1,4,6)$. Also, assuming that the realization of $U$ is $u=((1,0,2),(2,1,0))$, that is, $u_1=(1,0,2)$ and $u_2=(2,1,0)$. Then the sequence $\mathscr{Z}(\widetilde{Z},U)$ induced by $\widetilde{Z}$ and $U$ is $\mathscr{Z}(\widetilde{z},u) = (\widetilde{z}_{1,u_{1,0}},\widetilde{z}_{2,u_{2,0}},\widetilde{z}_{1,u_{1,1}},\widetilde{z}_{2,u_{2,1}},\widetilde{z}_{1,u_{1,2}},\widetilde{z}_{1,u_{2,2}}) = (3,6,2,4,5,1)$. In this way, $Z$ can be replaced by $\widetilde{Z}$ and $U$, which enables us to extend the results in Theorem~\ref{thm3} and Corollary~\ref{cor1} to the case that $u=km, k\in \mathbb{N}_+$. Since the formulations are similar, we place the details in Appendix~\ref{proof3}. Since $U$ has more possible values to take with the increase of $k$, we need to accordingly increase the samples of $U$ to reduce the estimated error of conditional mutual information. 

We close this part by discussing the connection between the transductive supersample and the supersample under the leave-one-out CMI framework \citep{Haghifam2022}, where the selector variable is defined as $U={\rm Unif}([n+1])$. The supersample under this framework is denoted by $\widetilde{Z}=(\widetilde{Z}_1,\ldots,\widetilde{Z}_{n+1}) \in (\mathcal{X}\times \mathcal{Y})^{n+1}$, where $\widetilde{Z}_i \sim P_{X,Y}, i \in [n+1]$ are independent and identically distributed (i.i.d.) data points. The role of $U$ is selecting an entry from $\widetilde{Z}$, which will be used as a test data point. And the remaining entries in $\widetilde{Z}$ are used as training data. Formally, the training set and test set are given by $\{\widetilde{Z}_{U}\}$ and $\{\widetilde{Z}_j | j \ne U\}$ respectively. Recall that the random variable $U_i$ is defined by $U_i \sim \mathrm{Unif}(\mathrm{Perm}((0,\ldots,k)))$ for a fixed index $i\in [m]$ under the transductive learning setting that $u=km$. If we redefine $U_i,i\in [m]$ as $U_i \sim \mathrm{Unif}(\mathrm{Perm}((1,\ldots,n+1)))$, one can verify that $U_{i,j} \sim \mathrm{Unif}([n+1])$ for $j\in [n+1]$, thereby recovering the definition of $U$ used in leave-one-out CMI framework. And, recall that $\widetilde{Z}_i,i\in [m]$ is a sequence of length $(k+1)$, which plays a role similar to the arbitrarily-sized subset used by \cite{Haghifam2022}. However, this connection only lies in the case that $u=km,k\in \mathbb{N}_+$. For common cases where $m$ and $u$ are arbitrary integers, we do not yet know how to extend the CMI framework to the transductive learning setting.

\subsection{Transductive PAC-Bayesian Bounds under the Random Splitting Setting}\label{pac_split}

In this section, we establish generalization bounds for the random splitting setting of transductive learning algorithms in the context of PAC-Bayesian theory, which is closely connected with information theory since the foundation of both is the change of measure technique. The overall procedure is similar to that used in Section~\ref{mi_bounds}, where two main ingredients are applying the change of measure technique and deriving an upper bound for the moment-generating function of the transductive generalization gap via the martingale approach. Similar to the PAC-Bayesian bounds derived in the inductive setting, the transductive PAC-Bayesian bounds also include the KL divergence between a posterior $Q$ and a prior $P$. The theoretical results are summarized in the following theorem.

\begin{theorem}\label{thm4}
    Suppose that $\ell(w,s) \in [0,B]$ holds for all $w \in \mathcal{W}$ and $s \in \{s_i\}_{i=1}^n$, where $B > 0$ is a constant. Let $P \in \mathcal{P}(\mathcal{W})$ be a prior distribution independent of $Z$. For any $0 < \delta < 1$, $\lambda > 0 $, and $Q \in \mathcal{P}(\mathcal{W})$ such that $Q \ll P$, we have
    \begin{equation}\label{average_all}
        \mathbb{E}_Z \mathbb{E}_{W \sim Q} [R_{\rm test}(W, Z) - R_{\rm train}(W, Z)] \leq \frac{\lambda B^2 C_{m,u}(m+u)}{8mu} + \frac{\mathbb{E}_Z \left[{\rm D}_{\rm KL}(Q || P) \right]}{\lambda}.
    \end{equation}
    With probability at least $1-\delta$ over the randomness of $Z$, we have
    \begin{equation}\label{average}
        \mathop{\rm sup}_{Q \in \mathcal{P}(\mathcal{W})} \mathbb{E}_{W \sim Q} [R_{\rm test}(W, Z) - R_{\rm train}(W, Z)] \leq \frac{\lambda B^2 C_{m,u}(m+u)}{8mu} + \frac{{\rm D}_{\rm KL}(Q || P) + \log (1/\delta)}{\lambda}.
    \end{equation}
    With probability at least $1-\delta$ over the randomness of $Z$ and $W \sim Q$ we have
    \begin{equation}\label{single}
        R_{\rm test}(W, Z) \leq R_{\rm train}(W, Z) + \frac{\lambda B^2 C_{m,u}(m+u)}{8mu} + \frac{1}{\lambda} \left( \log \frac{\dif Q}{\dif P} + \log (1/\delta) \right).
    \end{equation}
\end{theorem}

Eq.~(\ref{average_all}) is a bound for the average transductive generalzation gap $\mathbb{E}_Z\mathbb{E}_{W\sim P}[\mathcal{E}(W,Z)]$. Eq.~(\ref{average}) is an average bound that holds with high probability under the distribution $P_Z$. Eq.~(\ref{single}) is a singe-draw bound that holds with high probability under the distribution $P_{W,Z}$. Following \cite{Pierre2021}, denote by $\mathbb{E}_Z[Q]$ the probability distributions
\begin{equation}
    \frac{\dif \mathbb{E}_Z\left[Q \right]}{\dif P}(w) \triangleq \mathbb{E}_Z \left[ \frac{\dif Q}{\dif P}(w) \right], w\in \mathcal{W}.
\end{equation}
By the equality $\mathbb{E}_Z\left[ {\rm D}_{\rm KL}(Q||P) \right] = \mathbb{E}_Z\left[ {\rm D}_{\rm KL}(Q||\mathbb{E}_Z[Q]) \right] + {\rm D}_{\rm KL}(\mathbb{E}_Z[Q]||P)$ and the fact that $I(W;Z)=\mathbb{E}_Z\left[ {\rm D}_{\rm KL}(Q||\mathbb{E}_Z[Q]) \right]$, choosing $P = \mathbb{E}_Z[Q]$ as the prior and minimizing the right hand side (r.h.s.) of Eq.~(\ref{average_all}) w.r.t. $\lambda$ enable us to recover the information-theoretic bound in Eq.~(\ref{first}). Now let us turn to Eq.~(\ref{average}), whose r.h.s. holds for any probability measure $Q \in \mathcal{P}(\mathcal{W})$. For a fixed prior $P$, Lemma~\ref{lemma} shows that the optimal posterior $Q$ minimizing the term $\lambda \mathbb{E}_{W\sim Q}[R_{\rm train}(W,Z)] + {\rm D}_{\rm KL}(Q || P)$ is the Gibbs distribution $P_{e^{-\lambda R_{\rm train}}}$ given by
\begin{equation}
    \frac{\dif P_{e^{-\lambda R_{\rm train}}}}{\dif P}(w) = \frac{e^{-\lambda R_{\rm train}(w,Z)}}{\int_{\mathcal{W}} e^{-\lambda R_{\rm train}(w',Z)}\dif P(w')}.
\end{equation}
This is consistent with the result under the random sampling setting of transductive learning \citep[Theorem~3.1.2]{Catoni2007}. 
Notice that Eqs.~(\ref{average}) and (\ref{single}) are Catoni-typed bounds since they contain an extra parameter $\lambda$. By minimizing the right hand side of them, we obtain a formulation of the optimal value of $\lambda$, which contains the complexity term ${\rm D}_{\rm KL}(Q || P)$. Since $\lambda$ needs to be specified before observing $Z$ and the posterior $Q$ generally depends on $Z$, this approach is not feasible in practice. An alternative approach that enables $\lambda$ to be optimized, proposed by \cite{Pierre2021,Catoni2007}, is to optimize $\lambda$ over a predefined set $\Lambda$ with finite cardinality, as shown in the following corollary.

\begin{corollary}\label{cor_pac}
    Denote by $\Lambda$ a predefined set that satisfies $\vert \Lambda \vert < \infty$. Under the assumption of Theorem~\ref{thm4}, for any $0 < \delta < 1$ and $Q \in \mathcal{P}(\mathcal{W})$ such that $Q \ll P$, with probability at least $1-\delta$ over the randomness of $Z$, we have
    \begin{equation*}
    \begin{aligned}
        & \mathbb{E}_{W \sim Q} [R_{\rm test}(W, Z) - R_{\rm train}(W, Z)] \leq  \mathop{\rm inf}_{\lambda \in \Lambda} \left\{ \frac{\lambda C_{m,u}(m+u)}{8mu} + \frac{{\rm D}_{\rm KL}(Q || P) + \log (\vert \Lambda \vert/\delta)}{\lambda} \right\}.
    \end{aligned}
    \end{equation*}
    Particularly, for $\Lambda \triangleq \left\{ e^i, i\in \mathbb{N} \right\} \cap [1,mu/(m+u)]$, we have
    \begin{equation*}
    \begin{aligned}
        & \mathbb{E}_{W \sim Q} [R_{\rm test}(W, Z) - R_{\rm train}(W, Z)] \\
        \leq & \mathop{\rm inf}_{ \lambda \in \left[1, \frac{mu}{m+u} \right]} \left\{ \frac{\lambda C_{m,u}(m+u)}{8mu} + \frac{e}{\lambda} \left( {\rm D}_{\rm KL}(Q || P) + \log \left( \frac{1}{\delta}\log \left( \frac{mu}{m+u} \right) \right) \right) \right\}.
    \end{aligned}
    \end{equation*}
\end{corollary}
Corollary~\ref{cor_pac} shows that demonstrates $\lambda$ to be optimized resulting in an additional term $\log \log (mu/(m+u))$ appearing in the bound. Now we are in a place to compare our result to the latest explicit transductive PAC-Bayesian bound derived under the random splitting setting, which is given by \cite{Begin2014} and we restate it as follows.

\begin{theorem}[\citealp{Begin2014}, Corollary~7(b)]\label{thm_begin}
    Suppose that the number of training and test data points satisfies $m, u \geq 20$. Let $P \in \mathcal{P}(\mathcal{W})$ be a prior distribution independent of $Z$. For any $0<\delta<1$ and $Q \in \mathcal{P}(\mathcal{W})$ such that $Q \ll P$, with probability at least $1-\delta$ over the randomness of $Z$, we have
    \begin{equation*}
        \mathbb{E}_{W\sim Q} \left[ R_{\rm test}(W,Z) - R_{\rm train}(W,Z) \right] \leq \sqrt{\frac{m+u}{2mu} \left( {\rm D}_{\rm KL}(Q || P) + \log \left( \frac{3\log (m)}{\delta} \sqrt{\frac{mu}{m+u}}\right) \right)}.
    \end{equation*}
\end{theorem}

Different from our result, Theorem~\ref{thm_begin} presents a Langford-Seeger-Maurer-typed PAC-Bayesian bound and does not include an additional parameter $\lambda$. Accordingly, the overall formulation of the bound is more concise than ours, and the order of the bound w.r.t. the number of training data points $m$ and test data points $u$ is clear. However, the physical meaning of the bound in Theorem~\ref{thm_begin} is implicit since the optimal posterior can not be reflected from the bound. In contrast, the physical meaning of our result (Eq.~\ref{average}) is explicit as it reveals that the optimal posterior is the Gibbs distribution. To directly compare the order of our result with the bound given in Theorem~\ref{thm_begin}, inspired by the work of \cite{Pierre2021}, we reformulate the latter by introducing the parameter $\lambda$ by using the inequality $\sqrt{(\lambda a/2)(2b/\lambda)} \leq \lambda a/4 + b/\lambda$ that holds for any $a,b,\lambda > 0$:
\begin{equation*}
\begin{aligned}
    & \mathbb{E}_{W \sim Q} [R_{\rm test}(W, Z)] \\
    \leq & \mathbb{E}_{W \sim Q} [R_{\rm train}(W, Z)] + \mathop{\rm inf}_{\lambda > 0} \left\{ \frac{\lambda(m+u)}{8emu} + \frac{e}{\lambda} \left( {\rm D}_{\rm KL}(Q || P) + \log \left( \frac{3\log (m)}{\delta} \sqrt{\frac{mu}{m+u}}\right) \right) \right\}.
\end{aligned}
\end{equation*}
Using the inequality $\log x \leq \sqrt{x}$ that holds for all $x > 0$, we have $\log(mu/(m+u)) \leq \sqrt{mu/(m+u)}$. Therefore, Theorem~\ref{thm4} with a geometric grid (Corollary~\ref{cor_pac}) gives better result than Theorem~\ref{thm_begin} by saving a factor $\log \left(3\log (m) \right)$ when $m$ and $u$ are large. However, Theorem~\ref{thm_begin} may provide a tighter bound if $m$ and $u$ is small since its constant of the first term is smaller than that in Theorem~\ref{thm4}. However, we emphasize that Theorem~\ref{thm4} is essentially different than Theorem~\ref{thm_begin}, and its assumption is much weaker. Specifically, Theorem~\ref{thm_begin} requires that $\ell(\cdot)$ is zero-one loss, and the values of $m$ and $u$ must satisfy $m, u \geq 20$. In contrast, Theorem~\ref{thm4} applies to any bounded loss, and there are no constraints on the value of $m$ and $u$. The reason is that we derive the upper bound for the moment-generating function by the martingale technique, while \cite{Begin2014} firstly obtain an implicit bound by introducing a variant of the binary relative entropy \citep{Germain2009} and then convert it to an explicit one by the Pinsker’s inequality. As for the implicit one \citep[Corollary~7~(a)]{Begin2014}, we do not know how to compare it with ours, and we are accordingly not sure which one is sharper.

One of the most important insights delivered by PAC-Bayesian bounds is that the generalization ability of a model is related to the flatness of its loss landscape, and a flat minimum is beneficial for generalization. With the help of Theorem~\ref{thm4}, this result can be extended to the transductive learning setting when $\ell(\cdot)$ is the zero-one loss.

\begin{corollary}\label{cor2}
    Suppose that (\romannumeral1) $m \geq 2, u \geq 3$ or $m \geq 3, u \geq 2$ and (\romannumeral2) $R_{\rm test}(W,Z) \leq \mathbb{E}_{{\bm \epsilon} \sim \mathcal{N}({\bm 0}, \sigma^2 \mathbf{I}_d)} [R_{\rm test}(W+{\bm \epsilon},Z)]$ holds for all sampling sequence $Z$, where $W \in \mathbb{R}^d$ is the parameter returned by a given transductive learning algorithm and ${\bm \epsilon} \sim \mathcal{N}({\bm 0}, \sigma^2 \mathbf{I}_d)$. For any $\lambda > 0$ and $0<\delta<1$, with probability at least $1-\delta$ over the randomness of $Z$ we have
    \begin{small}
    \begin{equation*}
    \begin{aligned}
        R_{\rm test}(W, Z) \leq & \max_{\Vert \bm \epsilon \Vert_2 \leq \rho} R_{\rm train}(W+\bm{\epsilon}, Z) + \frac{(\lambda C_{m,u}+8)(m+u)}{8mu} \\
        & + \frac{1}{\lambda} \left( \frac{1}{2} \left[ 1 + d \log \bigg( 1 + \frac{\big( 1 + \widetilde{C}_{m,u} \big)^2\Vert W \Vert^2_2}{\rho^2} \bigg) \right] + \log \left( \frac{1}{6\delta} \right) + 2\log \left( \frac{6\pi mu}{m+u} \right) \right),
    \end{aligned}
    \end{equation*}
    \end{small}%
    where $\widetilde{C}_{m,u} \triangleq \sqrt{2\log (mu/(m+u))/d}$ and $\rho \triangleq \sigma (1+\widetilde{C}_{m,u})\sqrt{d}$.
\end{corollary}
The assumption $R_{\rm test}(W,Z) \leq \mathbb{E}_{{\bm \epsilon} \sim \mathcal{N}({\bm 0}, \sigma^2 \mathbf{I}_d)} [R_{\rm test}(W+{\bm \epsilon},Z)]$ used in Corollary~\ref{cor2} means that adding random Gaussian noise to the parameter of the model after training does not decrease its performance on unlabeled data points. We remark that the parameter $\lambda$ can not be directly optimized since its optimal value depends on $\Vert W \Vert_2$. The term $\max_{\Vert \bm \epsilon \Vert_2 \leq \rho} R_{\rm train}(W+\bm{\epsilon}, Z)$ characterizes the change of loss landscape within a ball with $W$ as the center and $\rho$ as the radius. Formally, we call $W$ as sharp minima if the loss values around it differ significantly from itself, namely $R_{\rm train}(W+\bm{\epsilon}, Z)$ is significantly larger than $R_{\rm train}(W, Z)$. Therefore, Corollary~\ref{cor2} suggests that a flat optima could have better transductive generalization performance. A classical approach \citep{Foret2021} to ensure the flatness of loss landscape is solving a minimax optimization problem $\mathop{\min}_{w} \mathop{\max}_{\Vert \bm{\epsilon} \Vert_2 \leq \rho} R_{\rm train}(w + \bm{\epsilon}, Z)$. By converting the minimax optimization into a bi-level optimization problem and solving it via the hypergradient algorithm, \cite{Chen2023} show that in terms of the recommendation task, GNNs with flatter minima have a better generalization ability than those with sharper minima. This observation serves as strong evidence of Corollary~\ref{cor2} and also demonstrates that we can study the generalization ability of deep transductive learning models (such as GNNs) by analyzing the flatness of its loss landscape. Particularly, recent work of \cite{Tang2023} reveals that the initial residual and identity mapping techniques adopted in GCNII \citep{Chen2020GCNII} can help the model maintain the generalization gap when the number of layers increases. Investigating how these techniques affect the flatness of the loss landscape and ultimately affect the generalization of the model is worth exploring. Besides, our results can also enhance the applicability of existing generalization analysis for GNNs. For example, combining Theorem~\ref{thm4} with the technique used by \cite{Neyshabur2018}, results of \cite{Liao2021} could be extended to the transductive learning setting. Recently, \cite{Lee2024} have combined the techniques of \cite{Neyshabur2018} and Theorem~\ref{thm_begin} to establish a generalization guarantee for a deterministic triplet classifier. They also derive the first generalization bounds for knowledge graph representation learning. The posterior distribution used by \cite{Neyshabur2018,Lee2024} is constructed by adding random Gaussian noise to the parameter of the model, which is similar to ours. Besides, the assumption of result given by \cite{Lee2024} is weaker than ours since it does not require that adding Gaussian perturbation should not decrease the test error. However, one can easily obtain analogous results of \cite{Neyshabur2018,Lee2024} by applying their technique to our new transductive PAC-Bayesian bound in Theorem~\ref{thm4}.

We close this part by giving some extra comments on Corollary~\ref{cor2}. Firstly, we have that $\rho = \mathcal{O}(\sqrt{d})$, and its numerical value is generally not available. The reason is that $\rho$ depends on $\sigma$ and we are unable to obtain the exact numerical value of $\sigma$ in the assumption. And, minimizing the term $\max_{\Vert \bm \epsilon \Vert_2 \leq \rho} R_{\rm train}(W+\bm{\epsilon}, Z)$ will result in the loss landscape around $W$ being excessively flat, since $\rho = \mathcal{O}(\sqrt{d})$. Therefore, $\rho$ is regarded as a hyperparameter in practice \citep{Foret2021}. Secondly, Corollary~\ref{cor2} still suffers from some issues. On the one hand, if $d$ is sufficiently large, the term $d \log(1 + 1/(\sigma^2 d)) \approx O(1))$ may be independent of $d$, which implies that there could be no impact of the dimension in the first term of the bound when the dimension is sufficiently large. On the other hand, $\sigma$ that appears in the assumption is data-free, which results in the variance of the posterior $Q \triangleq \mathcal{N}(O,\sigma^2 \mathbf{I}_d)$ being data-independent. We point out here that these issues essentially come from the proof of \cite{Foret2021} and they naturally appear since our proof for Corollary~\ref{cor2} generally follows theirs. However, we would like to clarify that the reason for presenting this corollary is to briefly reveal the application potential of our theoretical results. Indeed, the aforementioned issues could be tackled by using new proof techniques. For example, to address the shortcoming that the variance of the posterior is data-independent, we can adopt the Fisher information matrix as the variance of perturbed noise \citep{Kim2022}, which will make the derived result more in line with real-world scenarios.

\subsection{Transductive PAC-Bayesian Bounds under the Random Sampling Setting}\label{con_pac}

So far, all results are derived under the random splitting setting (Section~\ref{trc_pre}). In this section, we turn to the random sampling setting (Section~\ref{trc2}) where a sequence of data points is provided. Establishing risk bounds under this setting is challenging because the data points could come from different distributions and there could exist dependence among them. Theorem~10.1 of \cite{Vapnik1998} shows that if the data points are i.i.d., the risk bounds derived under the random splitting setting can be transported to the random sampling setting. Later, \cite{Catoni2003,Catoni2007} proposes another approach that directly derives risk bounds without transporting results from the random splitting setting, which only requires that the data distribution and the prior be exchangeable \citep[Chapter~2.1]{Catoni2003} or partially exchangeable \citep[Chapter~3.1]{Catoni2007}. Here, the prior is defined as a function $P: (\mathcal{X}\times \mathcal{Y})^{n} \to \mathcal{P}(\mathcal{W})$. To proceed, we first introduce the following definitions.

\begin{definition}[Exchangeable and Partially Exchangeable Distribution]
    Let $\Pi$ be the set of all bijections $\pi: [n] \to [n]$. We say $P_{S_1,\ldots,S_n}$ is an exchangeable distribution if $P_{S_1,\ldots,S_n}=P_{S_{\pi(1)},\ldots,S_{\pi(n)}}$ holds for all $\pi \in \Pi$. If $u=km,k\in \mathbb{N}_+$, let $\Pi$ be the set of all bijections $\pi: \{0,\ldots,k\} \to \{0,\ldots,k\}$. We say $P_{S_1,\ldots,S_{(k+1)m}}$ is a partially exchangeable distribution if for all $\pi \in \Pi, i \in [m]$, the following holds:
    \begin{equation*}
    \begin{aligned}
        & P_{S_1,\ldots,S_i,\ldots,S_m,\ldots,S_{1+mk},\ldots,S_{i+mk},\ldots,S_{(k+1)m}} = P_{S_1,\ldots,S_{i+m\pi(0)},\ldots,S_m,\ldots,S_{1+mk},\ldots,S_{i+m\pi(k)},\ldots,S_{(k+1)m}}.
    \end{aligned}
    \end{equation*}
\end{definition}

\begin{definition}[Exchangeable and Partially Exchangeable Prior]
    Let $\Pi$ be the set of all bijections $\pi: [n] \to [n]$. We say $P$ is an exchangeable prior if $P((s_1,\ldots,s_n))=P((s_{\pi(1)},\ldots,s_{\pi(n)}))$ holds for all $(s_1,\ldots,s_n) \in (\mathcal{X}\times \mathcal{Y})^{n}$ and $\pi \in \Pi$. If $u=km,k\in \mathbb{N}_+$, let $\Pi$ be the set of all bijections $\pi: \{0,\ldots,k\} \to \{0,\ldots,k\}$. We say $P$ is a partially exchangeable prior if for all $(s_1,\ldots,s_n) \in (\mathcal{X}\times \mathcal{Y})^{n}, \pi \in \Pi, i\in [m]$, the following holds:
    \begin{equation*}
    \begin{aligned}
        & P \left( (s_1,\ldots,s_i,\ldots,s_m,\ldots,s_{1+mk},\ldots,s_{i+mk},\ldots,s_{(k+1)m}) \right) \\
        = & P\left( (s_1,\ldots,s_{i+m\pi(0)},\ldots,s_m,\ldots,s_{1+mk},\ldots,s_{i+m\pi(k)},\ldots,s_{(k+1)m}) \right).
    \end{aligned}
    \end{equation*}
\end{definition}

Clearly, exchangeable distribution (or partially exchangeable distribution) refers to those distributions $P_{S_1,\ldots,S_n}$ that remain unchanged after applying a permutation on the full indices $(1,\ldots,n)$ (or partial indices $(i,m+i,\ldots,km+i),i\in [m]$) of the random variable sequence $(S_1,\ldots,S_n)$.  Similarly, applying permutation on the indices $(1,\ldots,n)$ (or partial indices $(i,m+i,\ldots,km+i),i\in [m]$) of input variables $(s_1,\ldots,s_n)$ does not change the value of an exchangeable prior (or partially exchangeable prior). Now we elucidate why these concepts are needed in assumptions. The concept of exchangeable distribution (or partially exchangeable distribution) is to handle the dependency between training error and test error. It allows us to implicitly take expectation over the random permutation on the data sequence and thus either connect the transductive training error with total error on full data or transport the result from the random splitting setting to the random sampling setting. To see this, denote by $Z=(Z_1,\ldots,Z_{n})$ the sequence obtained by sampling without replacement from $[n]$. Assuming that $S \sim P_{(S_1,\ldots,S_n)}$ and $P_{(S_1,\ldots,S_n)}$ is an exchangeable distribution, for any $w \in \mathcal{W}$ we have
\begin{equation}\label{perm1}
    \mathbb{E}_S \left[ e^{\frac{1}{u}\sum_{i=m+1}^{m+u} \ell(w,S_i) - \frac{1}{m}\sum_{i=1}^m \ell(w,S_i)} \right] = \mathbb{E}_S \mathbb{E}_Z \left[ e^{\frac{1}{u}\sum_{i=m+1}^{m+u} \ell(w,S_{Z_i}) - \frac{1}{m}\sum_{i=1}^m \ell(w,S_{Z_i})} \right].
\end{equation}
Notice that the term $\frac{1}{u}\sum_{i=m+1}^{m+u} \ell(w,S_{Z_i}) - \frac{1}{m}\sum_{i=1}^m \ell(w,S_{Z_i})$ in the exponential function of Eq.~(\ref{perm1}) is exactly the transductive generalization gap that we analyze in Section~\ref{mi_bounds} and Section~\ref{pac_split}. Thereby, we can apply the martingale technique to derive an upper bound. If the number of training data points $m$ and test data points $u$ satisfy a certain relation, for example, $u=km,k\in \mathbb{N}_+$, we can use $U = (U_1,\ldots,U_m) \sim \mathrm{Unif}(\mathrm{Perm}((0,\ldots,k)))^m$ instead of $Z$ to permute the data sequence and thus establish a connection between training error and total error. Concretely, assume that $S=(S_1,\ldots,S_{(k+1)m}) \sim P_{S_1,\ldots,S_{(k+1)m}}$ and $P_{S_1,\ldots,S_{(k+1)m}}$ is a partially exchangeable distribution, for any $w \in \mathcal{W}$ we have
\begin{equation}\label{perm2}
\begin{aligned}
    & \mathbb{E}_S \left[ e^{\frac{1}{m}\sum_{i=1}^m \ell(w,S_i)} \right] = \mathbb{E}_S \mathbb{E}_U \left[ e^{\frac{1}{m}\sum_{i=1}^m \ell(w,S_{i+mU_{i,0}})} \right] \\
    = & \mathbb{E}_S \mathbb{E}_U \Bigg[ \prod_{i=1}^m e^{\frac{1}{m} \ell(w,S_{i+mU_{i,0}})} \Bigg] = \mathbb{E}_S \Bigg[ \prod_{i=1}^m \mathbb{E}_U \left[ e^{\frac{1}{m} \ell(w,S_{i+mU_{i,0}})} \right] \Bigg] \\
    = & \mathbb{E}_S \Bigg[ \exp\bigg\{ \sum_{i=1}^m \log \left( \mathbb{E}_U \left[ e^{\frac{1}{m} \ell(w,S_{i+mU_{i,0}})} \right]\right)\bigg\} \Bigg] \\
    \leq & \mathbb{E}_S \bigg[ \exp\bigg\{ m \log \bigg( \frac{1}{m}\sum_{i=1}^m \mathbb{E}_U \left[ e^{\frac{1}{m} \ell(w,S_{i+mU_{i,0}})} \right]\bigg)\bigg\} \bigg] \leq \mathbb{E}_S\left[ \exp\{ \Phi_{1/m}(\bar{R}(w,S)) \right],
\end{aligned}
\end{equation}
where $\bar{R}(w,S) \triangleq \frac{1}{(k+1)m} \sum_{i=1}^{(k+1)m} \ell(w,S_i)$ is the total error. In this way, we can build a connection between the training error $\frac{1}{m}\sum_{i=1}^m \ell(w,S_i)$ and the total error $\bar{R}(w,S)$. The above technique first appeared in the proof of Theorem~3.1.2 given by \cite{Catoni2007} but was presented with different notations. Specifically, in Definition~3.1.1 of \cite{Catoni2007}, he introduces $m$ circular permutations to permute the sequence $(S_1,S_2,\ldots,S_{(k+1)m})$, where the $i$-th one only applied to partial indices $(S_{i},S_{i+m},\ldots,S_{i+km})$ for $i\in [m]$ and leaving the rest indices unchanged, which is mathematically equivalent to use $U_i \sim \mathrm{Unif}(\mathrm{Perm}((0,\ldots,k)))$ to permute the subsequence $(S_{i},S_{i+m},\ldots,S_{i+km})$ for $i\in [m]$. Further, if $m=u$, the selector sequence can be simplified as $U = (U_1,\ldots,U_m) \sim \mathrm{Unif}(\{0,1\})^m$. Assume that $S \sim P_{(S_1,\ldots,S_{2m})}$ and $P_{(S_1,\ldots,S_{2m})}$ is a partially exchangeable distribution, for any $w \in \mathcal{W}$,
\begin{equation}\label{perm3}
\begin{aligned}
    & \mathbb{E}_{S}\left[ e^{\frac{1}{m} \sum_{i=m+1}^{2m} \ell(w,S_i) - \frac{1}{m} \sum_{i=1}^{m} \ell(w,S_i)} \right] = \mathbb{E}_{S}\left[ e^{\frac{1}{m} \sum_{i=1}^{m} \left( \ell(w,S_{m+i}) - \ell(w,S_i) \right)} \right] \\
    = & \mathbb{E}_{S} \mathbb{E}_U \left[ e^{\frac{1}{m} \sum_{i=1}^{m} (-1)^{U_i} \left( \ell(w,S_{i+m}) - \ell(w,S_{i}) \right)} \right] \\
    = & \mathbb{E}_{S} \mathbb{E}_U \Bigg[ \prod_{i=1}^m e^{\frac{(-1)^{U_i}}{m} \left( \ell(w,S_{i+m}) - \ell(w,S_{i}) \right)} \Bigg] = \mathbb{E}_S \Bigg[ \prod_{i=1}^m \mathbb{E}_{U_i} \bigg[ e^{\frac{(-1)^{U_i}}{m} \left( \ell(w,S_{i+m}) - \ell(w,S_{i}) \right)} \bigg] \Bigg] \\
    = & \mathbb{E}_S \Bigg[ \prod_{i=1}^m \cosh \Big\{ \frac{1}{m}\big(\ell(w,S_{i+m}) - \ell(w,S_{i})\big) \Big\} \Bigg] 
    \leq \mathbb{E}_S \left[ e^{\frac{1}{2m^2}\sum_{i=1}^m \left( \ell(w,S_{m+i}) - \ell(w,S_i) \right)^2} \right].
\end{aligned}
\end{equation}
The last term in Eq.~(\ref{perm3}) can be bounded via various techniques. For the case that $\ell(\cdot)$ is the zero-one loss, \cite{Catoni2003} shows that it equals to $\mathbb{E}_S \big[ e^{\frac{1}{2m^2}\sum_{i=1}^{2m} \ell(w,S_{i})} \big]$ in the proof of Lemma~2.1 given by \cite{Catoni2003}. Further, \cite{Audibert2007} show that for any type of losses, it can be upper bounded by $\mathbb{E}_S \big[ e^{\frac{1}{m^2}\sum_{i=1}^{2m} \ell^2(w,S_{i})} \big]$ via the AM–GM inequality. To summarize, exchangeable distribution or partially exchangeable distribution enables us to handle the dependence between training error and test error, and a stronger assumption on the relation between $m$ and $u$ allows us to use more sophisticated techniques. Moreover, the concept of exchangeable prior (or partially exchangeable prior) is to ensure that the information about selecting training labels is not carried by the prior $P$. Recall that applying permutation on the indices of input variables does not change the value of the prior $P$. Although the prior $P$ is allowed to depend on the full dataset $S$, it will always return the same parameter regardless of the selection of training labels and thus could not carry any information on the training label selection. We remark that this assumption plays the same role as that requiring $P$ to be independent of $Z$ in the random splitting setting. 

Now we can extend Theorem~\ref{thm4} to the random sampling setting, inspired by the connection between the random splitting setting and the random sampling setting \citep[Theorem~10.1]{Vapnik1998}. For the convenience of comparison, we restate results of \cite{Catoni2007} and \cite{Audibert2007} in Theorem~\ref{thm_catoni} and Theorem~\ref{thm_audibert}, respectively. The first one requires that $u=km,k\in \mathbb{N}_+$ while the second one requires that $m=u$.
\begin{corollary}\label{cor3}
    Suppose that $P_{(S_1,\ldots,S_{n})}$ is an exchangeable distribution, and $\ell(w, s) \in [0,B]$ holds for all $w \in \mathcal{W}$ and $s \in (\mathcal{X} \times \mathcal{Y})^{n}$, where $B>0$ is a constant. For any $0<\delta<1$, $\lambda > 0 $, and $Q \in \mathcal{P}(\mathcal{W})$ such that $Q \ll P$, with probability $1-\delta$ over the randomness of $S$,
    \begin{equation*}
        \mathop{\rm sup}_{Q \in \mathcal{P}(\mathcal{W})} \mathbb{E}_{W\sim Q} \left[R_{\rm test}(W, S) - R_{\rm train}(W, S) \right] \leq \frac{{\rm D}_{\rm KL}(Q||P)+\log(1/\delta)}{\lambda} + \frac{\lambda B^2 C_{m,u}(m+u)}{8mu}.
    \end{equation*}
\end{corollary}
\begin{theorem}[\citealp{Catoni2007}, Theorem 3.1.2]\label{thm_catoni}
    Suppose that $P_{(S_1,\ldots,S_{(k+1)m})}$ is a partially exchangeable distribution, and $\ell(\cdot)$ is the zero-one loss. For any $0<\delta<1$, $\lambda > 0 $, and $Q \in \mathcal{P}(\mathcal{W})$ such that $Q \ll P$, with probability at least $1-\delta$ over the randomness of $S$,
    \begin{equation}
        \mathop{\rm sup}_{Q \in \mathcal{P}(\mathcal{W})} \mathbb{E}_{W\sim Q} \left[ \Phi_{\lambda/m}(R(W,S)) - R_{\rm train}(W,S) \right]  \leq \frac{{\rm D}_{\rm KL}(Q || P) + \log (1/\delta)}{\lambda},
    \end{equation}
    where $R(W,S) \triangleq \frac{mR_{\rm train}(W,S)+km R_{\rm test}(W,S)}{(k+1)m}$ is the total error computed on full data.
\end{theorem}
\begin{theorem}[\citealp{Audibert2007}, Lemma~10]\label{thm_audibert}
    Suppose that $P_{S_1,\ldots,S_{2m}}$ is a partially exchangeable distribution. For any $0<\delta<1$, $\lambda > 0 $, and $Q \in \mathcal{P}(\mathcal{W})$ such that $Q \ll P$, with probability at least $1-\delta$ over the randomness of $S$,
    \begin{small}
    \begin{equation*}
        \mathop{\rm sup}_{Q \in \mathcal{P}(\mathcal{W}) } \mathbb{E}_{W\sim Q} \left[R_{\rm test}(W, S) - R_{\rm train}(W, S) \right] \leq \mathbb{E}_{S,W\sim Q} \Bigg[ \frac{\lambda}{m^2} \sum_{i=1}^{2m} \ell^2(W,S_i) \Bigg] + \frac{{\rm D}_{\rm KL}(Q||P)+\log(1/\delta)}{\lambda}.
    \end{equation*}
    \end{small}
\end{theorem}

Overall, Corollary~\ref{cor3} is a non-trivial extension of Theorem~\ref{thm_catoni} to more common cases. Theorem~\ref{thm_catoni} requires that the data distribution and prior be partially exchangeable, which are weaker than Corollary~\ref{cor3}. However, Theorem~\ref{thm_catoni} only applies to the case that $u=km, k\in \mathbb{N}_+$ and $\ell(\cdot)$ is the zero-one loss, yet Corollary~\ref{cor3} applies to arbitrary values of $m,u$ and only requires $\ell(\cdot)$ be bounded. The main difference between our proof and Catoni's proof is the way to tackle the dependence between training error and test error. Specifically, we use the assumption of exchangeable distribution to build a connection between the random sampling setting and random splitting setting and then tackle this dependence by the martingale approach, as we show in Eq.~(\ref{perm1}). Different from ours, Theorem~3.1.2 of \cite{Catoni2007} requires an extra assumption that $u=km,k\in \mathbb{N}_+$, and uses the assumption of partially exchangeable distribution to connect the training error with total error, by which the dependence is eliminated. His key insight is that under the case $u=km,k\in \mathbb{N}_+$, permuting the sequence that contains $(k+1)m$ data points can be achieved by firstly dividing them into $k$ subsequences and then permuting the data points within each subsequence separately. Finally, merging these subsequences together gives a permutation on the entire sequence, as we show in Eq.~(\ref{perm2}). Further, more techniques can be adopted under the assumption that $m=u$, which enables us to connect the transductive generalization gap with the empirical measure associated with the full data points, as shown in Theorem~\ref{thm_audibert}. However, for more general cases where $u$ and $m$ are arbitrary integers, adopting the selector sequence $U$ is not sufficient to depict the randomness, resulting in the technique of \cite{Catoni2003,Catoni2007,Audibert2007} not applicable. The reason is that the number of errors observed in $S$ follows a hypergeometric distribution rather than a binomial distribution \citep{Begin2014}. In contrast, the martingale approach we use still works under these cases, and it can also be applied to scenarios where the objective is a function of matrices (See Section~\ref{opti_sec} for more details). To summarize, we improve previous results by relaxing the assumption on both the type of loss function and the values of $m$ and $u$. Additionally, the technique we adopt has a broad applicability. 

It is worth pointing out that introducing a selector sequence $(U_1,\ldots,U_m)$ to depict the randomness of selecting training labels is motivated from the CMI framework \citep{Steinke2020}, which itself could be tracked back to Catoni's circular permutation technique \citep[Chapter~3.1]{Catoni2007}, as revealed recently in Chapter~6.7 of \cite{Hellstrom2023}. However, we emphasize that the transductive supersample $\widetilde{Z}$ we propose is essentially a new concept. Our key motivation is that, under the random split setting, we could disentangle the uncertainty of training label selection into two parts. One is contained in $\widetilde{Z}$, and the other is contained in $U$, and $\widetilde{Z}$ is independent of $U$. Thus, we only need to perform sampling without replacement to obtain $\widetilde{Z}$ and then apply $U$ to permute it. This allows us to tackle the dependence induced by sampling without replacement. Also, for fixed realizations of $\widetilde{Z}$ that is given by $\widetilde{z}_i=(i,m+i,\ldots,km+i),i\in [m]$, we recover the definition of transductive risk under the random sampling setting. Therefore, the transductive supersample concept serves as a connection between the random sampling setting and random splitting setting and allows us to transport results from the former to the latter. Moreover, we remark that many techniques used in the inductive learning setting are still applicable in the transductive learning setting, for example, using chaining technique \citep{Audibert2007} or Bernstein assumption \citep{Grunwald2021} to derive bounds with a faster rate. Correspondingly, our derived bounds can be further improved by applying these techniques. Recently, \cite{Borja2024} extend Catoni's results under mild assumptions on the loss function. However, these results are derived under the inductive learning setting. It remains unclear whether they apply to the transductive learning setting.

\subsection{Upper Bounds for Adaptive Optimization Algorithms}\label{opti_sec}

As noted previously, one of the advantages of our theoretical results over previous results is that they fully account for the impact of the optimization algorithm on generalization. In this section, we demonstrate this advantage by analyzing AdaGrad \citep{Duchi2011} in the context of transductive learning. Unlike SGD, AdaGrad features an adaptively adjusted learning rate throughout the training process. Let $\{W_t\}_{t=1}^T$ represent the weight sequences on the training trajectory of AdaGrad. Following the work of \cite{Wang2022}, we adopt the setting that mini-batches data points are fixed. Denote $(B_1, \ldots, B_T)$ as the sequence of mini-batches, with $B_t, t\in [T]$ representing the set of data points used in the $t$-th iteration. For simplicity, we assume that the model only minimizes the loss on labeled data points, that is, $B_t = B_t(Z) \subseteq \{s_{Z_i}\}_{i=1}^m$ for $t\in [T]$. And, we assume that each mini-batch contains exactly $b$ data points. The average gradient on $B_t, t\in [T]$ is defined as
\begin{equation*}
    g(w, B_t(Z)) \triangleq \frac{1}{b} \sum_{s \in B_t(Z)} \nabla_{w} \ell(w,s),
\end{equation*}
where $\ell(\cdot)$ represents the objective function. Notice that the loss is computed exclusively on the labeled data point. It might appear that the features of unlabeled data points are not used. However, we point out that in certain scenarios, the features of unlabeled data are implicitly used by the model during loss computation, as seen in transductive graph learning (see Section~\ref{trc_graph} for more details). Additionally, proxy loss on unlabeled data points, such as using pseudo labels generated by the model itself to compute the loss, can be included in the objective function. This would not affect the formulation of the final theoretical result. Recall that the update rule of AdaGrad can be formulated as
\begin{gather*}
    V_{t} = \sum_{k=0}^{t-1} g(W_k, B_k(Z)) \odot g(W_k, B_k(Z)), W_t = W_{t-1} - \frac{\eta}{\sqrt{V_t}+\epsilon} \odot g(W_{t-1}, B_t(Z)), t\in [T],
\end{gather*}
where $W_0$ is the initial parameter and $\eta,\epsilon$ are two predefined hyper-parameters. Notice that $V_t$ is a random variable determined by the sequence $W^{[t-1]} \triangleq (W_0,\ldots,W_{t-1})$ and $Z$. For notational simplicity, we denote the ``adaptive gradient'' as $\Psi(W^{[t-1]},Z) \triangleq \left( {\eta}/({\sqrt{V_t}+\epsilon}) \right) \odot g(W_{t-1}, B_t(Z))$, which is computed by normalizing the current gradient with accumulate squared gradient. Inspired by the work of \cite{Neu2021} and \cite{Wang2022}, we introduce the following auxiliary weight process $\{\widetilde{W}_t\}_{t=1}^T$ defined as
\begin{equation*}
    \widetilde{W}_0 = W_0, \widetilde{W}_t = \widetilde{W}_{t-1} - \Psi(W^{[t-1]},Z) + N_t, N_t = \sigma_t N, t \in [T],
\end{equation*}
where $\{\sigma_t\}_{t\in [T]}$ are predefined hyperparameters and $N$ is a standard Gaussian random variable independent of $W^{[T]}$ and $Z$. Denote by $N_{1:t} \triangleq \sum_{k=1}^t N_k$, then the expectation of the transductive generalization gap $\mathbb{E}_{Z, W_T} \left[ R_{\rm test}(W_T, Z) - R_{\rm train}(W_T, Z) \right]$ can be decomposed into three terms:
\begin{equation}\label{decomposition}
\begin{aligned}
   & \mathbb{E}_{Z, W_T} \left[ R_{\rm test}(W_T, Z) - R_{\rm train}(W_T, Z) \right] \\
   = & \mathbb{E}_{Z, W_T, N_{1:T}} \left[ R_{\rm test}(W_T+N_{1:T}, Z) - R_{\rm train}(W_T+N_{1:T}, Z) \right] \\
   & + \mathbb{E}_{Z,W_T,N_{1:T}} \left[ R_{\rm train}(W_T+N_{1:T}, Z) - R_{\rm train}(W_T, Z) \right] \\
   & - \mathbb{E}_{Z,W_T,N_{1:T}} \left[ R_{\rm test}(W_T+N_{1:T}, Z) - R_{\rm test}(W_T, Z) \right].
\end{aligned}
\end{equation}

We first discuss how to derive upper bounds for the second and third terms in Eq.~(\ref{decomposition}), which require additional assumptions. The first approach relies on the following assumption: $\mathbb{E}_{Z,N_{1:T}}\left[ R_{\rm test}(w_T+N_{1:T},Z) - R_{\rm test}(w_T,Z) \right] \geq 0$ holds for any realization $w_T$ of $W_T$. This assumption means that adding random noise to the final learned parameter $W_T$ does not, on average, reduce the transductive error on unlabeled data points. A similar assumption is also used by \cite{Foret2021,Wang2022} and in Corollary~\ref{cor2} to establish information-theoretic or PAC-Bayesian bounds. Under this assumption, the third term in Eq.~(\ref{decomposition}) can be omitted since
\begin{equation}
\begin{aligned}
   & \mathbb{E}_{Z,W_T,N_{1:T}} \left[ R_{\rm test}({W}_T+N_{1:T}, Z) - R_{\rm test}(W_T, Z) \right] \\
   = & \int_{w_T} \mathbb{E}_{Z,N_{1:T}} \left[ R_{\rm test}(w_T + N_{1:T},Z) - R_{\rm test}(w_T,Z) | W_T=w_T \right] \dif P_{W_T}(w_T) \geq 0.
\end{aligned}
\end{equation}
The second term $\mathbb{E}_{Z,W_T,N_{1:T}} \left[ R_{\rm train}(W_T+N_{1:T}, Z) - R_{\rm train}(W_T, Z) \right]$ is directly remained, which reflects the flatness of the training loss landscape around $W_T$. The second approach relies on the assumption that the spectral norm of the Hessian is bounded, which enables us to provide an upper bound for the trace of the Hessian. By applying Taylor's expansion on $R_{\rm test}$ and $R_{\rm train}$ respectively and neglecting high order terms, we can derive upper bounds for the second and third terms in Eq.~(\ref{decomposition}). We present this result in the following theorem. 
\begin{theorem}\label{bound_gap}
    Suppose that (\romannumeral1) the loss $\ell(w,s)$ is second order differential w.r.t. $w$ and (\romannumeral2) $\mathop{\rm sup}_{s} \left\Vert \frac{\partial^2 \ell(w,s)}{\partial w^2} \right\Vert \leq B_H$ that holds for all $w \in \mathcal{W}$ and $s \in \{s_i\}_{i=1}^n$. Then we have
    \begin{equation*}
    \begin{aligned}
        & \mathbb{E}_{Z,W_T,N_{1:T}} \left[ R_{\rm train}(W_T+N_{1:T}, Z) - R_{\rm train}(W_T, Z) \right] \\
        \leq & \mathbb{E}_{Z,W_T,N_{1:T}} \left[ R_{\rm test}(W_T+N_{1:T}, Z) - R_{\rm test}(W_T, Z) \right] \\
        & + \bigg(\sum_{t=1}^T \sigma^2_t \bigg) \sqrt{\frac{32 d^2 B^2_H C_{m,u} (I(W_T;Z)+\log (d)) (m+u)}{mu}} + \mathcal{O}\Bigg(\bigg(\sum_{t=1}^T \sigma^2_t\bigg)^2\Bigg).
    \end{aligned}
    \end{equation*}
\end{theorem}

In the proof of Theorem~\ref{bound_gap}, we establish a concentration inequality for the largest eigenvalue of the Hessian associated with the transductive generalization gap by the matrix martingale technique. Once this concentration inequality is obtained, various methods such as covering numbers \citep{Ju2022} or transductive Rademacher complexity \citep{Yaniv2007} can be applied to bound the second and the third terms in Eq.~(\ref{decomposition}). To align with the motivation of this paper, we derive an upper bound within the framework of information theory. Notice that $B_H$ represents the maximum sharpness that may occur across the entire loss landscape. Therefore, although the assumption of the second approach differs from that of the first, it conveys the same geometric meaning: a flatter optima implies a smaller transductive generalization gap. This interpretation is consistent with both the first approach and Theorem~\ref{thm4}. Finally, inspired by the work of \cite{Neu2021,Wang2022}, we derive the following upper bound for the first term in Eq.~(\ref{decomposition}).
\begin{theorem}\label{thm5}
    Under the assumption of Theorem~\ref{thm1}, we have
    \begin{equation}\label{ada}
    \begin{aligned}
        & \mathbb{E}_{Z, W_T, N_{1:T}} \left[ R_{\rm test}(W_T+N_{1:T}, Z) - R_{\rm train}(W_T+N_{1:T}, Z) \right] \\
        \leq & \frac{1}{2}\sqrt{\frac{dC_{m,u}(m+u)}{mu}\sum_{t=1}^T \log \left( \frac{1}{d\sigma_t^2} \mathbb{E}_{W_0,\ldots,W_{t-1},Z} \left[ \Big\Vert \Psi(W_0,\ldots,W_{t-1},Z) \Big\Vert^2_2 \right] + 1 \right)}.
    \end{aligned}
    \end{equation}
\end{theorem}

The right hand of Eq.~(\ref{ada}) is described by the norm square of the ``adaptive gradient'' on the training trajectory. The expectation is bounded if the gradients at each point on this training trajectory are bounded, a condition that can generally be satisfied in practice. The behavior of this expectation depends on the specific learning algorithms and training data we use. Although Theorem~\ref{thm5} does not provide a convergence guarantee towards a minimum, it allows us to estimate the generalization behavior of the model from the perspective of optimization trajectory. Analyzing the optimization property of adaptive optimization algorithms under the transductive learning setting goes beyond the scope of this paper, and we left it for future work. Compared with results derived from algorithm stability \citep{Cong2021} or complexity of hypothesis space \citep{Tang2023}, Theorem~\ref{thm5} does not include any Lipschitz or smoothness constants. Since the Lipschitz constant serves as the upper bound of the norm of gradients, our result can more finely depict the impact of optimization algorithms on generalization ability. Another advantage of our result is that it does not require the activation function to be smooth \citep{Cong2021} or H\"older smooth \citep{Tang2023}. Therefore, it can be applied to neural networks using ReLU as the activation function. Moreover, we also derive analogous results for Adam \citep{Kingma2014}, given its widespread use in practice. Since their formulations and reflected insights are similar to those of Theorem~\ref{thm5}, we place them in Appendix~\ref{proof5}.

\section{Applications}\label{appli}

\subsection{Semi-supervised Learning}\label{ssl_sec}

Due to the expensive cost of collecting high-quality labeled data, semi-supervised learning \citep{Shahshahani1994,Blum1998,Thorsten1999,Zhu2003} has attracted increasing attention, whose goal is to train a model using a small amount of labeled data and a large volume of unlabeled data. Establishing generalization guarantees for semi-supervised learning algorithms has been extensively studied \citep{Mey2023}, with theoretical results varying based on the problem setting and underlying assumptions. In this section, we adopt the random sampling setting of transductive learning as outlined in Section~\ref{con_pac}, assuming that $u=km,k\in \mathbb{N}_+$. Recall that under this setting, the full data points are defined as $S=(S_1,\ldots,S_{(k+1)m})$, where $S_i \triangleq (X_i,Y_i), i\in [(k+1)m]$. To transport the results derived under the random splitting setting to the random sampling setting, we follow the work of \cite{Vapnik1982} and assume that $S_i \sim P_S, i\in [(k+1)m]$, meaning that each data point is independently drawn from the same distribution. This assumption about the data distribution is slightly stronger than the partially exchangeable assumption but allows us to derive results more conveniently. And, the numerical values of them are easy to estimate. Let us start from the case where $u=m$. Denote by $Z=(Z_1,\ldots,Z_{2m})$ the sequence obtained by sampling without replacement from $[2m]$, and let $U\triangleq(U_1,\ldots,U_m) \sim \mathrm{Unif}(\{0,1\})^m$ represent the selector sequence. We have
\begin{equation}\label{ssl_eq}
\begin{aligned}
    \mathbb{E}_{W,S} \left[\mathcal{E}(W,S) \right] = & \mathbb{E}_{W,S} \Bigg[\frac{1}{m}\sum_{i=m+1}^{2m} \ell(W,S_i) -  \frac{1}{m}\sum_{i=1}^m \ell(W,S_i) \Bigg] \\
    = & \mathbb{E}_{W,S,Z} \Bigg[\frac{1}{m}\sum_{i=m+1}^{2m} \ell(W,S_{Z_i}) -  \frac{1}{m}\sum_{i=1}^m \ell(W,S_{Z_i}) \Bigg] \\
    = & \mathbb{E}_{W,S,\widetilde{Z},U} \Bigg[\frac{1}{m}\sum_{i=m+1}^{2m} \ell \big( W,S_{\mathscr{Z}_i(\widetilde{Z},U)} \big) -  \frac{1}{m}\sum_{i=1}^m \ell \big( W,S_{\mathscr{Z}_i(\widetilde{Z},U)} \big) \Bigg],
\end{aligned}
\end{equation}
where the second and third equality are obtained by the assumption $S_i \sim P_S, i \in [2m]$ and Proposition~\ref{prop1}, respectively. Notice that Eq.~(\ref{ssl_eq}) can be viewed as the expectation of the transductive generalization gap $\mathbb{E}_{W,\widetilde{Z},U}[\mathcal{E}(W,\mathscr{Z}(\widetilde{Z},U))]$ over the data points $(S_1,\ldots,S_{2m})$. In other words, the transductive generalization gap studied in Section~\ref{cmi_sec} can be expressed as  $\mathbb{E}_{W,\widetilde{Z},U|S=s}[\mathcal{E}(W,\mathscr{Z}(\widetilde{Z},U))]$. Based on the results derived in Section~\ref{cmi_sec}, we obtain the following information-theoretic bounds for semi-supervised learning algorithms.

\begin{proposition}\label{pro2}
    Suppose that $r(\hat{y},y) \in [0,B]$ holds for all $\hat{y} \in \widehat{\mathcal{Y}}$ and $y \in \mathcal{Y}$, where $B > 0$ is a constant. Also, suppose that $S_1, \ldots, S_{2m} \sim P_S$. Denote by $f_w(X) \in \mathbb{R}^K$ the prediction of the model and $F_{i} \triangleq (f_W(X_{\widetilde{Z}_{i,0}}), f_W(X_{\widetilde{Z}_{i,1}}))$ the sequence of predictions. Denote by $L_{i,:} \triangleq (\ell(W,S_{\widetilde{Z}_{i,0}}), \ell(W,S_{\widetilde{Z}_{i,1}}))$ the sequence of loss values and $\Delta_{i} \triangleq \ell(W,S_{\widetilde{Z}_{i,0}}) - \ell(W,S_{\widetilde{Z}_{i,1}})$ the difference of loss value. 
    We have
    \begin{align}
    & \left\vert \mathbb{E}_{W,S} \left[R_{\text{\rm test}}(W,S) - R_{\text{\rm train}}(W,S) \right] \right\vert \leq \frac{B}{m} \sum_{i=1}^m \mathbb{E}_{S,\widetilde{Z}} \sqrt{2I^{S,\widetilde{Z}}(F_i;U_i)}, \label{semi_bound1} \\
    & \left\vert \mathbb{E}_{W,S} \left[R_{\text{\rm test}}(W,S) - R_{\text{\rm train}}(W,S) \right] \right\vert \leq \frac{B}{m} \sum_{i=1}^m \mathbb{E}_{S,\widetilde{Z}} \sqrt{2I^{S,\widetilde{Z}}(L_i;U_i)} \label{semi_bound2}, \\
    & \left\vert \mathbb{E}_{W,S} \left[R_{\text{\rm test}}(W,S) - R_{\text{\rm train}}(W,S) \right] \right\vert \leq \frac{B}{m} \sum_{i=1}^m \mathbb{E}_{S,\widetilde{Z}} \sqrt{2I^{S,\widetilde{Z}}(\Delta_i;U_i)} \label{semi_bound3}.
\end{align}
\end{proposition}

Proposition~\ref{pro2} extends Corollary~\ref{cor1} to the random sampling setting. Notice that the mutual information terms in Proposition~\ref{pro2} are conditioned on two random variables $S$ and $\widetilde{Z}$, whereas in  Corollary~\ref{cor1}, they are conditioned only on $\widetilde{Z}$. The reason for this difference is that Corollary~\ref{cor1} is derived under the random splitting setting, where the training data is selected on a fixed set of data points. Thus, the only source of randomness comes from the sampling without replacement process represented by $\widetilde{Z}$. In contrast, for the random sampling setting, we first sample a sequence of data points from $P_S$ and then divide them into training and test sets according to $\widetilde{Z}$, which is obtained by sampling without replacement from $[2m]$. Therefore, the randomness of training data is depicted by both $(S_1,\ldots,S_{2m})$ and $\widetilde{Z}$. Additionally, we point out that the bounds converge as the number of test data points $u$ increases, given our assumption that $u=m$. Following the same procedure, we can obtain results similar to Eqs.~(\ref{semi_bound1}) and (\ref{semi_bound2}) for the common case where $u=km,k\in \mathbb{N}_+$. These results and detailed derivations are provided in Appendix~\ref{ssl_pro}.

\subsection{Transductive Graph Learning}\label{trc_graph}

Graph-structured data, composed of multiple objects and their relationships, plays a crucial role in various real-world applications, such as recommendation systems \citep{Wang2019neural,He2020light,Huang2021mix}, drug discovery \citep{Sun2020,Pietro2021}, and traffic flow forecasting \citep{Song2020,Li2021Spatial,Lan2022}. 
Graph learning tasks can be categorized into transductive tasks and inductive tasks. In this paper, we focus on transductive graph tasks, which can be further divided into node-level tasks and edge-level tasks. For transductive node-level tasks, some nodes are randomly selected from a fixed graph and their labels are revealed to the model. The learning goal is to predict the labels of the remaining nodes based on the revealed labels, features of all nodes, and the graph structure. For transductive edge-level tasks, sub-graphs (that is, subsets of full nodes along with their edges) are sampled from a fixed graph and provided to the model. The learning objective is to predict the connections between nodes that are not included in the sampled sub-graphs. Both types of transductive learning tasks fit within the random splitting setting of transductive learning, allowing us to apply the results derived in this work to analyze the generalization abilities of GNNs on these tasks. 

We now demonstrate the application of the derived results on the transductive node classification task. Assume that the model is a two-layer GCN \citep{Kipf2017semisupervised}. Let $\widetilde{\mathbf{A}} \in \mathbb{R}^{n\times n}$ be the normalized adjacent matrix with self-loops and $\mathbf{X} \in \mathbb{R}^{n \times d_0}$ be the node feature matrix. Let $\mathbf{W}_1 \in \mathbb{R}^{d_0\times d_1}$ and $\mathbf{W}_2 \in \mathbb{R}^{d_1\times |\mathcal{Y}|}$ be the learnable parameters, whose collection is denoted by $\mathbf{W} \triangleq \left[\text{vec}\left[ \mathbf{W}_1 \right],\text{vec}\left[ \mathbf{W}_2 \right]\right]$. Here, ${\rm vec}[\cdot]$ is the vectorization operator that reshapes a matrix into a column vector. The prediction of the model is $\widehat{\mathbf{Y}} = \text{Softmax}(\widetilde{\mathbf{A}}\text{ReLU}(\mathbf{H})\mathbf{W}_2) \in \mathbb{R}^{n\times |\mathcal{Y}|}$ with $\mathbf{H} \triangleq \widetilde{\mathbf{A}}\mathbf{X}\mathbf{W}_1$. Without loss of generality, we assume that the sequence sampled without replacement from $[n]$ is given by $Z=(1,\ldots,n)$. Let $\mathbf{Y} \in \{0, 1\}^{n \times |\mathcal{Y}|}$ be the label matrix where each row represents the one-hot vector of the corresponding node's label. Suppose that $\ell(\cdot)$ is the cross-entropy loss, then the transductive training error is $R_{\rm train}(\mathbf{W},Z) = - \frac{1}{m} \sum_{i=1}^m \sum_{j=1}^{|\mathcal{Y}|} \mathbf{Y}_{ij} \log \widehat{\mathbf{Y}}_{ij}$. The gradient of $R_{\rm train}(\mathbf{W},Z)$ is given by $g(\mathbf{W},Z) = \left[ \frac{\partial R_{\rm train}(\mathbf{W},Z)}{\partial \text{vec}\left[ \mathbf{W}_1 \right]}, \frac{\partial R_{\rm train}(\mathbf{W},Z)}{\partial \text{vec}\left[ \mathbf{W}_2 \right]} \right]^\top$ with
\begin{equation*}
\begin{aligned}
    & \frac{\partial R_{\rm train}(\mathbf{W},Z)}{\partial \text{vec}\left[ \mathbf{W}_1 \right]} = \frac{1}{m} \sum_{i=1}^m (\widehat{\mathbf{Y}}_{i,:} - \mathbf{Y}_{i,:}) \otimes \Bigg( \sum_{j=1}^n \widetilde{\mathbf{A}}_{ij} \text{ReLU}(\mathbf{H}_{j,:}) \Bigg), \\
    & \frac{\partial R_{\rm train}(\mathbf{W},Z)}{\partial \text{vec}\left[ \mathbf{W}_2 \right]} = \frac{1}{m} \sum_{i=1}^m \sum_{j=1}^n \widetilde{\mathbf{A}}_{ij} \Bigg( \text{ReLU}' \bigg(\sum_{k=1}^n \mathbf{H}_{k,:} \bigg) \odot \bigg( (\widehat{\mathbf{Y}}_{i,:} - \mathbf{Y}_{i,:})\mathbf{W}^\top_2 \bigg) \Bigg) \otimes \mathbf{H}_{j,:},
\end{aligned}
\end{equation*}
where $\text{ReLU}'(\cdot)$ is the derivative of the ReLU function. Here we assume that the transductive training error is computed over all labeled nodes, which is a common setting in practice \citep{Kipf2017semisupervised,Gasteiger2018combining,Chien2021adaptive}. By substituting $g(\mathbf{W},Z)$ into Theorem \ref{thm5}, we can derive an upper bound for GCNs trained with adaptive optimization algorithms. Notably, the influence of GNN architectures and graph-structured data on model generalization is captured by the norm of the gradients. Following the analysis of \cite{Cong2021,Tang2023}, one can derive fine-grained upper bounds for other GNN models and gain insights into their generalization behavior.

\begin{figure}[!t]
\centering
\subfigure[MNIST with $k=1$]{
\includegraphics[width=0.35\textwidth]{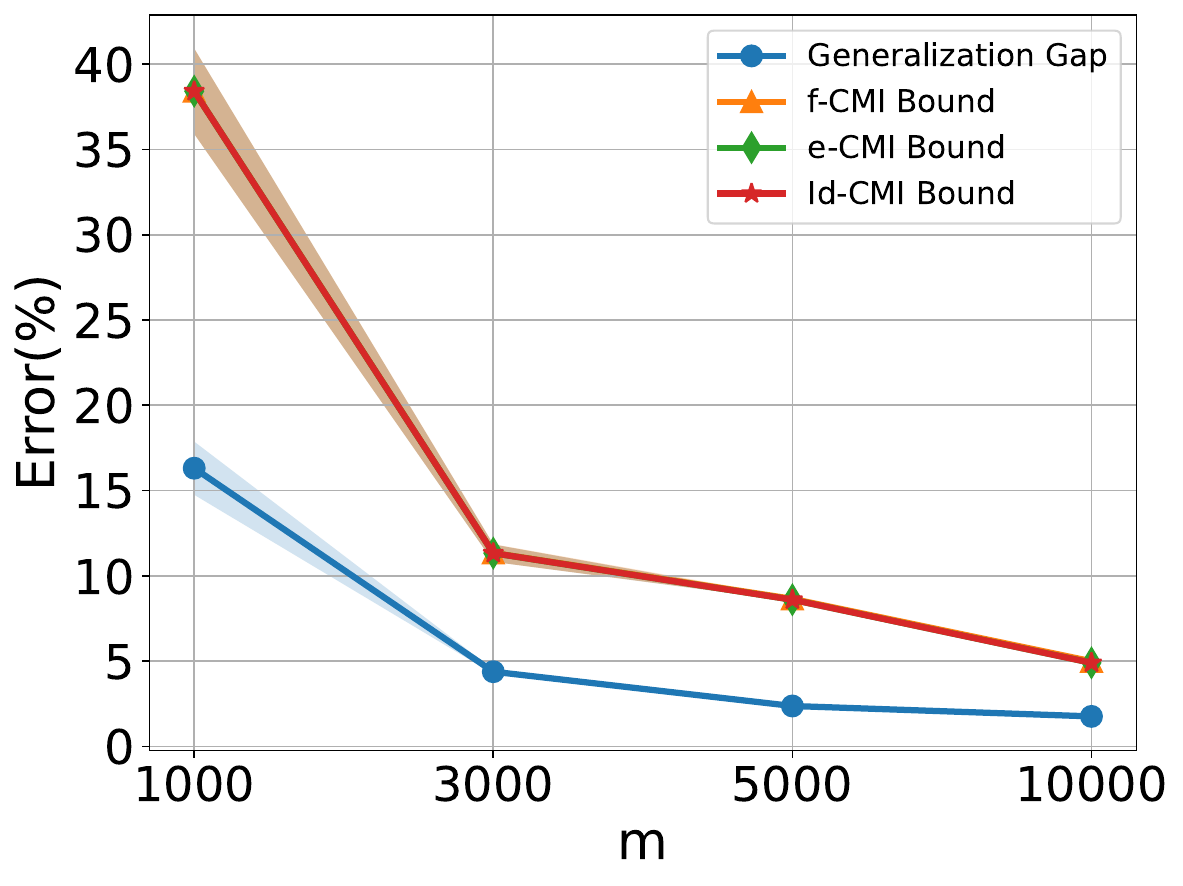}}
\hspace{0.2mm}
\subfigure[MNIST with $k=2$]{\includegraphics[width=0.35\textwidth]{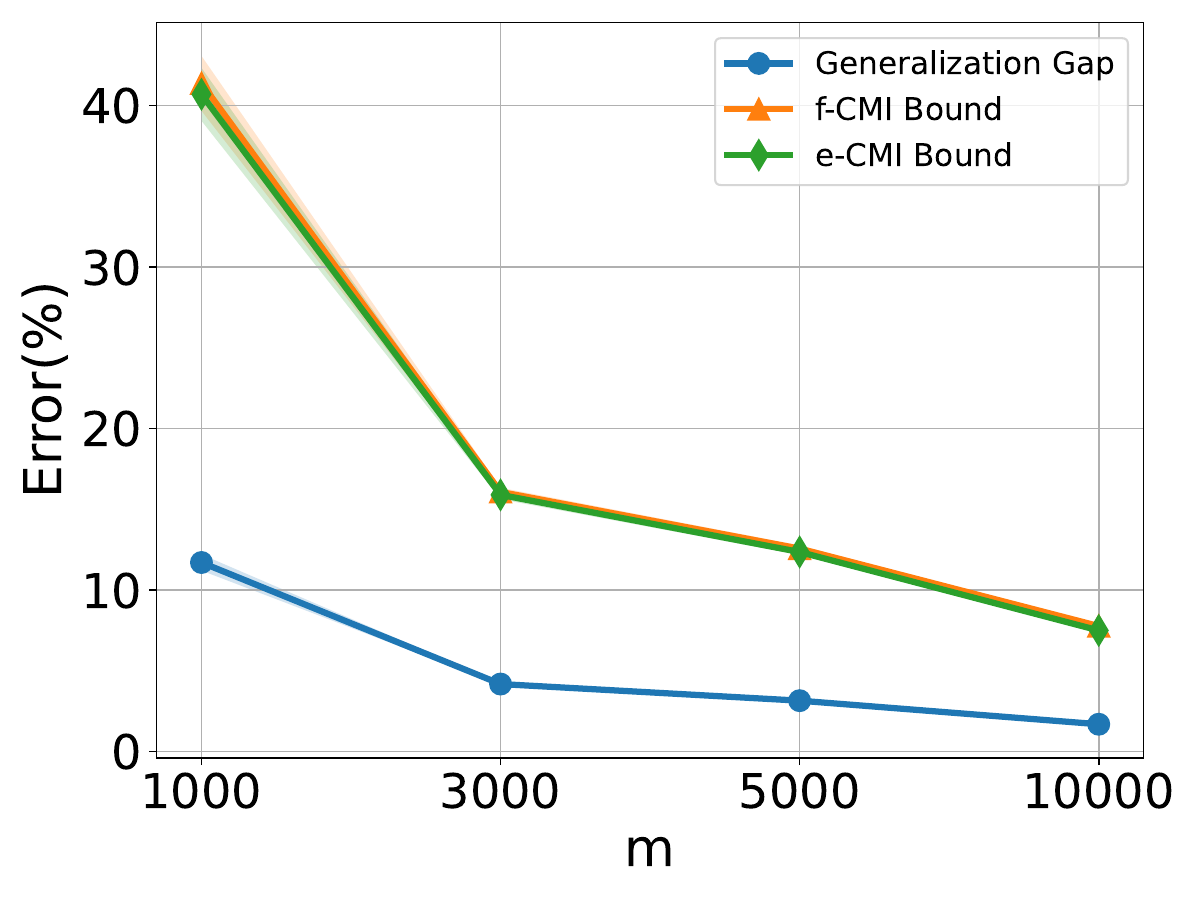}}\\
\subfigure[CIFAR-$10$ with $k=1$]{
\includegraphics[width=0.35\textwidth]{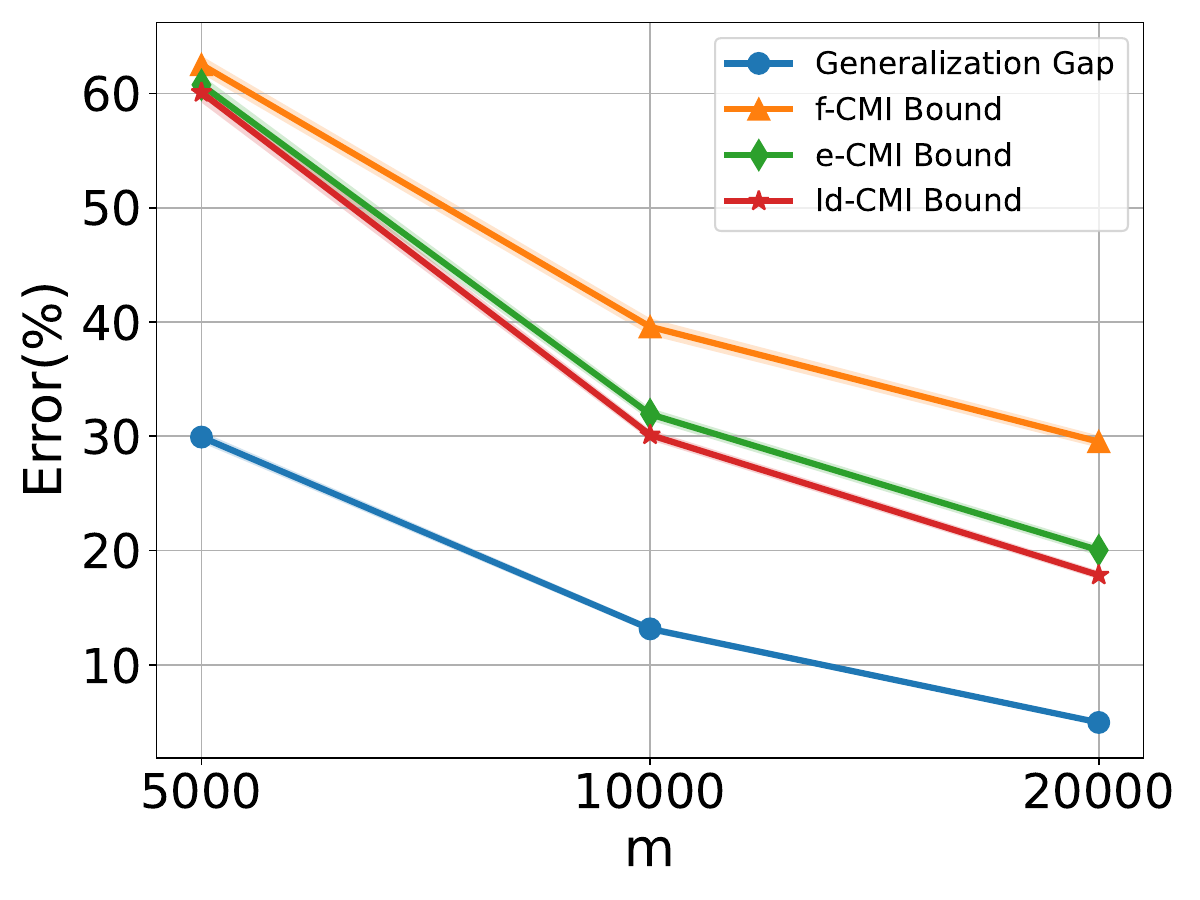}\label{1c}}
\hspace{0.2mm}
\subfigure[CIFAR-$10$ with $k=2$]{\includegraphics[width=0.35\textwidth]{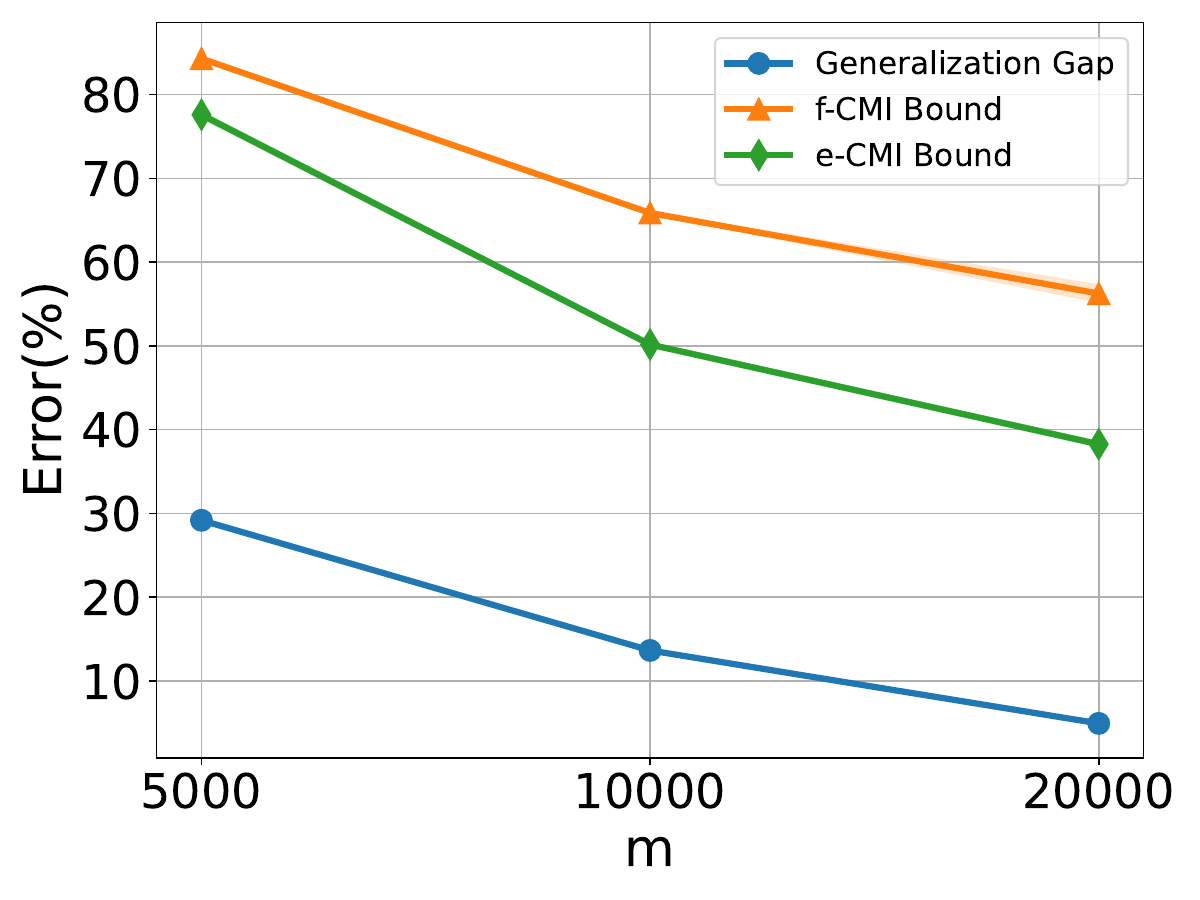}\label{1d}}
\caption{Estimations of the transductive generalization gap and the derived bounds on MNIST and CIFAR-$10$ with different values of $m$ and $k$.} 
\label{ssl}
\vskip -0.1in
\end{figure}

\begin{figure}[t]
\centering
\subfigure{
\includegraphics[width=0.3\textwidth]{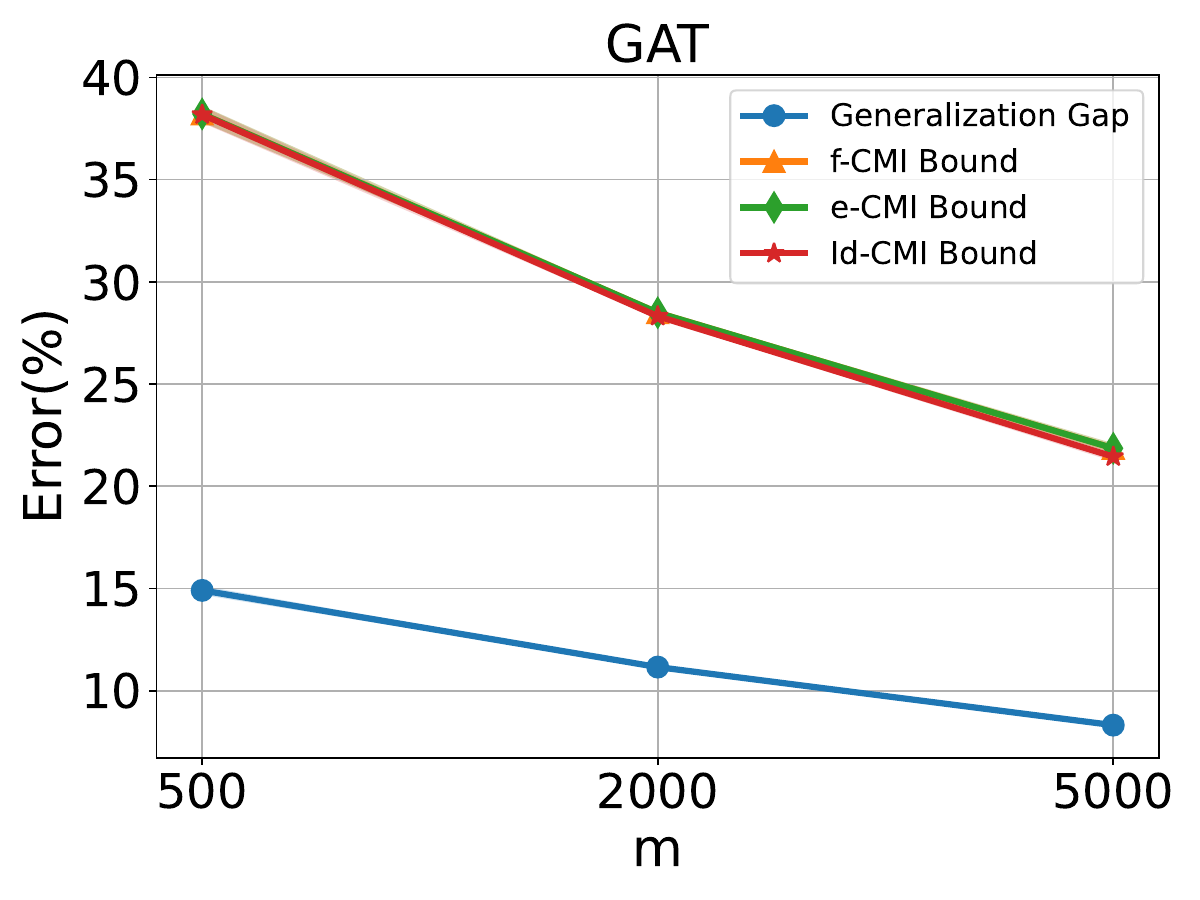}}
\hspace{-0.3mm}
\subfigure{\includegraphics[width=0.3\textwidth]{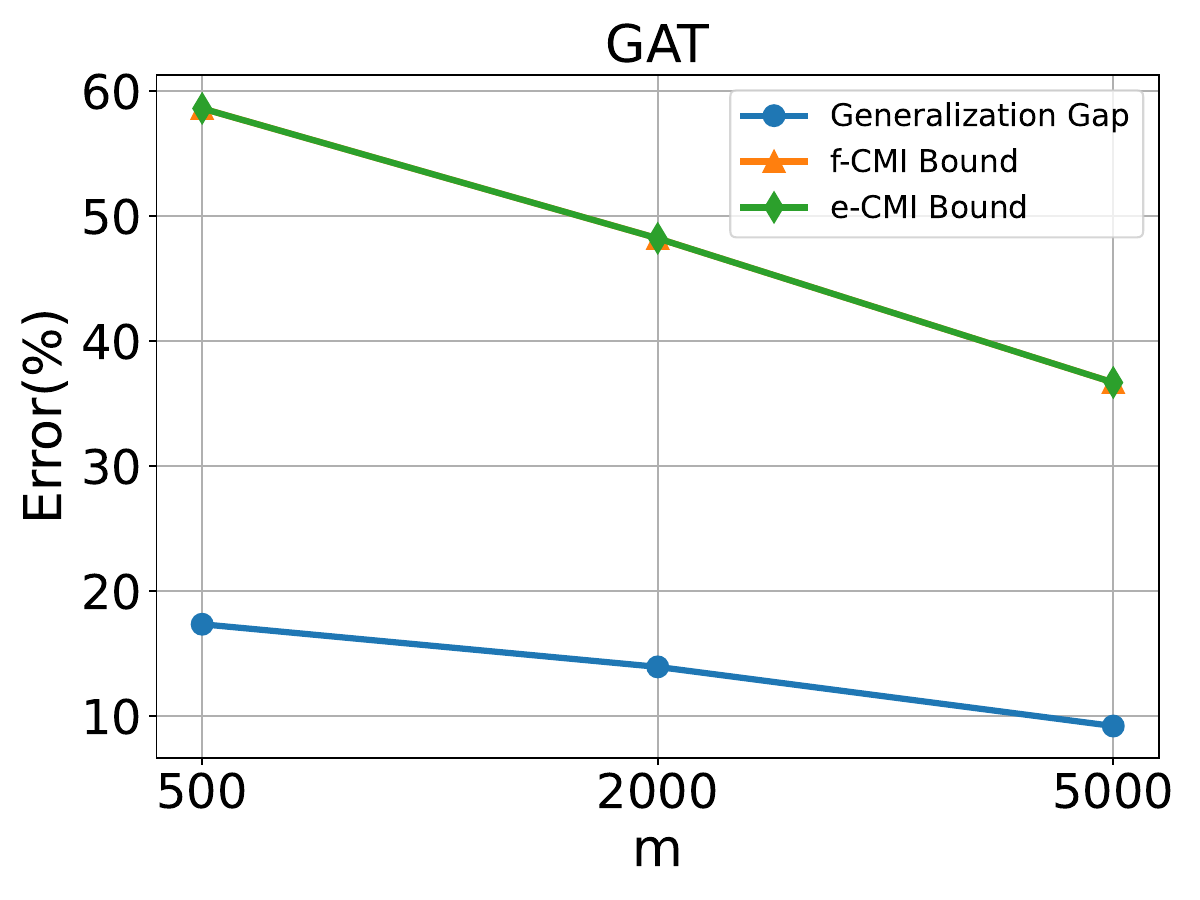}}
\hspace{-0.3mm}
\subfigure{\includegraphics[width=0.3\textwidth]{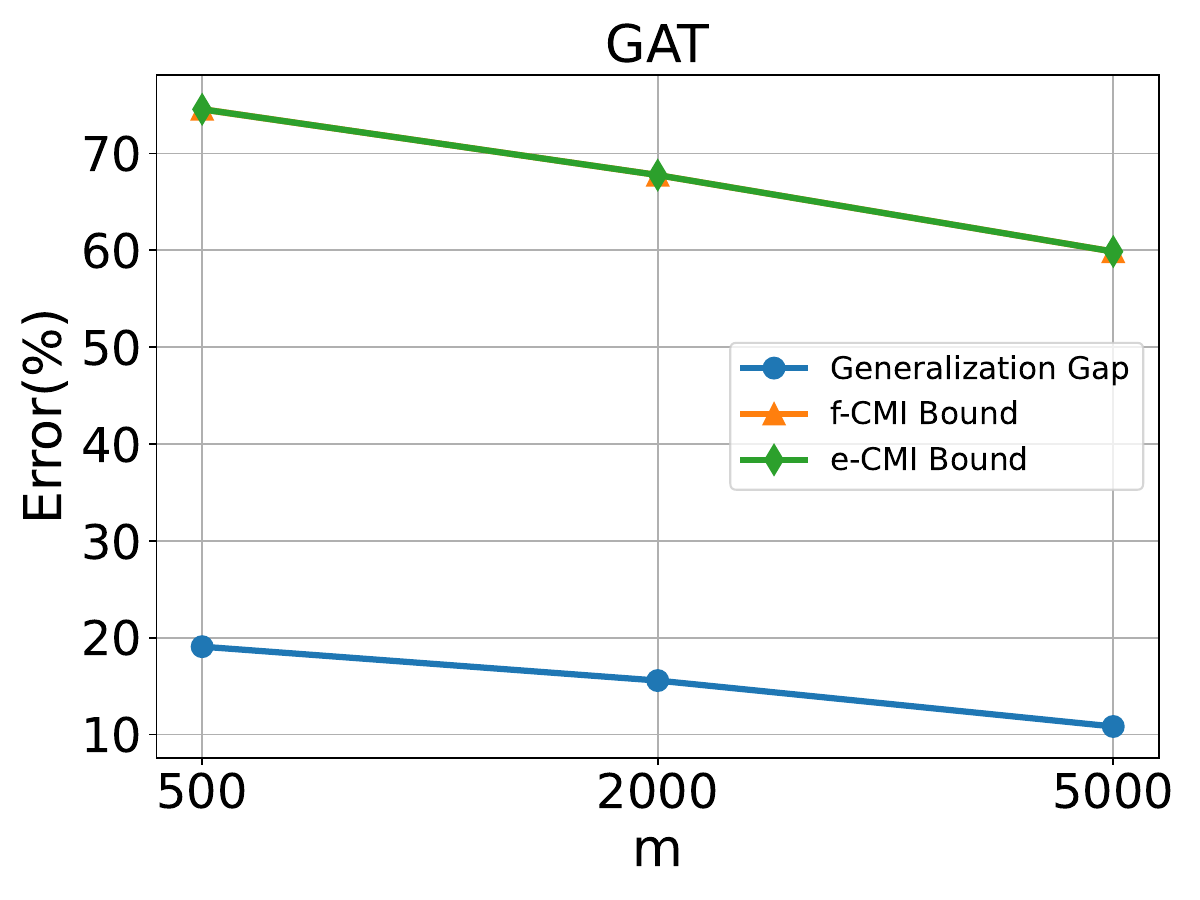}}\\
\subfigure{
\includegraphics[width=0.3\textwidth]{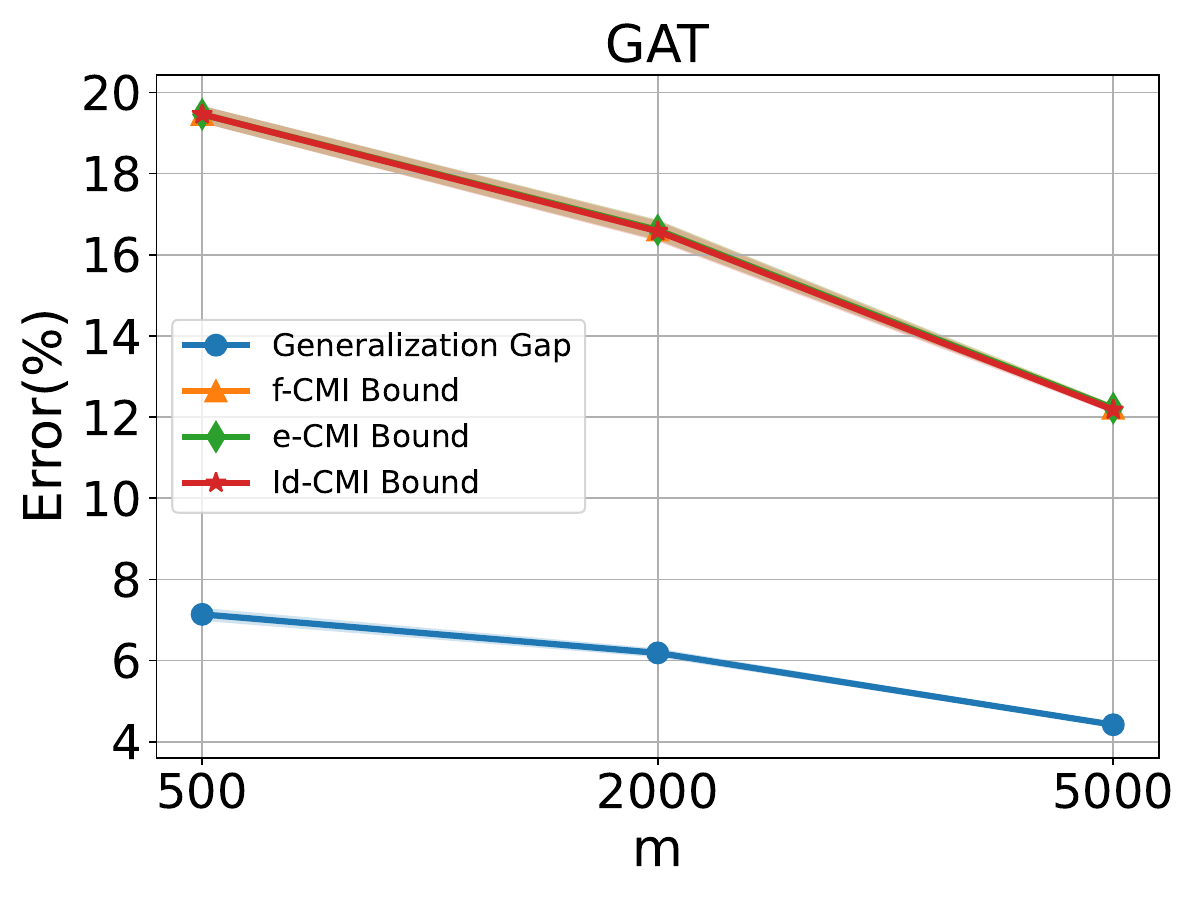}}
\hspace{-0.3mm}
\subfigure{\includegraphics[width=0.3\textwidth]{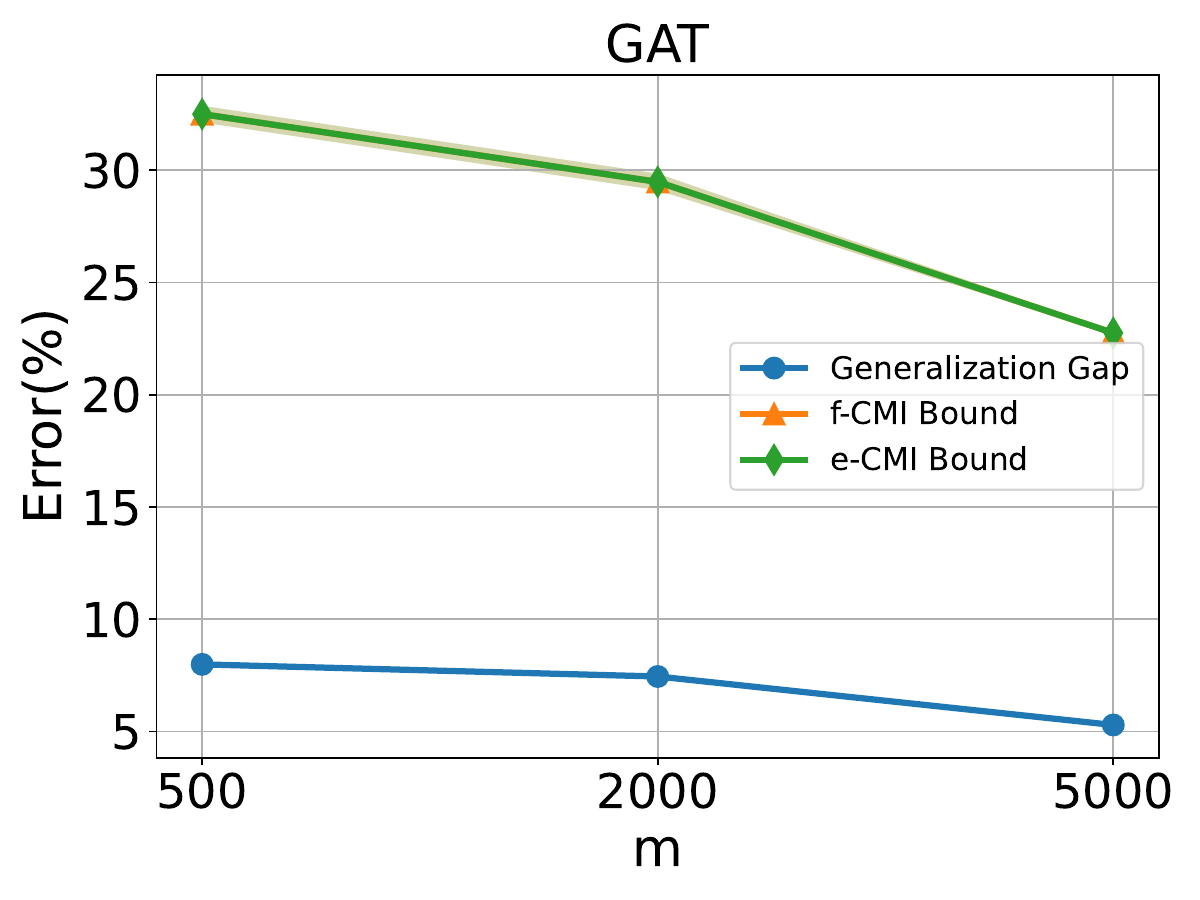}}
\hspace{-0.3mm}
\subfigure{\includegraphics[width=0.3\textwidth]{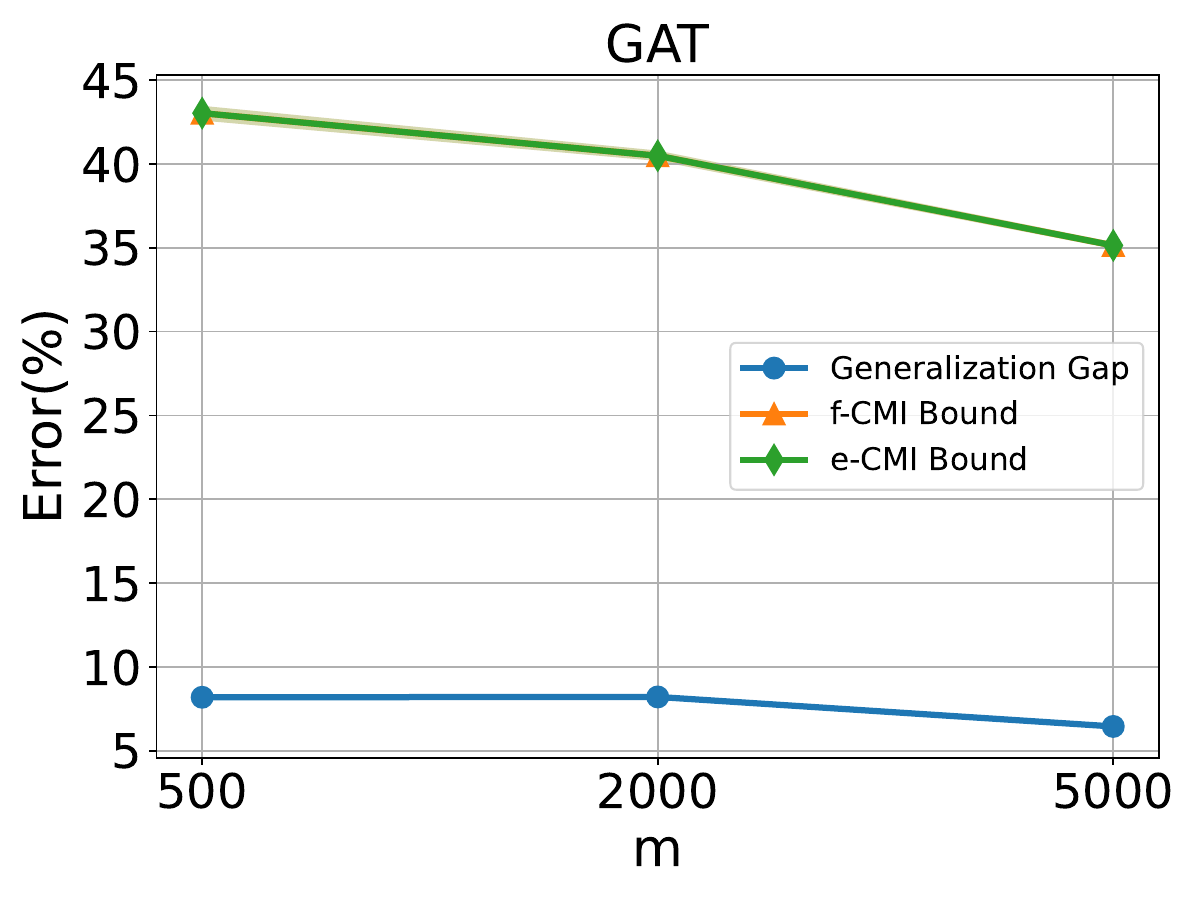}}\\
\subfigure{\includegraphics[width=0.3\textwidth]{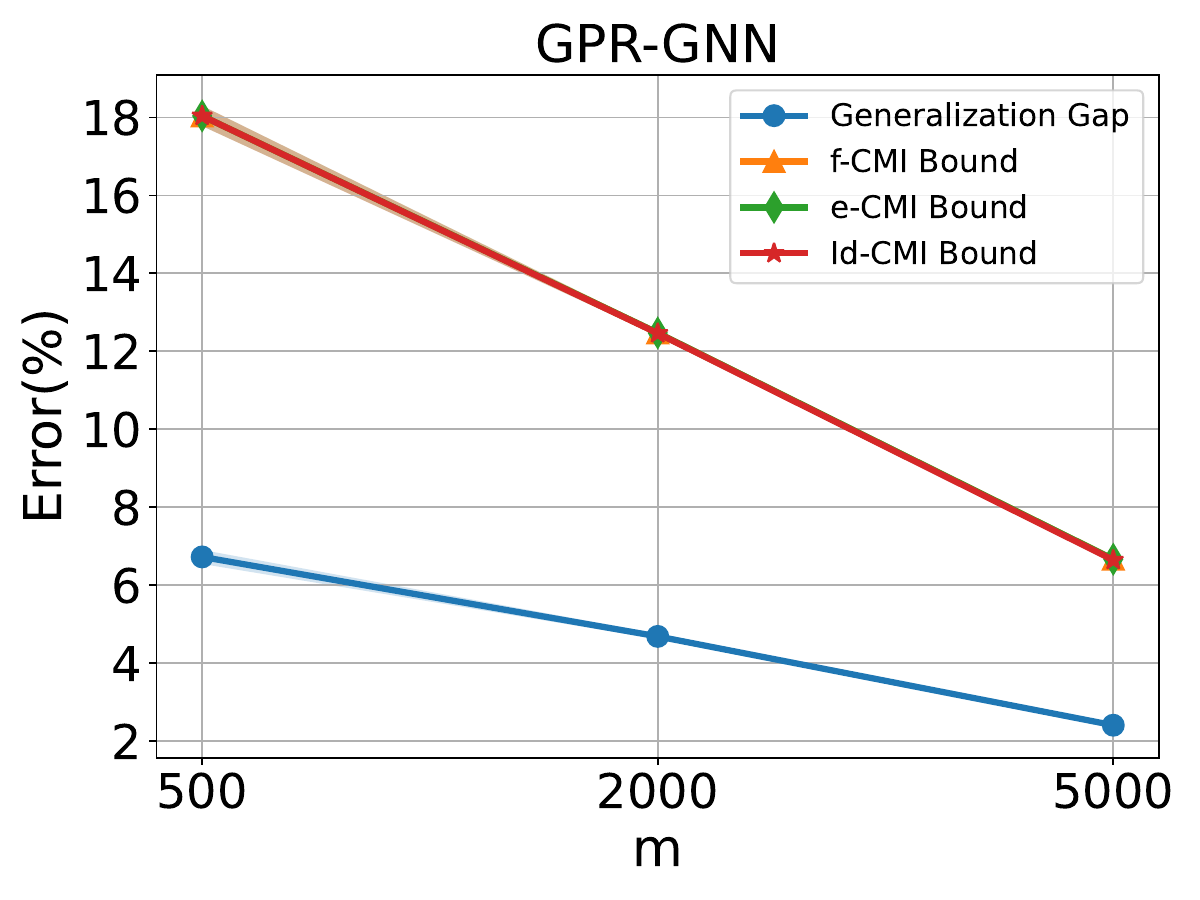}}
\hspace{-0.3mm}
\subfigure{\includegraphics[width=0.3\textwidth]{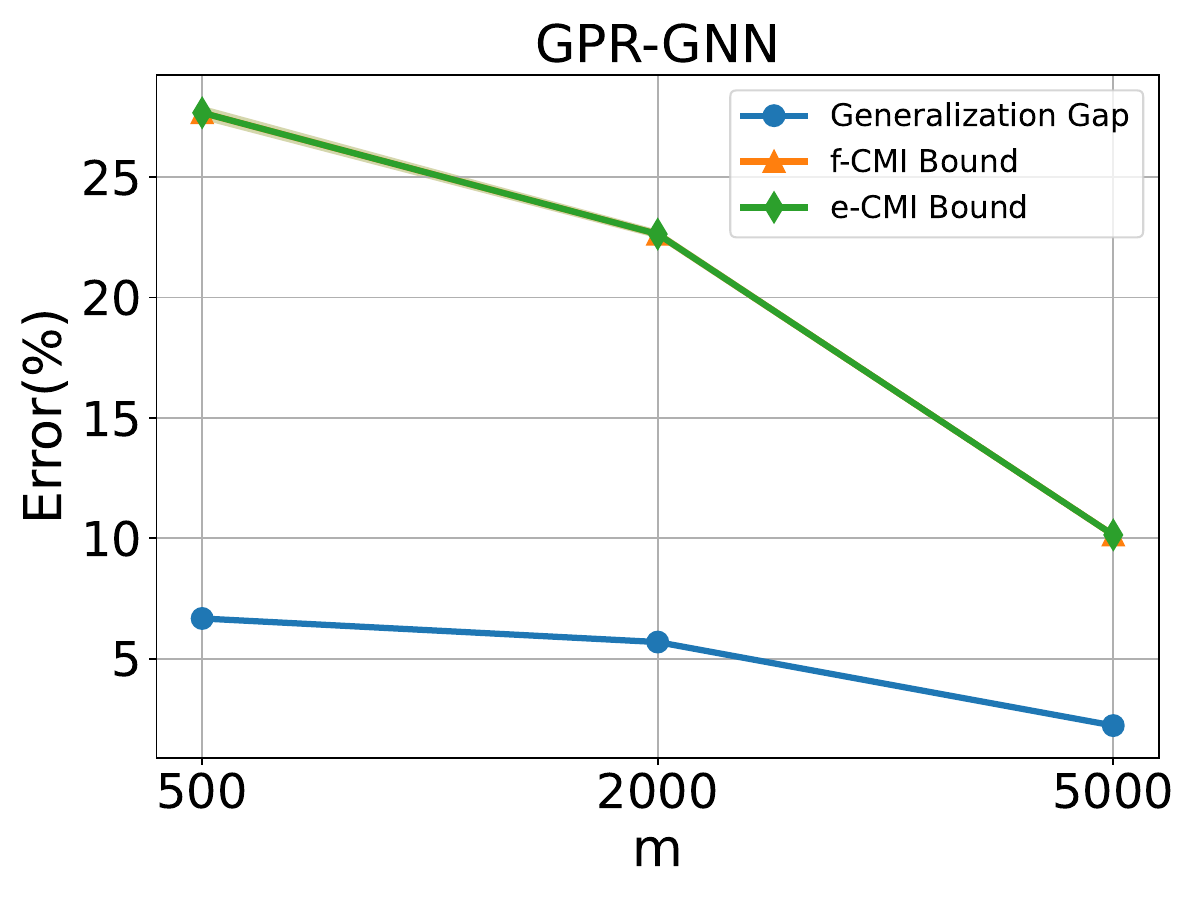}}
\hspace{-0.3mm}
\subfigure{\includegraphics[width=0.3\textwidth]{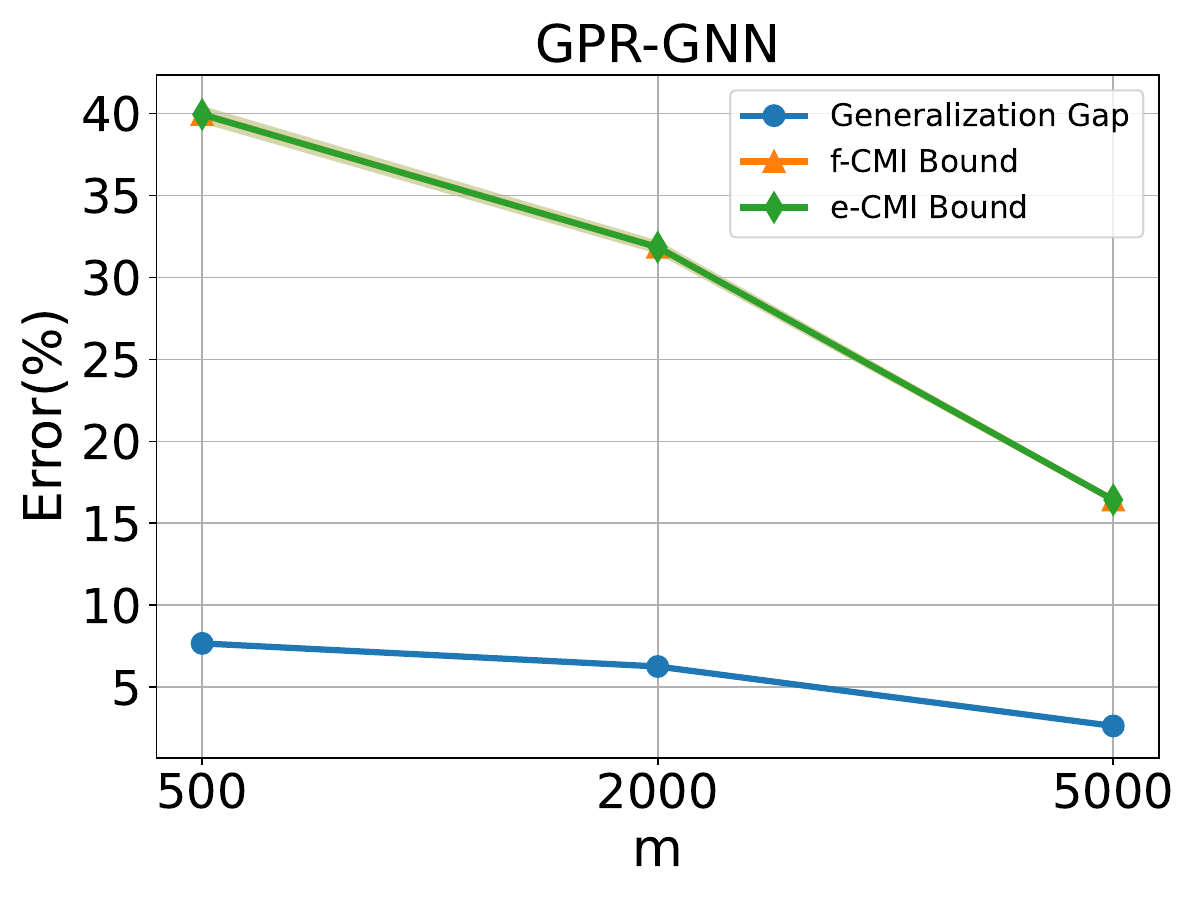}}\\
\subfigure{
\includegraphics[width=0.3\textwidth]{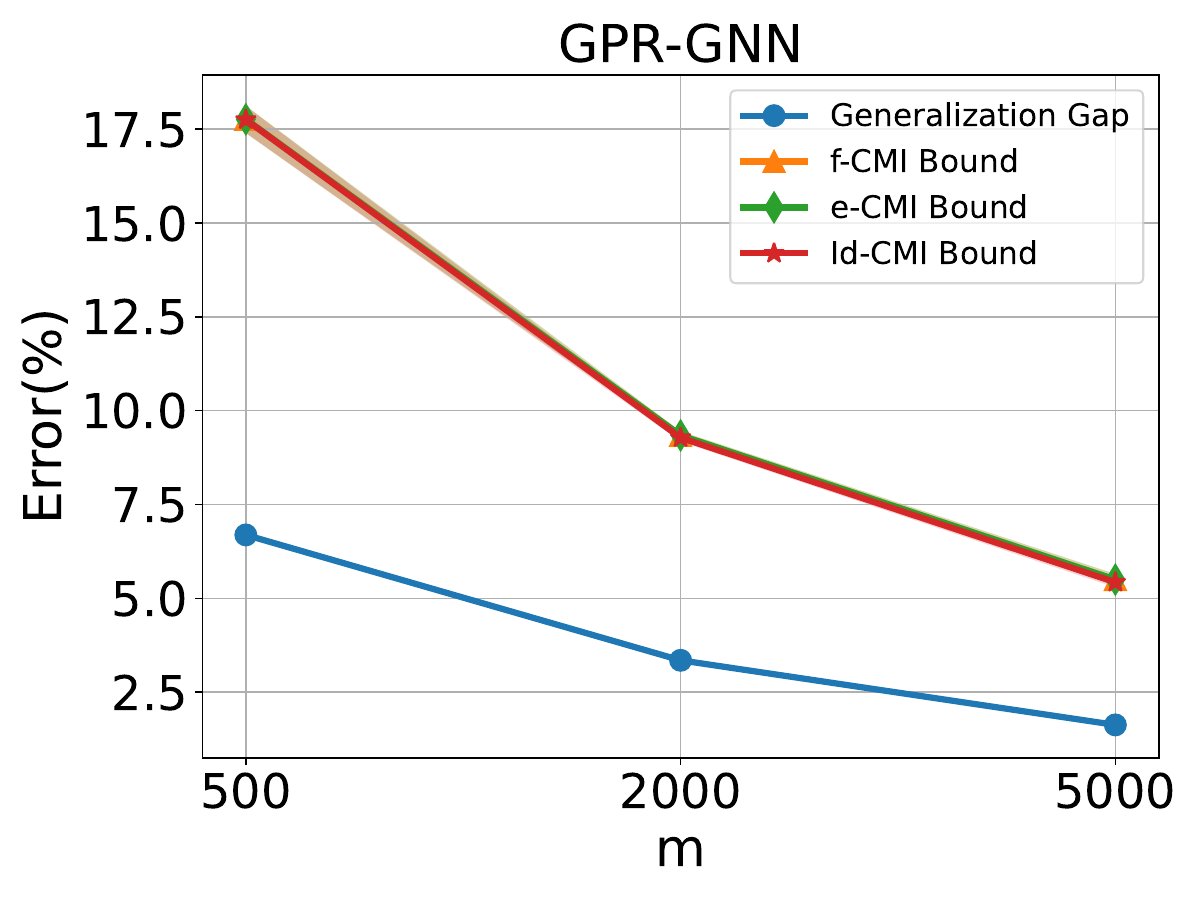}}
\hspace{-0.3mm}
\subfigure{\includegraphics[width=0.3\textwidth]{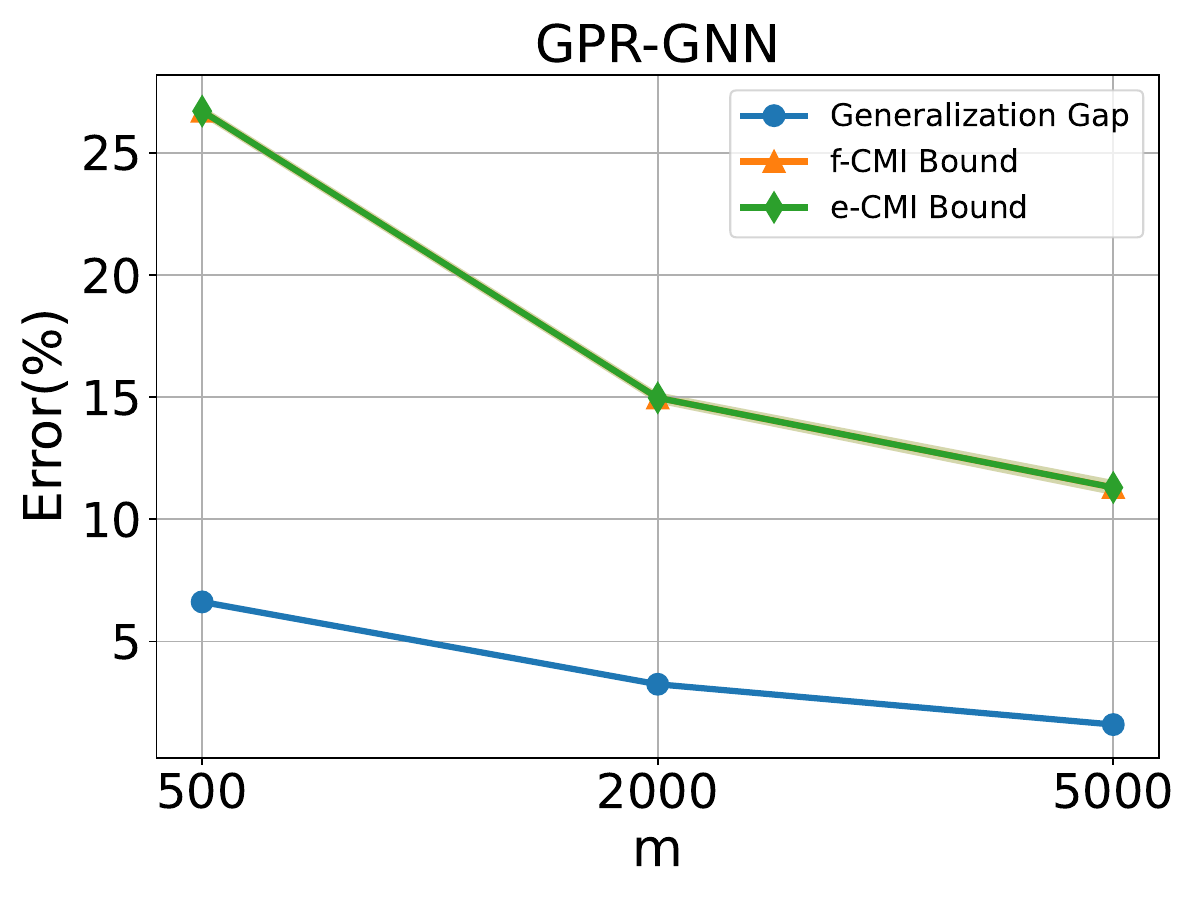}}
\hspace{-0.3mm}
\subfigure{\includegraphics[width=0.3\textwidth]{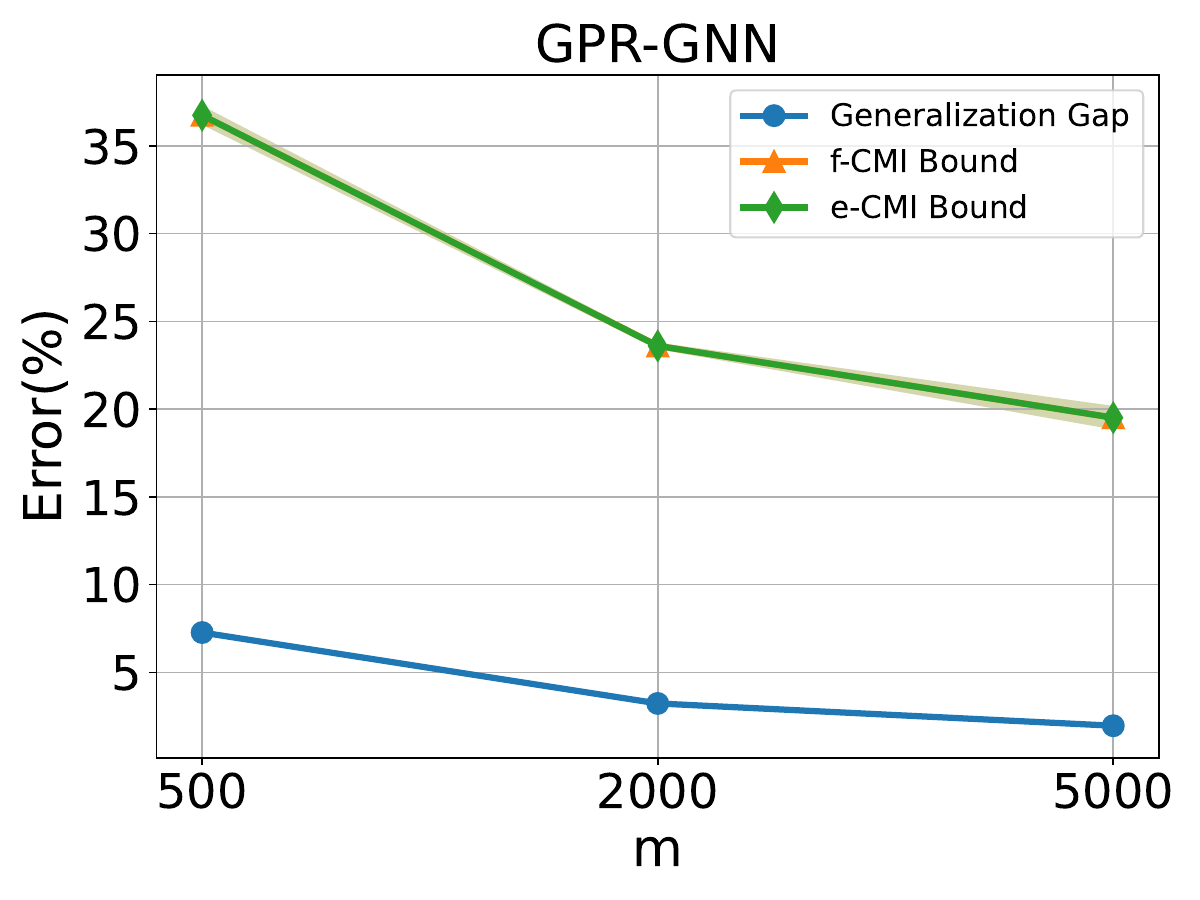}}\\
\caption{Estimations of the transductive generalization gap and the derived bounds on cSBMs with GAT and GPR-GNN. The first (second) and third (fourth) rows correspond to $\phi=-0.5$ ($\phi=0.5$). The left, middle, and right figures in each row correspond to $k=1$, $k=2$ and $k=3$.}
\label{graph1}
\vskip -0.2in
\end{figure}

\begin{figure}[!t]
\centering
\subfigure[GAT]{
\includegraphics[width=0.45\textwidth]{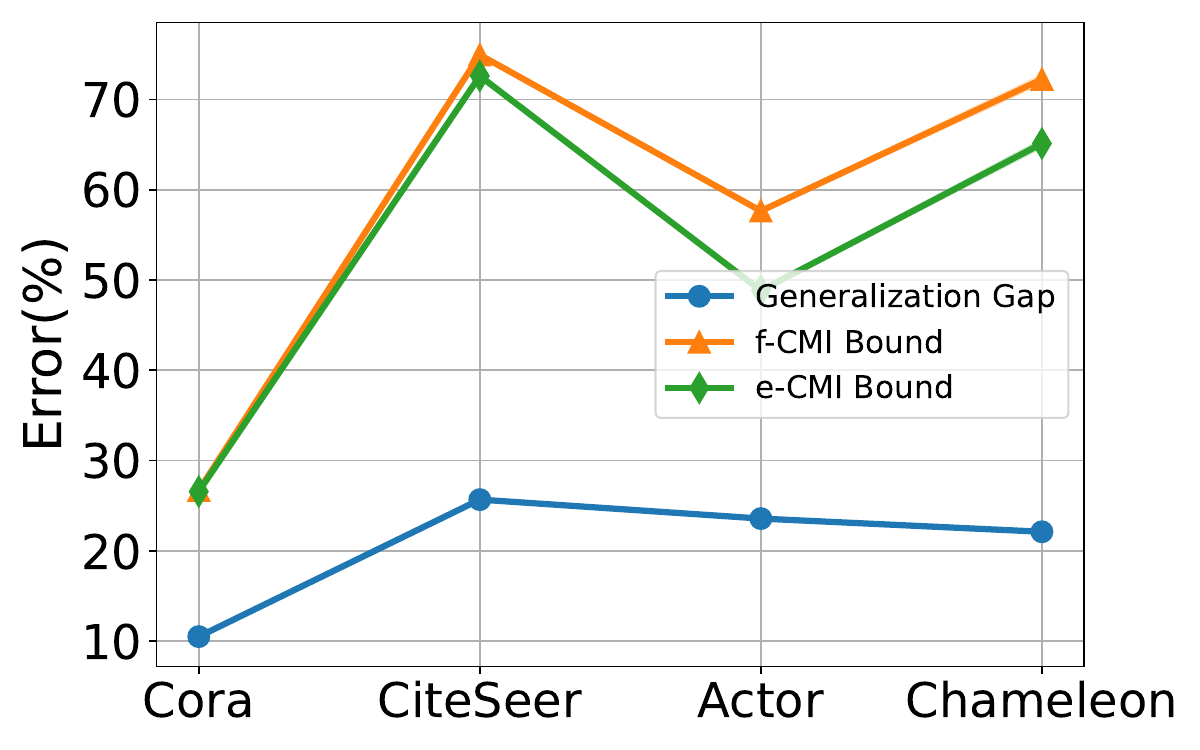}}
\hspace{0.2mm}
\subfigure[GPR-GNN]{\includegraphics[width=0.45\textwidth]{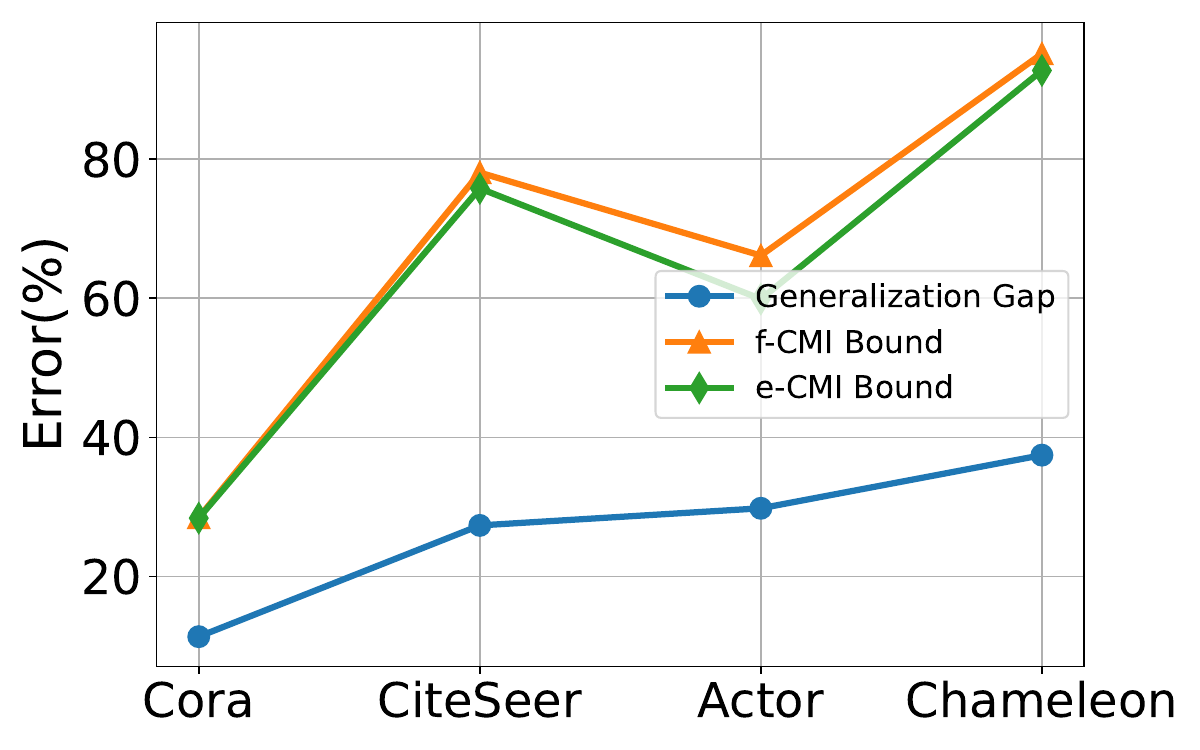}}
\caption{Estimations of the transductive generalization gap and the derived bounds on real-world datasets with GAT and GPR-GNN.}
\label{graph2}
\vskip -0.15in
\end{figure}

\section{Experiments}\label{experi}

\subsection{Experimental Setup}

For semi-supervised learning, we select image classification on the MNIST and CIFAR-$10$ datasets as learning tasks. Following the work of \cite{Guo2020}, the loss for unlabeled images is defined as the mean square error between the prediction of augmented images and their vanilla (non-augmented) counterparts by the model, which is known as a consistency regularization in the field of semi-supervised learning. The loss for labeled images is the cross-entropy loss. Following the work of \cite{Harutyunyan2021,Guo2020}, we utilize a four-layer CNN and a Wide \text{ResNet}-$28$-$10$ \citep{ZagoruykoK2016} as the model for MNIST and CIFAR-$10$, respectively. For both datasets, we train the model for $1000$ iterations using the Adam optimizer. The learning rate is set to $0.001$, and each mini-batch contains $128$ images. The objective function of training is the summation of losses on labeled and unlabeled losses. For evaluation, we use the zero-one loss as the criterion. Following the work of \cite{Harutyunyan2021}, we make the training process deterministic by fixing the sequence of mini-batch and the initialization of model parameters through random seeds. For transductive graph learning, we select semi-supervised node classification on both synthetic and real-world datasets as learning tasks. Specifically, we use cSBMs \citep{Deshpande2018} as synthetic data and Cora, CiteSeer \citep{Sen2008,Yang2016}, Actor, and Chameleon as real-world data. For each dataset, we adopt GAT \citep{Petar2018} and GPR-GNN \citep{Chien2021adaptive} as the models, since they are regarded as representative models of spatial and spectral GNNs, respectively. In experiments on both synthetic and real-world datasets, we train the model for $300$ iterations using the Adam optimizer with a learning rate of $0.01$ without weight decay. Following the setting of \cite{Kipf2017semisupervised,Gasteiger2018combining,Chien2021adaptive}, the GNN model passes over all labeled nodes during each iteration. Additional experimental details, including methods for estimating the expected generalization gap and the derived bounds, are provided in Appendix~\ref{exp_detail}.

\subsection{Experimental Results}

Figure~\ref{ssl} illustrates the results of semi-supervised learning algorithms on the MNIST and CIFAR-$10$ datasets, where $m$ represents the number of labeled images, and $k$ denotes the ratio of unlabeled to labeled images. The value of $f$-CMI, e-CMI, and Id-CMI bounds are calculated using Eqs.~(\ref{semi_bound1}), (\ref{semi_bound2}) and (\ref{semi_bound3}), respectively. For cases where $k \geq 2$, we report only the $f$-CMI and e-CMI bound values. It can be observed that the numerical values of our established bounds are non-vacuous, with the discrepancy between the estimated value and the generalization gap decreasing as $m$ increases. Conversely, this discrepancy tends to increase when $k$ grows larger. This phenomenon occurs because a higher $k$ value results in greater estimation error for the conditional mutual information, as discussed in Section~\ref{cmi_sec}. Additionally, it is observed that the e-CMI bound does not exceed the $f$-CMI bound, and the Id-CMI bound does not surpass the e-CMI bound for $k=2$. These observations align with the findings of \cite{Hellstrom2022,Wang2023}, and they remain valid in the transductive learning setting. The results of transductive graph learning are presented in Figures~\ref{graph1} and \ref{graph2}. Here, the $f$-CMI, e-CMI and Id-CMI bounds are computed according to Eqs.~(\ref{fmi}), (\ref{ecmi}), and (\ref{idcmi}). Generally, the trends observed are consistent with those seen in semi-supervised learning. Notably, in Figure~\ref{graph1}, the estimated values of the $f$-CMI, e-CMI, and Id-CMI bounds are nearly identical, which could be due to the small magnitude of the generalization gap. In contrast, as shown in Figures~\ref{1c} and \ref{1d}, when the generalized gap is large, the differences among these information-theoretic bounds become more apparent.

\section{Conclusion}\label{conclu}
In this work, we investigate the generalization ability of transductive learning algorithms. We develop a variety of generalization upper bounds under both the random sampling setting and the random splitting setting within the framework of information theory and PAC-Bayes. To address the dependency between training and test data points due to sampling without replacement, our key techniques involve employing martingale methods and introducing the concept of transductive supersamples. We anticipate that the findings and methodologies presented in this work will shed light on understanding and analyzing the generalization performance of transductive learning algorithms. Promising avenues for future research include extending our theoretical results to additional transductive learning scenarios and devising novel information measures tailored to the transductive learning setting. It is worth noting that while our study emphasizes establishing generalization upper bounds for transductive learning algorithms, further efforts to explore lower bounds remain a worthwhile area of investigation.

\section*{Acknowledgements}
We gratefully appreciate the editor and anonymous referees for their valuable and constructive comments. This research was supported by the National Natural Science Foundation of China (No.62476277), the National Key Research and Development Program of China (No.2024YFE0203200), the CCF-ALIMAMA TECH Kangaroo Fund (No.CCF-ALIMAMA OF 2024008), and the Huawei-Renmin University joint program on Information Retrieval. We also acknowledge the support provided by the fund for building worldclass universities (disciplines) of Renmin University of China and by the funds from Beijing Key Laboratory of Big Data Management and Analysis Methods, Gaoling School of Artificial Intelligence, Renmin University of China, from Engineering Research Center of Next-Generation Intelligent Search and Recommendation, Ministry of Education, from Intelligent Social Governance Interdisciplinary Platform, Major Innovation \& Planning Interdisciplinary Platform for the ``DoubleFirst Class” Initiative, Renmin University of China, from Public Policy and Decision-making Research Lab of Renmin University of China, and from Public Computing Cloud, Renmin University of China.

\newpage
\appendix

\section{Lemma}\label{not_lemma}

\begin{lemma}[\citealp{Polyanskiy2022},
Theorem~4.6]\label{lemma}
    Let $P$ and $Q$ be two probability measures on the same measurable space $(\mathcal{X},\mathcal{A})$, where $\mathcal{A}$ is a $\sigma$-algebra on $\mathcal{X}$. Denote by $\mathcal{F} \triangleq \left\{ f: \mathcal{X} \to \mathbb{R} \right \}$ the family of bounded measurable function on $\mathcal{X}$. Then we have
    \begin{equation}
        \mathrm{D_{KL}}(P||Q) = \mathop{\rm sup}_{f \in \mathcal{F}} \left\{\mathbb{E}_P \left[ f(X) \right] - \log \mathbb{E}_Q \left[ \exp\left\{ f(X) \right\} \right] \right\}.
    \end{equation}
    Denote by $\mathcal{P}(\mathcal{X})$ the set that contains all probability distributions on $(\mathcal{X}, \mathcal{A})$. For any $f \in \mathcal{F}$ we have
    \begin{equation}\label{dual_dv}
        \log \left(\mathbb{E}_{Q} [\exp (f(X))]\right) = \mathop{\rm sup}_{P \in \mathcal{P}(\mathcal{X})} \left\{ \mathbb{E}_{P} [f(X)] - \mathrm{D_{KL}}(P||Q) \right\}.
    \end{equation}
\end{lemma}

\section{Proof of Theorem~\ref{thm1}}\label{pf1}
\noindent
{\bf Proof}. We first establish an upper bound for the moment-generating function $\mathbb{E}_{Z} \left[ e^{\lambda \mathcal{E}(w, {Z})} \right]$ by the martingale technique for any $w \in \mathcal{W}$. Without loss of generality, we assume that $u \geq m$. Inspired by the work of \cite{Yaniv2006,Cortes2008,Yaniv2007}, we construct the following Doob's martingale difference sequences
\begin{equation*}
   \xi_i \triangleq \mathbb{E}[\mathcal{E}(w, {Z})|Z_1,\ldots,Z_i] - \mathbb{E}[\mathcal{E}(w, {Z})|Z_1,\ldots,Z_{i-1}], i \in [n].
\end{equation*}
One can verify that $\mathcal{E}(w, {Z}) - \mathbb{E}[\mathcal{E}(w, {Z})] = \sum_{i=1}^n \xi_i$ and $\mathbb{E}[\xi_i|Z_1,\ldots,Z_{i-1}]=0$ holds. Notice that $\xi_i$ is a function of $Z_1, \ldots, Z_i$. For $i\in [n]$, define
\begin{equation*}
\begin{aligned}
   \xi^{\rm inf}_i & \triangleq \mathop{\rm inf}_z \mathbb{E}[\mathcal{E}(w, {Z})|Z_1,\ldots,Z_{i-1},Z_i=z] - \mathbb{E}[\mathcal{E}(w, {Z})|Z_1,\ldots,Z_{i-1}], \\
   \xi^{\rm sup}_i & \triangleq \mathop{\rm sup}_z \mathbb{E}[\mathcal{E}(w, {Z})|Z_1,\ldots,Z_{i-1},Z_i=z] - \mathbb{E}[\mathcal{E}(w, {Z})|Z_1,\ldots,Z_{i-1}], \\
\end{aligned}
\end{equation*}
we have $\xi^{\rm inf}_i \leq \xi_i \leq \xi^{\rm sup}_i$. One can find that for $i\in [n]$,
\begin{equation}\label{martingale}
\begin{aligned}
   & \xi^{\rm sup}_i - \xi^{\rm inf}_i\\
   = & \mathop{\rm sup}_z \mathbb{E}[\mathcal{E}(w, {Z})|Z_1,\ldots,Z_{i-1},Z_i=z] - \mathop{\rm inf}_z \mathbb{E}[\mathcal{E}(w, {Z})|Z_1,\ldots,Z_{i-1},Z_i=z] \\
   = & \mathop{\rm sup}_{z,\tilde{z}} \Big\{ \mathbb{E}[\mathcal{E}(w, {Z})|Z_1,\ldots,Z_{i-1},Z_i=z] - \mathbb{E}[\mathcal{E}(w, {Z})|Z_1,\ldots,Z_{i-1},Z_i=\tilde{z}] \Big\}.
\end{aligned}
\end{equation}
Notice that once $Z_1,\ldots,Z_m$ are given, the values of $Z_{m+1},\ldots,Z_{n}$ do not affect the value of $\mathcal{E}(w,Z)$. Thus, $\xi^{\rm sup}_i - \xi^{\rm inf}_i = 0$ holds for $i = m+1, \ldots, n$. Now we discuss the case that $i\in [m]$. To proceed with the proof, we first illustrate the meaning of Eq.~(\ref{martingale}) by considering a fixed realization $z_{j}$ of $Z_{j}$ with $j\in [i-1]$. Denote by $z_i$ and $\tilde{z}_i$ the realizations of $Z_i$ and $\widetilde{Z}_i$, respectively. Let $Z_{i+1:n}=(z_1,\ldots,z_{i-1},z_i,Z_{i+1},\ldots,Z_n)$ be the sequence where $Z_{i+1},\ldots,Z_n$ are obtained by sampling without replacement from $[n]\backslash \{z_1,\ldots,z_{i-1},z_i\}$, and $\widetilde{Z}_{i+1:n} = (z_1,\ldots,z_{i-1},\tilde{z}_i,\widetilde{Z}_{i+1},\ldots,\widetilde{Z}_n)$ be the sequence where $\widetilde{Z}_{i+1},\ldots,\widetilde{Z}_n$ are obtained by sampling without replacement from $[n]\backslash \{z_1,\ldots,z_{i-1},\tilde{z}_i\}$. The meaning of Eq.~(\ref{martingale}) is the maximum value of the expectation $\mathbb{E}[\mathcal{E}(w,Z_{i+1:n})-\mathcal{E}(w,\widetilde{Z}_{i+1:n})]$ over any possible values $z_1,\ldots,z_{i-1},z_i,\tilde{z}_i$. To this end, it is sufficient to consider two cases: (1) $z_i$ appears at positions $i+1$ to $m$ of $\widetilde{Z}_{i+1:n}$, and (2) $z_i$ appears at positions $m+1$ to $n$ of $\widetilde{Z}_{i+1:n}$. For case~(1), the expectation is equal to zero since for each realization of $Z_{i+1:n}$ we can always find a corresponding and unique realization of $\widetilde{Z}_{i+1:n}$ such that they are equal to each other. For case~(2), the maximum of the expectation is $\frac{m+u}{mu}[\ell(w,s_{\tilde{z}_i}) - \ell(w,s_{z_i})]$. To see this, notice that for each realization of $Z_{i+1:n}$, we can always find a corresponding and unique realization of $\widetilde{Z}_{i+1:n}$ such that $Z_{i+1:n}$ and $\widetilde{Z}_{i+1:n}$ only differs at the $i$-th position. The last step is to compute the probability of $z_i$ appearing at positions $m+1$ to $n$ of $\widetilde{Z}_{i+1:n}$. To this end, we need to sample $m-i$ elements among the rest $n-i-1$ elements (that is, the set $[n] \backslash \{z_1,\ldots,z_i,\tilde{z}_{i}\}$), and then apply permutation on them. Thus, the probability is given by $\frac{u!(m-i)!C_{n-i-1}^{m-i}}{(n-i)!}$. Putting all above ingredients together and notice that $\frac{m+u}{mu}[\ell(w,s_{\tilde{z}_i}) - \ell(w,s_{z_i})] \leq \frac{(m+u)B}{mu}$, we obtain
\begin{equation}
   \xi^{\rm sup}_i - \xi^{\rm inf}_i = \begin{cases}
         \frac{u!(m-i)!C_{n-i-1}^{m-i}}{(n-i)!} \cdot \frac{(m+u)B}{mu} = \frac{(m+u)B}{m(m+u-i)}, & i = 1,\ldots,m, \\
         0, & i = m+1, \ldots, n.
        \end{cases}
\end{equation}
Then we have
\begin{equation}
\begin{aligned}
   \mathbb{E} \left[ \exp \left\{ \lambda\sum_{i=1}^n \xi_i \right\} \right] = & \mathbb{E} \left[ \mathbb{E} \left[ \exp \left\{ \lambda\sum_{i=1}^{n-1} \xi_i \right\} \exp \left\{ \lambda \xi_n \right\} \Bigg| Z_1, \ldots, Z_{n-1} \right] \right] \\
   = & \mathbb{E} \left[ \exp \left\{ \lambda\sum_{i=1}^{n-1} \xi_i \right\} \mathbb{E} \left[ \exp \left\{ \lambda \xi_n \right\} | Z_1, \ldots, Z_{n-1} \right] \right] \\
   \leq & \exp \left\{\frac{\lambda^2 (\xi^{\rm sup}_n-\xi^{\rm inf}_n)^2}{8} \right\} \mathbb{E} \left[ \exp \left\{ \lambda\sum_{i=1}^{n-1} \xi_i \right\} \right],
\end{aligned}
\end{equation}
where the second line is due to the tower property of conditional expectation and the last line is obtained by Hoeffding's inequality \citep{Hoeffding1963}. By iteration we obtain 
\begin{equation}
\begin{aligned}
   & \mathbb{E} \left[ \exp \left\{ \lambda\sum_{i=1}^n \xi_i \right\} \right] \\
   \leq & \exp \left\{ \frac{\lambda^2}{8}\sum_{i=1}^{n} (\xi^{\rm sup}_i-\xi^{\rm inf}_i)^2 \right\} = \exp \left\{ \frac{\lambda^2B^2(m+u)^2}{8m^2}\sum_{i=1}^{m} \frac{1}{(m+u-i)^2} \right\} \\
   = & \exp \left\{ \frac{\lambda^2B^2(m+u)^2}{8m^2}\sum_{i=u}^{m+u-1} \frac{1}{i^2} \right\} < \exp \left\{ \frac{\lambda^2B^2(m+u)^2}{8m^2}\sum_{i=u}^{m+u-1} \frac{1}{i^2-1/4} \right\} \\
   = & \exp \left\{ \frac{\lambda^2B^2(m+u)^2}{8m^2}\sum_{i=u}^{m+u-1} \left( \frac{1}{i-1/2} - \frac{1}{i+1/2} \right) \right\} \\
   = & \exp \left\{ \frac{\lambda^2B^2(m+u)^2}{8m(u-1/2)(m+u-1/2)} \right\}.
\end{aligned}
\end{equation}
For the case that $u \leq m$, by the symmetry of $m$ and $u$, we have
\begin{equation}
    \mathbb{E} \left[ \exp \left\{ \lambda\sum_{i=1}^n \xi_i \right\} \right] \leq \exp \left\{ \frac{\lambda^2B^2(m+u)^2}{8u(m-1/2)(m+u-1/2)} \right\}.
\end{equation}
Accordingly, the final bound is obtained by the smallest one of these two bounds,
\begin{equation}\label{sub}
\begin{aligned}
    \mathbb{E} \left[ \exp \left\{ \lambda\sum_{i=1}^n \xi_i \right\} \right] & \leq \exp \left\{ \frac{\lambda^2B^2(m+u)^2}{8mu(m+u-1/2)} \cdot \frac{2\max(m,u)}{2\max(m,u)-1} \right\} \\
    & = \exp \left\{ \frac{\lambda^2 B^2(m+u)C_{m,u}}{8mu} \right\}.
\end{aligned}
\end{equation}
Combining Eq.~(\ref{sub}) and the facts that $\mathcal{E}(w, {Z}) - \mathbb{E}[\mathcal{E}(w, {Z})] = \sum_{i=1}^n \xi_i$ and $\mathbb{E}[\mathcal{E}(w, {Z})] = 0$, for any $w \in \mathcal{W}$ we have
\begin{equation}\label{exp}
\begin{aligned}
   & \mathbb{E}_{Z} \left[ \exp \left\{ \lambda(R_{\rm test}(w, Z) - R_{\rm train}(w, Z)) \right\} \right] \leq \exp \left\{ \frac{\lambda^2 B^2(m+u)C_{m,u}}{8mu} \right\}.
\end{aligned}
\end{equation}
Denote by $Z'$ the independent copy of $Z$, which is independent of $W$ and has the same distribution as $Z$. Then we have
\begin{equation}
\begin{aligned}
   & \log \mathbb{E}_{W, Z'} \left[ \exp \left\{ \lambda(R_{\rm test}(W, Z') - R_{\rm train}(W, Z')) \right\} \right] \\
   = & \log \left( \int_w \mathbb{E}_{Z'} \left[ \exp \left\{ \lambda(R_{\rm test}(w, Z') - R_{\rm train}(w, Z')) \right\} \right] \dif P_{W}{(w)} \right) \\
   \leq & \log \left( \int_w \exp \left\{ \frac{\lambda^2(m+u)C_{m,u}}{8mu} \right\} \dif P_{W}{(w)} \right) \\
   = & \frac{\lambda^2 B^2 C_{m,u}(m+u)}{8mu}.
\end{aligned}
\end{equation}
By Lemma~\ref{lemma}, for any $\lambda \in \mathbb{R}$ we have
\begin{equation}\label{dv}
\begin{aligned}
   & {\rm D}_{\rm KL}(P_{{Z}, W} || P_{Z', W}) \\
   \geq & \mathbb{E}_{{Z}, W} \left[ \lambda(R_{\rm test}(W, Z) - R_{\rm train}(W, Z)) \right] - \log \mathbb{E}_{Z', W} \left[ \exp \left\{ \lambda(R_{\rm test}(W, Z') - R_{\rm train}(W, Z')) \right\} \right] \\
   \geq & \mathbb{E}_{{Z}, W} \left[ \lambda(R_{\rm test}(W, Z) - R_{\rm train}(W, Z)) \right] - \frac{\lambda^2 B^2 C_{m,u}(m+u)}{8mu},
\end{aligned}
\end{equation}
which implies that
\begin{equation}
    \left\vert \mathbb{E}_{{Z}, W} \left[ R_{\rm test}(W, Z) - R_{\rm train}(W, Z) \right] \right\vert \leq \sqrt{\frac{B^2C_{m,u}I(W;Z)(m+u)}{2mu}}.
\end{equation}
This finishes the proof for Eq.~(\ref{first}). For Eq.~(\ref{sqexp}), notice that Eq.~(\ref{exp}) can be rewritten as $\mathbb{E}_Z[ \exp \{\lambda \mathcal{E}(w,Z)\}] \leq \exp \{ \lambda^2 \sigma_{m,u} \}$, where $\sigma_{m,u} \triangleq \frac{B^2 C_{m,u}(m+u)}{8mu}$. Replacing $\mathcal{E}(w,Z)$ with $-\mathcal{E}(w,Z)$, we have $\mathbb{E}_Z[ \exp \{-\lambda \mathcal{E}(w,Z)\}] \leq \exp \{ \lambda^2 \sigma_{m,u} \}$. Therefore, one can find that
\begin{equation}
\begin{aligned}
    \mathbb{P} \left\{ \left\vert \mathcal{E}(w,Z) \right\vert \geq t \right\} \leq & \mathbb{P} \left\{ \mathcal{E}(w,Z) \geq t \right\} + \mathbb{P} \left\{ \mathcal{E}(w,Z) \leq -t \right\} \leq 2 \exp \left\{ -\frac{t^2}{4\sigma_{m,u}} \right\},
\end{aligned}
\end{equation}
where the first and the second inequality are due to Boole’s inequality and the Chernoff technique, respectively. For any $k \in \mathbb{N}_+$, we have
\begin{equation}
\begin{aligned}
    \mathbb{E} \left[ \left\vert \mathcal{E}(w,Z) \right\vert^k \right] = & \int_{0}^\infty \mathbb{P} \{ \left\vert \mathcal{E}(w,Z) \right\vert^k \geq u \} \dif u = k \int_{0}^\infty \mathbb{P} \{ \left\vert \mathcal{E}(w,Z) \right\vert \geq t \} {\,t}^{k-1} \dif t \\
    \leq & 2k \int_{0}^\infty \exp \left\{ -\frac{t^2}{4\sigma_{m,u}} \right\} {\,t}^{k-1}\dif t = (4\sigma_{m,u})^{\frac{k}{2}} k\Gamma(k/2),
\end{aligned}
\end{equation}
which implies that
\begin{equation}
    \mathbb{E} \left[ \exp\{ \lambda \mathcal{E}^2(w,Z) \} \right] = 1 + \sum_{k=1}^\infty \frac{\lambda^k}{k!} \mathbb{E} \left[ \left\vert \mathcal{E}(w,Z) \right\vert^{2k} \right] \leq 1 + 2\sum_{k=1}^\infty (4\lambda \sigma_{m,u})^k.
\end{equation}
By Lemma~\ref{lemma}, for any $\lambda \in \mathbb{R}$ we have
\begin{small}
\begin{equation*}
\begin{aligned}
   & {\rm D}_{\rm KL}(P_{{Z}, W} || P_{Z', W}) \\
   \geq & \mathbb{E}_{{Z}, W} \left[ \lambda (R_{\rm test}(W, Z) - R_{\rm train}(W, Z))^2 \right] - \log \mathbb{E}_{Z', W} \left[ \exp \left\{ \lambda (R_{\rm test}(W, Z') - R_{\rm train}(W, Z'))^2 \right\} \right] \\
   \geq & \mathbb{E}_{{Z}, W} \left[ \lambda (R_{\rm test}(W, Z) - R_{\rm train}(W, Z))^2 \right] - \log \left( 1 + 2\sum_{k=1}^\infty (4\lambda \sigma_{m,u})^k \right),
\end{aligned}
\end{equation*}
\end{small}
Let $\lambda \to 1/(8\sigma_{m,u})$ and plug into $\sigma_{m,u} = \frac{B^2 C_{m,u}(m+u)}{8mu}$, we obtain
\begin{equation}
    \mathbb{E}_{{Z}, W} \left[ (R_{\rm test}(W, Z) - R_{\rm train}(W, Z))^2 \right] \leq \frac{B^2C_{m,u}(m+u)(I(W;Z)+\log 3)}{mu}.
\end{equation}
This finishes the proof.

\section{Proof of Theorem~\ref{thm2}}
\noindent
{\bf Proof}. For any $k\in \mathbb{N}_+$, denote by $Z^{(1)}, \ldots, Z^{(k)}$ the $k$ independent copies of $Z$. For $j \in [k]$ we run a transductive algorithm on $Z^{(j)}$ and obtain a output $W^{(j)} \sim P_{W|Z^{(j)}}$. By this way, $(Z^{(j)}, W^{(j)})$ can be regarded as an independent copy of $(Z,W)$ for $j\in [k]$. Now suppose that there is a monitor that returns 
\begin{equation*}
    (J^*, R^*) \triangleq \mathop{\rm argmax}_{j\in [k], r \in \{\pm 1\}} r\mathcal{E}(W^{(j)}, Z^{(j)}), \ W^* \triangleq W^{(J^*)}.
\end{equation*}
One can verify that
\begin{equation*}
    R^*\mathcal{E}(W^{(J^*)}, Z^{(J^*)}) = \mathop{\rm max}_{j\in [k]} \vert \mathcal{E}(Z^{(j)}, W^{(j)}) \vert.
\end{equation*}
Taking expectation on both side, we have
\begin{equation*}
\begin{aligned}
   & \mathbb{E}_{Z^{(1)}, \ldots, Z^{(k)}, J^*, R^*, W^*} \left[ R^*\mathcal{E}(W^{(J^*)}, Z^{(J^*)}) \right] = \mathbb{E}_{Z^{(1)}, \ldots, Z^{(k)}, W^{(1)}, \ldots, W^{(k)}} \left[ \mathop{\rm max}_{j\in [k]} \vert \mathcal{E}(Z^{(j)}, W^{(j)}) \vert \right].
\end{aligned}
\end{equation*}
Following the same procedure as that in Appendix~\ref{pf1}, we have
\begin{equation*}
\begin{aligned}
    & \log \left( \mathbb{E}_{J^*,R^*,W^*} \mathbb{E}_{Z^{(1)}, \ldots, Z^{(k)}} \left[ \exp \left\{ \lambda R^*\mathcal{E}(W^{(J^*)}, Z^{(J^*)}) \right\} \right] \right) \leq \frac{\lambda^2B^2C_{m,u}(m+u)}{8mu}.
\end{aligned}
\end{equation*}
By Lemma~\ref{lemma}, the following inequality holds for any $\lambda \in \mathbb{R}$
\begin{small}
\begin{equation*}
\begin{aligned}
    & D(P_{Z^{(1)}, \ldots, Z^{(k)}, J^*, R^*, W^*} || P_{Z^{(1)}, \ldots, Z^{(k)}} \otimes P_{J^*, R^*, W^*}) \\
    \geq & \mathbb{E}_{Z^{(1)}, \ldots, Z^{(k)}, J^*, R^*, W^*} \left[ \lambda R^*\mathcal{E}(W^{(J^*)}, Z^{(J^*)}) \right] - \log \left( \mathbb{E}_{J^*,R^*,W^*} \mathbb{E}_{Z^{(1)}, \ldots, Z^{(k)}} \left[ e^{ \lambda R^*\mathcal{E}(W^{(J^*)}, Z^{(J^*)}) } \right] \right) \\
    \geq & \lambda \mathbb{E}_{Z^{(1)}, \ldots, Z^{(k)}, J^*, R^*, W^*} \left[ R^*\mathcal{E}(W^{(J^*)}, Z^{(J^*)}) \right] - \frac{\lambda^2B^2C_{m,u}(m+u)}{8mu}.
\end{aligned}
\end{equation*}
\end{small}%
which implies that
\begin{equation}\label{high}
\resizebox{0.95\hsize}{!}{$
    \begin{aligned}
   & \mathbb{E}_{Z^{(1)}, \ldots, Z^{(k)}, J^*, R^*, W^*} \left[ R^*\mathcal{E}(W^{(J^*)}, Z^{(J^*)}) \right] \leq \sqrt{ \frac{B^2C_{m,u}I(Z^{(1)}, \ldots, Z^{(k)};J^*, R^*, W^*)(m+u)}{2mu}}. 
\end{aligned}
$}
\end{equation}
Next we provide an upper bound for the mutual information term. Notice that
\begin{equation}\label{mutual}
\begin{aligned}
    & I(Z^{(1)}, \ldots, Z^{(k)}; J^*, R^*, W^*) \\
    \leq & I(Z^{(1)}, \ldots, Z^{(k)}; J^*, R^*, W^*, W^{(1)}, \ldots, W^{(k)}) \\
    = & I(Z^{(1)}, \ldots, Z^{(k)};W^{(1)}, \ldots, W^{(k)}) + I(Z^{(1)}, \ldots, Z^{(k)}; J^*, R^*, W^*|W^{(1)}, \ldots, W^{(k)}) \\
    = & \sum_{j=1}^k I(Z^{(j)};W^{(j)}) + I(Z^{(1)}, \ldots, Z^{(k)}; J^*, R^*, W^*|W^{(1)}, \ldots, W^{(k)}) \\
    \leq & k I(Z;W) + \log (2k).
\end{aligned}
\end{equation}
where we have used the fact that $(Z^{(j)}, W^{(j)}), j\in [k]$ are independent copies of $(Z,W)$, and mutual information is determined only by the distribution of two probability measures. Plugging Eq.~(\ref{mutual}) into Eq.~(\ref{high}) yields
\begin{equation}\label{abs}
\begin{aligned}
    & \mathbb{E}_{Z^{(1)}, \ldots, Z^{(k)}, W^{(1)}, \ldots, W^{(k)}} \left[ \mathop{\rm max}_{j\in [k]} \vert \mathcal{E}(Z^{(j)}, W^{(j)}) \vert \right] \leq \sqrt{\frac{B^2C_{m,u}(m+u)}{2mu}(\log (2k) + k I(Z,W)) }.
\end{aligned}
\end{equation}
Since $(Z^{(j)}, W^{(j)})$ are independent copies of $(Z,W)$, for any $\alpha > 0$ we have
\begin{equation*}
    \mathbb{P}_{Z^{(1)}, W^{(1)}, \ldots, Z^{(k)}, W^{(k)}} \left\{ \mathop{\rm max}_{j\in [k]} \vert \mathcal{E}(Z^{(j)}, W^{(j)}) \vert < \alpha \right\} = \left( \mathbb{P}_{Z,W} \left\{ \vert \mathcal{E}(Z, W) \vert < \alpha \right\} \right)^k.
\end{equation*}
By Markov's inequality,
\begin{equation*}
\begin{aligned}
    & \mathbb{P}_{Z^{(1)},W^{(1)}, \ldots, Z^{(k)}, W^{(k)}} \left\{ \mathop{\rm max}_{j\in [k]} \vert \mathcal{E}(Z^{(j)}, W^{(j)}) \vert \geq \alpha \right\} \\
    \leq & \frac{1}{\alpha} \mathbb{E}_{Z^{(1)}, \ldots, Z^{(k)}, W^{(1)}, \ldots, W^{(k)}} \left[ \mathop{\rm max}_{j\in [k]} \vert \mathcal{E}(Z^{(j)}, W^{(j)}) \vert \right] \\
    \leq & \frac{1}{\alpha} \sqrt{ \frac{B^2C_{m,u}(m+u)}{2mu}(\log (2k) + k I(Z,W)) }.
\end{aligned}
\end{equation*}
Therefore,
\begin{equation*}
\begin{aligned}
    & \mathbb{P}_{Z,W} \left\{ \vert \mathcal{E}(Z, W) \vert \geq \alpha \right\} = 1 - \mathbb{P}_{Z,W} \left\{ \vert \mathcal{E}(Z, W) \vert < \alpha \right\} \\
    = & 1 - \left( \mathbb{P}_{Z^{(1)}, W^{(1)}, \ldots, Z^{(k)}, W^{(k)}} \left\{ \mathop{\rm max}_{j\in [k]} \vert \mathcal{E}(Z^{(j)}, W^{(j)}) \vert < \alpha \right\} \right)^{\frac{1}{k}} \\
    = & 1 - \left( 1 - \mathbb{P}_{Z^{(1)}, W^{(1)}, \ldots, Z^{(k)}, W^{(k)}} \left\{ \mathop{\rm max}_{j\in [k]} \vert \mathcal{E}(Z^{(j)}, W^{(j)}) \vert \geq \alpha \right\} \right)^{\frac{1}{k}} \\
    \leq & 1 - \left( 1 - \frac{1}{\alpha} \sqrt{\frac{B^2C_{m,u}(m+u)}{2mu}(\log (2k) + k I(Z,W)) } \right)^{\frac{1}{k}}.
\end{aligned}
\end{equation*}
Let $\alpha = 2 \sqrt{\frac{B^2C_{m,u}(m+u)(\log (2k) + k I(Z,W))}{2mu}}$ and $k = \lfloor \frac{1}{\delta} \rfloor$, we get Eq.~(\ref{highpro}) using the inequality $(1/2)^\delta \geq 1 - \delta$ that holds for any $\delta \in (0,1)$. Let $k=1$, we get Eq.~(\ref{absexp}) from Eq.~(\ref{abs}).

\section{Proof of Proposition~\ref{pro_density}}
\noindent
{\bf Proof}. We start from Eq.~(\ref{exp}). Using Theorem~2.6 (\uppercase\expandafter{\romannumeral4}) of \cite{Wainwright2019} and let $\lambda = 1- \frac{m+u}{mu} \in \left[ \frac{1}{6},1 \right)$, for any $w \in \mathcal{W}$ we have
\begin{equation}
    \mathbb{E}_{Z}\left[ \exp\left\{ \frac{2(mu-m-u)\mathcal{E}^2(w,Z)}{B^2C_{m,u}(m+u)} \right\}\right]  \leq \sqrt{\frac{mu}{m+u}}.
\end{equation}
Denote by $Z'$ the independent copy of $Z$, we have
\begin{equation}\label{leq_sqrt}
\begin{aligned}
    & \mathbb{E}_{Z \otimes W} \left[ \exp\left\{ \frac{2(mu-m-u)\mathcal{E}^2(W,Z)}{B^2C_{m,u}(m+u)} \right\}\right] \\
    = & \mathbb{E}_{Z'}\mathbb{E}_W \left[ \exp\left\{ \frac{2(mu-m-u)\mathcal{E}^2(W,Z')}{B^2C_{m,u}(m+u)} \right\}\right] \\
    = & \int_w \mathbb{E}_{Z'} \left[ \exp\left\{ \frac{2(mu-m-u)\mathcal{E}^2(w,Z')}{B^2C_{m,u}(m+u)} \right\}\right] \dif P_{W}{(w)} \leq \sqrt{\frac{mu}{m+u}}.
\end{aligned}
\end{equation}
Then we have
\begin{equation}\label{inq_sqrt}
\begin{aligned}
    & \mathbb{E}_{W \otimes Z} \left[ \exp\left\{ \frac{2(mu-m-u)\mathcal{E}^2(W,Z)}{B^2C_{m,u}(m+u)} \right\}\right] \\
    \geq & \mathbb{E}_{W \otimes Z} \left[ \mathds{1} \left\{ \frac{\dif P_{W,Z}}{\dif P_W P_Z} > 0 \right\} \exp\left\{ \frac{2(mu-m-u)\mathcal{E}^2(W,Z)}{B^2C_{m,u}(m+u)} \right\}\right] \\
    = & \mathbb{E}_{W, Z} \left[ \left( \frac{\dif P_{W,Z}}{\dif P_W P_Z} \right)^{-1} \exp\left\{ \frac{2(mu-m-u)\mathcal{E}^2(W,Z)}{B^2C_{m,u}(m+u)} \right\} \right] \\
    = & \mathbb{E}_{W,Z} \left[ \exp\left\{ \frac{2(mu-m-u)\mathcal{E}^2(W,Z)}{B^2C_{m,u}(m+u)} - \log \frac{\dif P_{W,Z}}{\dif P_W P_Z}\right\}\right] .
\end{aligned}
\end{equation}
Combining Eq.~(\ref{inq_sqrt}) with Eq.~(\ref{leq_sqrt}) and applying Markov's inequlity gives
\begin{equation}
\begin{aligned}
    & \mathbb{P}_{W,Z} \left( \frac{2(mu-m-u)\mathcal{E}^2(W,Z)}{B^2C_{m,u}(m+u)} - \log\left(\sqrt{\frac{mu}{m+u}} \right) - \log \frac{\dif P_{W,Z}}{\dif P_W P_Z} \geq \log(1/\delta) \right) \\
    = & \mathbb{P}_{W,Z} \left( \exp\left\{ \frac{2(mu-m-u)\mathcal{E}^2(W,Z)}{B^2C_{m,u}(m+u)} - \log\left(\sqrt{\frac{mu}{m+u}} \right) - \log \frac{\dif P_{W,Z}}{\dif P_W P_Z} \right\} \geq 1/\delta \right) \\
    \leq & \delta\mathbb{E}_{W,Z} \left[ \exp\left\{ \frac{2(mu-m-u)\mathcal{E}^2(W,Z)}{B^2C_{m,u}(m+u)} - \log\left(\sqrt{\frac{mu}{m+u}} \right) - \log \frac{\dif P_{W,Z}}{\dif P_W P_Z} \right\}\right] \\
    \leq & \delta.
\end{aligned}
\end{equation}
This finishes the proof.

\section{Proof of Proposition~\ref{prop1}}
\noindent
{\bf Proof}. Denote by $\mathcal{U}$ and $\widetilde{\mathcal{Z}}$ the set containing all values of $U$ and the transductive supersample $\widetilde{Z}$, respectively.
Let $\overline{\mathscr{Z}} \triangleq \{\mathscr{Z}(\widetilde{z},u) | \widetilde{z} \in \widetilde{\mathcal{Z}}, u \in \mathcal{U}\}$ be the set containing all sequences induce by $\widetilde{Z}$ and $U$. Now we show that $(\widetilde{z},u) \mapsto \mathscr{Z}(\widetilde{z},u)$ is a bijection. Clearly, it can be verified that $(\widetilde{z},u) \mapsto \mathscr{Z}(\widetilde{z},u)$ is a surjection. So it is sufficient to show that this mapping is injective. For any two element $\mathscr{Z}(\widetilde{z}_1,u_1), \mathscr{Z}(\widetilde{z}_2,u_2) \in \overline{\mathscr{Z}}$, we claim that $\mathscr{Z}(\widetilde{z}_1,u_1)= \mathscr{Z}(\widetilde{z}_2,u_2) \Rightarrow \widetilde{z}_1 = \widetilde{z}_2$ holds. If $\widetilde{z}_1 \ne \widetilde{z}_2$, by Eq.~(\ref{trc_2}) one can find that there must exist $i \in [m]$ such that either $\mathscr{Z}_i(\widetilde{z}_1,u_1) \ne \mathscr{Z}_i(\widetilde{z}_2,u_2)$ or $\mathscr{Z}_{m+i}(\widetilde{z}_1,u_1) \ne \mathscr{Z}_{m+i}(\widetilde{z}_2,u_2)$. 
Recall that $u_i$ only determines the relative position of $\mathscr{Z}_i(\widetilde{z},u)$ and $\mathscr{Z}_{m+i}(\widetilde{z},u)$ for $i\in [m]$. Since two sequences are equal if and only if the element at each position is equal, we have $\mathscr{Z}(\widetilde{z},u_1)= \mathscr{Z}(\widetilde{z},u_2) \Rightarrow u_1=u_2$, which implies that $(\widetilde{z},u) \mapsto \mathscr{Z}(\widetilde{z},u)$ is a injection. Since $\widetilde{Z}$ is obtained by sampling without replacement from $[n]$, elements in $\widetilde{\mathcal{Z}}$ are different from each other, and $\vert \widetilde{\mathcal{Z}} \vert = (2m)!/2^m$ holds. Clearly, we have $\vert \mathcal{U} \vert = 2^m$. For each $\widetilde{z} \in \widetilde{\mathcal{Z}}$, applying $U$ to permute it generates a different sequence. Thus, there are $(2m)!$ elements in $\overline{\mathscr{Z}}$ and they are different from each other, which implies that $\overline{\mathscr{Z}} = {\rm Perm}((1,\ldots,2m))$ holds. Let $\mathcal{Z}$ be the set containing all values of the sequence $Z$, then we have 
\begin{equation}
\begin{aligned}
    &  \mathbb{E}_{W,Z} \left[ \mathcal{E}(W,Z) \right] = \mathbb{E}_{W,Z} \left[ R_{\rm test}(W,Z) - R_{\rm train}(W,Z) \right] \\
    = & \frac{1}{(2m)!} \sum_{z\in \mathcal{Z}} \mathbb{E}_{W|Z=z} \left[ \frac{1}{m}\sum_{i=m+1}^{2m} \ell(W,s_{z_i}) - \frac{1}{m}\sum_{i=1}^{m} \ell(W,s_{z_i}) \right] \\
    = & \frac{1}{(2m)!} \sum_{u \in \mathcal{U}} \sum_{\widetilde{z}\in \widetilde{\mathcal{Z}}} \mathbb{E}_{W|\widetilde{Z}=\widetilde{z},U=u} \left[ \frac{1}{m}\sum_{i=m+1}^{2m} \ell(W,s_{\mathscr{Z}_i(\widetilde{z},u)}) - \frac{1}{m}\sum_{i=1}^{m} \ell(W,s_{\mathscr{Z}_i(\widetilde{z},u)}) \right] \\
    = & \mathbb{E}_{W,\widetilde{Z},U} [ R_{\rm test}(W,\widetilde{Z},U) - R_{\rm train}(W,\widetilde{Z},U) ].
\end{aligned}
\end{equation}
For the case that $u=km,k\in \mathbb{N}_+$, one can verify that $(\widetilde{z},u) \mapsto \mathscr{Z}(\widetilde{z},u)$ is also a bijection and thus $\overline{\mathscr{Z}} = {\rm Perm}((1,\ldots,(k+1)m))$, following the above process. Then we have
\begin{equation}
\begin{aligned}
    & \mathbb{E}_{W,Z} \left[ \mathcal{E}(W,Z) \right] = \mathbb{E}_{W,Z} \left[ R_{\rm test}(W,Z) - R_{\rm train}(W,Z) \right] \\
    = & \frac{1}{(km+m)!} \sum_{z\in \mathcal{Z}} \mathbb{E}_{W|Z=z} \Bigg[ \frac{1}{km}\sum_{i=m+1}^{(k+1)m} \ell(W,s_{z_i}) - \frac{1}{m}\sum_{i=1}^{m} \ell(W,s_{z_i}) \Bigg] \\
    = & \frac{1}{(km+m)!} \sum_{u \in \mathcal{U}} \sum_{\widetilde{z}\in \widetilde{\mathcal{Z}}} \mathbb{E}_{W|\widetilde{Z}=\widetilde{z},U=u} \Bigg[ \frac{1}{km}\sum_{i=m+1}^{(k+1)m} \ell(W,s_{\mathscr{Z}_i(\widetilde{z},u)}) - \frac{1}{m}\sum_{i=1}^{m} \ell(W,s_{\mathscr{Z}_i(\widetilde{z},u)}) \Bigg] \\
    = & \mathbb{E}_{W,\widetilde{Z},U} [ R_{\rm test}(W,\widetilde{Z},U) - R_{\rm train}(W,\widetilde{Z},U) ],
\end{aligned}
\end{equation}
where
\begin{equation}
\begin{aligned}
   & R_{\rm test}(W,\widetilde{Z},U) \triangleq \frac{1}{km}\sum_{i=m+1}^{(k+1)m} \ell \left( W,s_{\mathscr{Z}_i(\widetilde{Z},U)} \right) = \frac{1}{km}\sum_{i=1}^m \sum_{j=1}^k \left( W,s_{\widetilde{Z}_{i,U_{i,j}}} \right), \\
   & R_{\rm train}(W,\widetilde{Z},U) \triangleq \frac{1}{m}\sum_{i=1}^{m} \ell \left( W,s_{\mathscr{Z}_i(\widetilde{Z},U)} \right) = \frac{1}{m}\sum_{i=1}^m \left( W,s_{\widetilde{Z}_{i,U_{i,0}}} \right).
\end{aligned}
\end{equation}
This completes the proof.

\section{Proof of Theorem~\ref{thm3}}\label{proof3}
\noindent
{\bf Proof}. By Proposition~\ref{prop1}, we have
\begin{equation}
\begin{aligned}
    & \mathbb{E}_{W,Z} \left[ R_{\rm test}(W,Z) - R_{\rm train}(W,Z) \right] \\
    = & \mathbb{E}_{W,\widetilde{Z},U} \left[ R_{\rm test}(W,\widetilde{Z},U) - R_{\rm train}(W,\widetilde{Z},U) \right] \triangleq \mathbb{E}_{W,\widetilde{Z},U} \left[ \mathcal{E}(W,\widetilde{Z},U) \right] \\
    = & \mathbb{E}_{W,\widetilde{Z},U} \bigg[ \frac{1}{m}\sum_{i=m+1}^{2m} \ell \left( W,s_{\widetilde{Z}_{i,1-U_i}} \right) - \frac{1}{m}\sum_{i=1}^{m} \ell \left( W,s_{\widetilde{Z}_{i,U_i}} \right) \bigg].
\end{aligned}
\end{equation}
where we use $\mathcal{E}(W,\widetilde{Z},U)$ as the abbreviation of $R_{\rm test}(W,\widetilde{Z},U) - R_{\rm train}(W,\widetilde{Z},U)$. Denote by $w$ and $\widetilde{z}$ the fixed realizations of $W$ and $\widetilde{Z}$. For any $\lambda \in \mathbb{R}$, by Hoeffding's inequality \citep{Hoeffding1963} we have
\begin{equation}\label{exp2}
\begin{aligned}
    & \mathbb{E}_{U} \left[ \exp \left\{ \lambda \mathcal{E}(w,\widetilde{z},U) \right\} \right] \\
    = & \mathbb{E}_{U} \left[ \exp \left\{ \frac{\lambda}{m} \sum_{i=1}^m \ell(w, s_{\widetilde{z}_{i,1-U_i}}) - \ell(w, s_{\widetilde{z}_{i,U_i}}) \right\} \right] \leq \exp \left\{ \frac{\lambda^2 B^2}{2m} \right\}.
\end{aligned}
\end{equation}
Let $U'$ be the independent copy of $U$, we have
\begin{equation}
\begin{aligned}
   & \log \mathbb{E}_{U',W|\widetilde{Z}=\widetilde{z}} \left[ \exp \left\{ \lambda \mathcal{E}(W,\widetilde{z},U') \right\} \right] \\
   = & \log \left( \int_{w} \mathbb{E}_{U'} \left[ \exp \left\{ \lambda \mathcal{E}(w,\widetilde{z},U') \right\} \right] {\,\mathrm{d}}P_{W|\widetilde{Z}=\widetilde{z}}(w) \right) \leq \frac{\lambda^2B^2}{2m},
\end{aligned}
\end{equation}
where we have used the fact that $P_{U',W|\widetilde{Z}=\widetilde{z}}=P_{W|\widetilde{Z}=\widetilde{z}}P_{U'}$, due to $U'$ is independent to both $\widetilde{Z}$ and $W$. By Lemma~\ref{lemma}, for any $\lambda \in \mathbb{R}$,
\begin{equation}
\begin{aligned}
   I^{\widetilde{z}}(U;W) = & {\rm D}_{\rm KL}(P_{U, W|\widetilde{Z}=\widetilde{z}} || P_{U', W|\widetilde{Z}=\widetilde{z}}) \\
   \geq & \mathbb{E}_{U, W|\widetilde{Z}=\widetilde{z}} \left[ \lambda \mathcal{E}(W,\widetilde{z},U) \right] - \log \mathbb{E}_{U',W|\widetilde{Z}=\widetilde{z}} \left[ \exp \left\{ \lambda \mathcal{E}(W,z,U') \right\} \right] \\
   \geq & \lambda \mathbb{E}_{U, W|\widetilde{Z}=\widetilde{z}} \left[ \mathcal{E}(W,\widetilde{z},U) \right] - \frac{\lambda^2B^2}{2m},
\end{aligned}
\end{equation}
which implies that
\begin{equation*}
    \left\vert \mathbb{E}_{U, W|\widetilde{Z}=\widetilde{z}} \left[ \mathcal{E}(W,\widetilde{z},U) \right] \right\vert \leq \sqrt{\frac{2B^2}{m}I^{\widetilde{z}}(U;W)}.
\end{equation*}
Taking expectation over $\widetilde{Z}$ on both side, we have obtain
\begin{equation}
\begin{aligned}
    & \left\vert \mathbb{E}_{W,Z} \left[ R_{\rm test}(W,Z) - R_{\rm train}(W,Z) \right] \right\vert = \left\vert \mathbb{E}_{W,\widetilde{Z},U} \left[ \mathcal{E}(W,\widetilde{Z},U) \right] \right\vert \\
    \leq & \mathbb{E}_{\widetilde{Z}} \left\vert \mathbb{E}_{U, W|\widetilde{Z}} \left[ \mathcal{E}(W,\widetilde{Z},U) \right] \right\vert \leq \mathbb{E}_{\widetilde{Z}} \sqrt{\frac{2B^2}{m}I^{\widetilde{Z}}(U;W)}.
\end{aligned}
\end{equation}
For the second part, note that Eq.~(\ref{exp2}) can be rewritten as $\mathbb{E}_{U} \left[ \exp \left\{ \lambda \mathcal{E}(w,\widetilde{z},U) \right\} \right] \leq \exp \{ \lambda^2 B^2 /2m \}$. Similarly we have $\mathbb{E}_{U} \left[ \exp \left\{ -\lambda \mathcal{E}(w,\widetilde{z},U) \right\} \right] \leq \exp \{ \lambda^2 B^2 /2m \}$. Following the same procedure as that in Appendix~\ref{pf1}, we have
\begin{equation}
    \mathbb{E}_{U,W|\widetilde{Z}=\widetilde{z}} \left[ \mathcal{E}^2(W,\widetilde{z},U) \right] \leq \frac{4B^2}{m}(I^{\widetilde{z}}(U;W)+\log 3).
\end{equation}
Taking expectation on both side, we have obtain
\begin{equation*}
    \mathbb{E}_{W,Z} \left[ (R_{\rm test}(W,Z) - R_{\rm train}(W,Z))^2 \right] = \mathbb{E}_{W,\widetilde{Z},U} \left[ \mathcal{E}^2(W,\widetilde{Z},U) \right] \leq \frac{4B^2}{m}(I(U;W|\widetilde{Z})+\log 3).
\end{equation*}
Now let us consider a special case that $\ell(\cdot)$ is the zero-one loss. Denote by $w$ and $\widetilde{z}$ the fixed realizations of $W$ and $\widetilde{Z}$. By Proposition~\ref{prop1}, for any $\lambda > 0$ we have
\begin{equation}
\begin{aligned}
    \log \left( \mathbb{E}_U \left[ e^{-\lambda R_{\rm train}(w,\widetilde{z},U)} \right] \right) = & \log \left( \mathbb{E}_U \left[ e^{-\frac{\lambda}{m} \sum_{i=1}^m \ell(w, s_{\widetilde{z}_{i,U_i}})} \right] \right) \\
    = & \log \left( \mathbb{E}_U \left[ \prod_{i=1}^m e^{-\frac{\lambda}{m} \ell(w, s_{\widetilde{z}_{i,U_i}})} \right] \right) \\
    = & \log \left( \prod_{i=1}^m \mathbb{E}_{U_i} \left[ e^{-\frac{\lambda}{m} \ell(w, s_{\widetilde{z}_{i,U_i}})} \right] \right) \\
    = & \sum_{i=1}^m \log \left( \mathbb{E}_{U_i} \left[ e^{-\frac{\lambda}{m} \ell(w, s_{\widetilde{z}_{i,U_i}})} \right] \right) \\
    \leq & m \log \left( \frac{1}{m} \sum_{i=1}^m \mathbb{E}_{U_i} \left[ e^{-\frac{\lambda}{m} \ell(w, s_{\widetilde{z}_{i,U_i}})} \right] \right).
\end{aligned}
\end{equation}
The fact that $\ell(\cdot)$ is the zero-one loss implies
\begin{equation}
    e^{-\frac{\lambda}{m} \ell(w, s_{\widetilde{z}_{i,U_i}})} = 1 - \ell(w, s_{\widetilde{z}_{i,U_i}}) + e^{-\frac{\lambda}{m}}\ell(w, s_{\widetilde{z}_{i,U_i}}).
\end{equation}
Recall that $\Phi_{a}(p) = -a^{-1}\log(1-[1-e^{-a}]p)$ where $a \in \mathbb{R}$ and $0<p<1$, define
\begin{equation}
    R(W) \triangleq \frac{1}{2} \left( R_{\rm train}(W,Z) + R_{\rm test}(W,Z) \right) = \frac{1}{2m}\sum_{i=1}^{2m} \ell(W,s_{Z_i}),
\end{equation}
we have
\begin{equation}
\begin{aligned}
    & m \log \left( \frac{1}{m} \sum_{i=1}^m \mathbb{E}_{U_i} \left[ e^{-\frac{\lambda}{m} \ell(w, s_{\widetilde{z}_{i,U_i})})} \right] \right) \\
    = & m \log \left( \frac{1}{m} \sum_{i=1}^m \mathbb{E}_{U_i} \left[ 1 - \ell(w, s_{\widetilde{z}_{i,U_i})}) + e^{-\frac{\lambda}{m}}\ell(w, s_{\widetilde{z}_{i,U_i})}) \right] \right) \\
    = & m \log \left( 1 - \frac{1}{m} \sum_{i=1}^m \mathbb{E}_{U_i} \left[ \ell(w, s_{\widetilde{z}_{i,U_i})})\right] + \frac{e^{-\frac{\lambda}{m}}}{m} \sum_{i=1}^m \mathbb{E}_{U_i} \left[ \ell(w, s_{\widetilde{z}_{i,U_i})})\right] \right) \\
    = & m \log \left( 1 - \frac{1}{2m} \sum_{i=1}^{2m} \ell(w, s_i) + \frac{e^{-\frac{\lambda}{m}}}{2m} \sum_{i=1}^{2m} \ell(w, s_i) \right) = -\lambda \Phi_{\lambda/m}(R(w)).
\end{aligned}
\end{equation}
Combining the above results and rearranging the term we get
\begin{equation}
    \mathbb{E}_{U} \left[ e^{\lambda( \Phi_{\lambda/m}(R(w)) - R_{\rm train}(w,\widetilde{z},U))} \right] \leq 1.
\end{equation}
By Lemma~\ref{lemma}, for any posterior $Q$ and prior $P$ we have
\begin{equation}
\begin{aligned}
    & \mathbb{E}_{U} \left[ e^{\mathop{\rm sup}_{Q \in \mathcal{P}(\mathcal{W})} \mathbb{E}_{W\sim Q_{W|\widetilde{z},U}} \left[\lambda \left( \Phi_{\lambda/m}(R(W)) - R_{\rm train}(W,\widetilde{z},U) \right) \right] - {\rm D}_{\rm KL}(Q || P)} \right] \\
    = & \mathbb{E}_{U} \mathbb{E}_{W\sim P} \left[e^{\lambda \left( \Phi_{\lambda/m}(R(W)) - R_{\rm train}(W,\widetilde{z},U) \right)} \right] \\
    = & \int_w \mathbb{E}_{U} \left[ e^{ \lambda \left( \Phi_{\lambda/m}(R(w)) - R_{\rm train}(w,\widetilde{z},U) \right) } \right] {\,\mathrm{d}}P{(w)} \leq 1.
\end{aligned}
\end{equation}
Taking the expectation over $\widetilde{Z}$ on both side and applying the Markov's inequality, for any $0<\delta<1$ and $\lambda > 0$, with probability over the randomness of $\widetilde{Z}$ and $U$:
\begin{equation}
    \mathop{\rm sup}_{Q \in \mathcal{P}(\mathcal{W})} \mathbb{E}_{W\sim Q} \left[ \Phi_{\lambda/m}(R(W)) - R_{\rm train}(W,\widetilde{Z},U) \right] \leq \frac{{\rm D}_{\rm KL}(Q || P) + \log(1/\delta)}{\lambda}.
\end{equation}

We close this proof by presenting the results for the cases that $u=km$ with $k \in \mathbb{N}_+$. Recall that the transductive training and test error under these cases are defined as
\begin{equation}
\begin{aligned}
    & R_{\rm test}(W,\widetilde{Z},U) \triangleq \frac{1}{km}\sum_{i=m+1}^{(k+1)m} \ell \left( W,s_{\mathscr{Z}_i(\widetilde{Z},U)} \right) = \frac{1}{km}\sum_{i=1}^m \sum_{j=1}^k \left( W,s_{\widetilde{Z}_{i,U_{i,j}}} \right), \\
    & R_{\rm train}(W,\widetilde{Z},U) \triangleq \frac{1}{m}\sum_{i=1}^{m} \ell \left( W,s_{\mathscr{Z}_i(\widetilde{Z},U)} \right) = \frac{1}{m}\sum_{i=1}^m \left( W,s_{\widetilde{Z}_{i,U_{i,0}}} \right).
\end{aligned}
\end{equation}
By Proposition~\ref{prop1} we have 
\begin{equation*}
\begin{aligned}
    & \mathbb{E}_{W,Z} \left[ R_{\rm test}(W,Z) - R_{\rm train}(W,Z) \right] = \mathbb{E}_{W,\widetilde{Z},U} \left[ R_{\rm test}(W,\widetilde{Z},U) - R_{\rm train}(W,\widetilde{Z},U) \right], \\
    & \mathbb{E}_{W,Z} \left[ (R_{\rm test}(W,Z) - R_{\rm train}(W,Z))^2 \right] = \mathbb{E}_{W,\widetilde{Z},U} \left[ (R_{\rm test}(W,\widetilde{Z},U) - R_{\rm train}(W,\widetilde{Z},U))^2 \right].
\end{aligned}
\end{equation*} 
Following the proof in this part and plugging into $m=\frac{n}{k+1}$ we obtain
\begin{equation}
\begin{aligned}
    & \left\vert \mathbb{E}_{W,Z} \left[ R_{\rm test}(W,Z) - R_{\rm train}(W,Z) \right] \right\vert \leq \mathbb{E}_{\widetilde{Z}} \sqrt{\frac{2(k+1)B^2}{n}I^{\widetilde{Z}}(U;W)}, \\
    & \mathbb{E}_{W,Z} \left[ (R_{\rm test}(W,Z) - R_{\rm train}(W,Z))^2 \right] \leq \frac{4(k+1)B^2}{n}(I(U;W|\widetilde{Z})+\log 3).
\end{aligned}
\end{equation}
This completes the proof. 

\section{Proof of Corollary~\ref{cor1}}\label{cor1_proof}
\noindent
{\bf Proof}. Denote by 
\begin{equation}
g(F_i,U_i,\widetilde{Z}) \triangleq r(F_{i,1-U_i}, y_{\widetilde{Z}_{i,1-U_i}}) - r(F_{i,U_i}, y_{\widetilde{Z}_{i,U_i}})    
\end{equation}
the function of $(F_i,U_i,\widetilde{Z})$. Recall that $F_{i,U_i} = f_W(x_{\widetilde{Z}_{i,U_i}})$ and $F_{i,1-U_i} = f_W(x_{\widetilde{Z}_{i,1-U_i}})$ hold. Let $f_i$ and $\widetilde{z}$ be the fixed realizations of $F_i$ and $\widetilde{Z}$. For any $\lambda \in \mathbb{R}$ and $i\in [m]$, by Hoeffding's inequality \citep{Hoeffding1963} we have
\begin{equation}
    \mathbb{E}_{U_i} \left[ \exp \left\{ \lambda g(f_i, U_i, \widetilde{z}) \right\} \right] \leq \exp \left\{ \frac{\lambda^2 B^2}{2} \right\}.
\end{equation}
Let $U'_i$ be the independent copy of $U_i$, by Lemma~\ref{lemma},
\begin{equation}
\begin{aligned}
   I(F_i;U_i|\widetilde{Z}=\widetilde{z}) & \geq \lambda \mathbb{E}_{F_i,U_i|\widetilde{Z}=\widetilde{z}} \left[ g(F_i, U_i, \widetilde{z}) \right] - \log \mathbb{E}_{F_i,U'_i|\widetilde{Z}=\widetilde{z}} \left[ \exp \left\{ \lambda g(F_i, U'_i,\widetilde{z}) \right\} \right] \\
   & \geq \lambda \mathbb{E}_{F_i,U_i|\widetilde{Z}=\widetilde{z}} \left[ g(F_i, U_i,\widetilde{z}) \right] - \frac{\lambda^2B^2}{2},
\end{aligned}
\end{equation}
which implies that
\begin{equation}
    \left\vert \mathbb{E}_{F_i,U_i|\widetilde{Z}=\widetilde{z}} \left[ g(F_i, U_i,\widetilde{z}) \right] \right\vert \leq B\sqrt{2I^{\widetilde{z}}(F_i;U_i)}.
\end{equation}
Taking expectation over $\widetilde{Z}$ on both side yields
\begin{equation}
    \mathbb{E}_{\widetilde{Z}}\left\vert \mathbb{E}_{F_i,U_i|\widetilde{Z}} \left[ g(F_i, U_i,\widetilde{Z}) \right] \right\vert \leq B\mathbb{E}_{\widetilde{Z}}\sqrt{2I^{\widetilde{Z}}(F_i;U_i)}.
\end{equation}
Then we have
\begin{equation}\label{equ}
\begin{aligned}
   & \left\vert \mathbb{E}_{W,Z} \left[ R_{\rm test}(W,Z) - R_{\rm train}(W,Z) \right] \right\vert = \left\vert \mathbb{E}_{W,\widetilde{Z},U} \left[ \mathcal{E}(W,\widetilde{Z},U) \right] \right\vert \\
   = & \left\vert \mathbb{E}_{W,\widetilde{Z},U} \left[ \frac{1}{m} \sum_{i=m+1}^{2m}  \ell \left( W,s_{\mathscr{Z}_i(\widetilde{Z},U)} \right) - \frac{1}{m} \sum_{i=1}^m \ell \left( W,s_{\mathscr{Z}_i(\widetilde{Z},U)} \right) \right] \right\vert \\
   = & \left\vert \frac{1}{m} \sum_{i=1}^m \mathbb{E}_{\widetilde{Z}} \mathbb{E}_{W,U_i|\widetilde{Z}} \left[ \ell \left( W,s_{\widetilde{Z}_{i,1-U_i}} \right) - \ell \left( W,s_{\widetilde{Z}_{i,U_i}} \right) \right] \right\vert \\
   \leq & \frac{1}{m} \sum_{i=1}^m \mathbb{E}_{\widetilde{Z}} \left\vert \mathbb{E}_{W,U_i|\widetilde{Z}} \left[ \ell \left( W,s_{\widetilde{Z}_{i,1-U_i}} \right) - \ell \left( W,s_{\widetilde{Z}_{i,U_i}} \right) \right] \right\vert \\
   = & \frac{1}{m} \sum_{i=1}^m \mathbb{E}_{\widetilde{Z}} \left\vert \mathbb{E}_{W,U_i|\widetilde{Z}} \left[ r \big(f_W(x_{\widetilde{Z}_{i,1-U_i}}), y_{\widetilde{Z}_{i,1-U_i}} \big) - r \big(f_W(x_{\widetilde{Z}_{i,U_i}}), y_{\widetilde{Z}_{i,U_i}} \big) \right] \right\vert \\
   = & \frac{1}{m} \sum_{i=1}^m \mathbb{E}_{\widetilde{Z}} \left\vert \mathbb{E}_{F_i,U_i|\widetilde{Z}} \left[ r (F_{i,1-U_i}, y_{\widetilde{Z}_{i,1-U_i}} ) - r (F_{i,U_i}, y_{\widetilde{Z}_{i,U_i}} \big) \right] \right\vert \\
   = & \frac{1}{m} \sum_{i=1}^m \mathbb{E}_{\widetilde{Z}} \left\vert \mathbb{E}_{F_i,U_i|\widetilde{Z}} \left[ g(F_i, U_i,\widetilde{Z}) \right] \right\vert \leq \frac{B}{m} \sum_{i=1}^m \mathbb{E}_{\widetilde{Z}} \sqrt{2I^{\widetilde{Z}}(F_i;U_i)}.
\end{aligned}
\end{equation}
Denote by $g(L_i, U_i) = L_{i,1-U_i} - L_{i,U_i}$ and $g(\Delta_i, U_i) \triangleq (-1)^{U_i}\Delta_i$, following the above procedure we have
\begin{align}
   & \left\vert \mathbb{E}_{W,Z} \left[ R_{\rm test}(W,Z) - R_{\rm train}(W,Z) \right] \right\vert \leq \frac{B}{m} \sum_{i=1}^m \mathbb{E}_{\widetilde{Z}} \sqrt{2I^{\widetilde{Z}}(L_i;U_i)}, \label{eql} \\
   & \left\vert \mathbb{E}_{W,Z} \left[ R_{\rm test}(W,Z) - R_{\rm train}(W,Z) \right] \right\vert \leq \frac{B}{m} \sum_{i=1}^m \mathbb{E}_{\widetilde{Z}} \sqrt{2I^{\widetilde{Z}}(\Delta_i;U_i)}.
\end{align}
For the cases that $u=km,k\in \mathbb{N}_+$, define
$F_{i} \triangleq (f_W(X_{\widetilde{Z}_{i,0}}), \ldots, f_W(X_{\widetilde{Z}_{i,k}})), i\in [m]$ and $L_{i,:} \triangleq (\ell(W,S_{\widetilde{Z}_{i,0}}),\ldots, \ell(W,S_{\widetilde{Z}_{i,k}})), i\in [m]$ be the prediction of model and the sequence of loss values. Denote by $g(F_i,U_i,\widetilde{Z}) \triangleq \frac{1}{k}\sum_{j=1}^k r \big( F_{i,U_{i,j}}, y_{\widetilde{Z}_{i,U_{i,j}}} \big) - r \big( F_{i,U_{i,0}}, y_{\widetilde{Z}_{i,U_{i,0}}} \big)$ and $g(F_i,U_i,\widetilde{Z}) \triangleq \frac{1}{k}\sum_{j=1}^k L_{i,U_{i,j}}, - L_{i,U_{i,0}}$ the function of $(F_i,U_i,\widetilde{Z})$ and $(L_i,U_i,\widetilde{Z})$, respectively. Following the above process one can verify that Eqs.~(\ref{equ},\ref{eql}) still hold under the cases that $u=km,k\in \mathbb{N}_+$. This finishes the proof.

\section{Proof of Theorem~\ref{thm4}}
\noindent
{\bf Proof}. We start from Eq.~(\ref{exp}). By Lemma~\ref{lemma}, for any $\lambda > 0$ we have
\begin{equation}\label{pac_1}
\begin{aligned}
    & \mathbb{E}_{Z} \left[e^{\mathop{\rm sup}_{Q \in \mathcal{P}(\mathcal{W})} \lambda \mathbb{E}_{W \sim Q} \left[R_{\rm test}(W, Z) - R_{\rm train}(W, Z) \right] - {\rm D}_{\rm KL}(Q || P) } \right] \\
    = & \mathbb{E}_{Z} \mathbb{E}_{W \sim P} \left[ e^{\lambda(R_{\rm test}(W, Z) - R_{\rm train}(W, Z))} \right] \\
    = & \int_w \mathbb{E}_{Z} \left[ e^{\lambda(R_{\rm test}(w, Z) - R_{\rm train}(w, Z))} \right] {\,\mathrm{d}}P{(w)} \\
    \leq & e^{\frac{\lambda^2B^2C_{m,u}(m+u)}{8mu}}.
\end{aligned}
\end{equation}
By Jensen's inequality we have
\begin{equation}\label{pac_2}
\begin{aligned}
    & \mathbb{E}_{Z} \left[ e^{\mathop{\rm sup}_{Q \in \mathcal{P}(\mathcal{W})} \lambda \mathbb{E}_{W \sim Q} \left[R_{\rm test}(W, Z) - R_{\rm train}(W, Z) \right] - {\rm D}_{\rm KL}(Q || P)} \right] \\
    \geq & e^{\mathbb{E}_{Z} \left[ \mathop{\rm sup}_{Q \in \mathcal{P}(\mathcal{W})} \lambda \mathbb{E}_{W \sim Q} \left[R_{\rm test}(W, Z) - R_{\rm train}(W, Z) \right] - {\rm D}_{\rm KL}(Q || P) \right] }.
\end{aligned}
\end{equation}
Combining Eq.~(\ref{pac_1}) with Eq.~(\ref{pac_2}), for any posterior $Q \in \mathcal{P}(\mathcal{W})$ and $\lambda > 0$ we have
\begin{equation}
    \mathbb{E}_{Z}\mathbb{E}_{W \sim Q} \left[R_{\rm test}(W, Z) - R_{\rm train}(W, Z) \right] \leq  \frac{\lambda B^2C_{m,u}(m+u)}{8mu} + \frac{\mathbb{E}_{Z}\left[{\rm D}_{\rm KL}(Q || P) \right]}{\lambda}.
\end{equation}
This gives Eq.~(\ref{average_all}). Using Markov's inequality, for any $\lambda > 0$ and $0 < \delta < 1$:
\begin{footnotesize}
\begin{equation*}
\begin{aligned}
    & \mathbb{P}_Z \left( \mathop{\rm sup}_{Q \in \mathcal{P}(\mathcal{W})} \lambda \mathbb{E}_{W \sim Q} \left[R_{\rm test}(W, Z) - R_{\rm train}(W, Z) \right] - {\rm D}_{\rm KL}(Q || P) - \frac{\lambda^2B^2C_{m,u}(m+u)}{8mu} \geq \log (1/\delta) \right) \\
    = & \mathbb{P}_Z \left( \exp \left\{ \mathop{\rm sup}_{Q \in \mathcal{P}(\mathcal{W})} \lambda \mathbb{E}_{W \sim Q} \left[R_{\rm test}(W, Z) - R_{\rm train}(W, Z) \right] - {\rm D}_{\rm KL}(Q || P) - \frac{\lambda^2B^2C_{m,u}(m+u)}{8mu} \right\}  \geq \frac{1}{\delta} \right) \\
    \leq & \delta \mathbb{E}_{Z} \left[ \exp \left\{ \mathop{\rm sup}_{Q \in \mathcal{P}(\mathcal{W})} \lambda \mathbb{E}_{W \sim Q} \left[R_{\rm test}(W, Z) - R_{\rm train}(W, Z) \right] - {\rm D}_{\rm KL}(Q || P) - \frac{\lambda^2B^2C_{m,u}(m+u)}{8mu} \right\} \right] \leq \delta,
\end{aligned}
\end{equation*}
\end{footnotesize}%
where the last inequality is due to Eq.~(\ref{pac_1}). By rearranging the term we obtain Eq.~(\ref{average}). To obtain Eq.~(\ref{single}), firstly we have
\begin{equation}\label{inq_sqrt2}
\begin{aligned}
    & \mathbb{E}_Z \mathbb{E}_{W \sim P} \left[ \exp\left\{ \lambda \left[R_{\rm test}(W, Z) - R_{\rm train}(W, Z) \right] \right\} \right] \\
    \geq & \mathbb{E}_Z \mathbb{E}_{W \sim P} \left[ \mathds{1} \left\{ \frac{\dif Q}{\dif P} > 0 \right\} \exp\left\{ \lambda \left[R_{\rm test}(W, Z) - R_{\rm train}(W, Z) \right] \right\} \right] \\
    = & \mathbb{E}_Z \mathbb{E}_{W \sim Q} \left[ \left( \frac{\dif Q}{\dif P} \right)^{-1} \exp\left\{ \lambda \left[R_{\rm test}(W, Z) - R_{\rm train}(W, Z) \right] \right\} \right] \\
    = & \mathbb{E}_Z \mathbb{E}_{W \sim Q} \left[ \exp\left\{ \lambda \left[R_{\rm test}(W, Z) - R_{\rm train}(W, Z) \right] - \log \frac{\dif Q}{\dif P} \right\} \right] .
\end{aligned}
\end{equation}
Combining Eq.~(\ref{inq_sqrt2}) with Eq.~(\ref{pac_1}) and using Markov's inequality, for any $0 < \delta < 1$:
\begin{equation}
\begin{aligned}
    & \mathbb{P}_{Z,W\sim Q} \left( \lambda \left[ R_{\rm test}(W, Z) - R_{\rm train}(W, Z) \right] - \log \frac{\dif Q}{\dif P} - \frac{\lambda^2B^2C_{m,u}(m+u)}{8mu} \geq \log(1/\delta) \right) \\
    = & \mathbb{P}_{Z,W\sim Q} \left( \exp\left\{ \lambda \left[ R_{\rm test}(W, Z) - R_{\rm train}(W, Z) \right] - \log \frac{\dif Q}{\dif P} - \frac{\lambda^2B^2C_{m,u}(m+u)}{8mu} \right\} \geq \frac{1}{\delta} \right) \\
    \leq & \delta \mathbb{E}_{Z,W\sim Q} \left[ \exp\left\{ \lambda \left[ R_{\rm test}(W, Z) - R_{\rm train}(W, Z) \right] - \log \frac{\dif Q}{\dif P} - \frac{\lambda^2B^2C_{m,u}(m+u)}{8mu} \right\} \right] \leq \delta.
\end{aligned}
\end{equation}
By rearranging the term we obtain Eq.~(\ref{single}). Interestingly, we can also get a degenerated version of Theorem~\ref{thm1}, following the technique of \cite{Begin2014}. Denote by $R(W) \triangleq \frac{1}{m+u} \sum_{i=1}^{m+u} \ell(W,s_{Z_i}) = \frac{m}{m+u} R_{\rm train}(W,Z) + \frac{u}{m+u} R_{\rm test}(W,Z)$ the average error on $\{s_i\}_{i=1}^n$. Denote by $\mathcal{D}(p,q) \triangleq p \log \frac{p}{q} + (1-p) \log \frac{1-p}{1-q}$ the KL divergence between two Bernoulli distributions with success probability $p$ and $q$. Then the $\mathcal{D}$-function introduced by \cite{Begin2014} is expressed by $\mathcal{D}^*_{\beta}(p,q) \triangleq \mathcal{D}(p,q) + \frac{1-\beta}{\beta} \mathcal{D}(\frac{q-\beta p}{1-\beta},q)$. By Theorem~5 and Theorem~6 of \cite{Begin2014}, for fixed realization $w$ of $W$ we have
\begin{equation}
    \mathbb{E}_Z \left[ \exp \left\{ m \mathcal{D}^*_{\beta}(R_{\rm train}(w,Z), R(w)) \right\} \right] \leq 3\log(m) \sqrt{\frac{mu}{m+u}},
\end{equation}
which implies that
\begin{equation}
\begin{aligned}
   & \log \mathbb{E}_{W\otimes {Z}} \left[ \exp \left\{ m \mathcal{D}^*_{\beta}(R_{\rm train}(W,Z), R(W)) \right\} \right]
   \leq \log \left( 3\log (m) \sqrt{\frac{mu}{m+u}} \right).
\end{aligned}
\end{equation}
By Lemma~\ref{lemma} we have
\begin{equation}\label{pac1}
\begin{aligned}
    & \mathrm{D_{KL}}(P_{Z, W} || P_{Z \otimes W}) \\
    \geq & \mathbb{E}_{Z, W} \left[ m \mathcal{D}^*_{\beta}(R_{\rm train}(W,Z), R(W)) \right] - \log \mathbb{E}_{Z \otimes W} \left[ e^{m \mathcal{D}^*_{\beta}(R_{\rm train}(W,Z), R(W)) } \right] \\
    \geq & m \mathbb{E}_{Z, W} \left[ \mathcal{D}^*_{\beta}(R_{\rm train}(W,Z), R(W)) \right] - \log \left( 3\log (m) \sqrt{\frac{mu}{m+u}} \right).
\end{aligned}
\end{equation}
By Pinsker's inequality \citep[Theorem~4.19]{Boucheron2013} and plugging in $\beta = \frac{m}{m+u}$, the expectation term can be lower bounded by
\begin{equation}\label{pac2}
\begin{aligned}
    & \mathbb{E}_{Z, W} \left[\mathcal{D}^*_{\beta}(R_{\rm train}(W,Z), R(W)) \right] \\
    = & \frac{u}{m} \mathbb{E}_{Z, W} \left[\mathcal{D}\left( \frac{m+u}{u}R(W) - \frac{m}{u}R_{\rm train}(W,Z), R(W) \right) \right] + \mathbb{E}_{Z, W} \left[\mathcal{D}(R_{\rm train}(W,Z), R(W)) \right]\\
    \geq & \frac{2m}{u} \mathbb{E}_{Z, W} \left[ (R_{\rm train}(W,Z) - R(W))^2 \right] + 2\mathbb{E}_{Z, W} \left[ (R_{\rm train}(W,Z) - R(W))^2 \right] \\
    = & \frac{2(m+u)}{u} \mathbb{E}_{Z, W} \left[ \left( R_{\rm train}(W,Z) - \frac{m}{m+u}R_{\rm train}(W,Z) - \frac{u}{m+u}R_{\rm test}(W,Z) \right)^2 \right] \\
    = & \frac{2u}{m+u} \mathbb{E}_{Z, W} \left[ \left( R_{\rm train}(W,Z) - R_{\rm test}(W,Z) \right)^2 \right] \\
    \geq & \frac{2u}{m+u} \left( \mathbb{E}_{Z, W}[ R_{\rm train}(W,Z) - R_{\rm test}(W,Z)] \right)^2.
\end{aligned}
\end{equation}
Combining Eq.~(\ref{pac1}) with Eq.~(\ref{pac2}) we obtain 
\begin{equation*}
    \vert \mathbb{E}_{Z, W}[ R_{\rm train}(W,Z) - R_{u}(W,Z)] \vert \leq \sqrt{\frac{m+u}{2mu} \left[ I(Z;W) + \log \left( 3\log (m) \sqrt{\frac{mu}{m+u}} \right) \right]}.
\end{equation*}
Compared with the one presented in Theorem~\ref{thm1}, this bound is degenerated since it contains a extra factor $\log \left( 3\log (m) \sqrt{\frac{mu}{m+u}} \right)$. This finishes the proof.

\section{Proof of Corollary~\ref{cor_pac}}
\noindent
{\bf Proof}. We start from Eq.~(\ref{pac_1}): for a fixed $\lambda \in \Lambda$ we have
\begin{equation}
    \mathbb{E}_{Z} \left[e^{\mathop{\rm sup}_{Q \in \mathcal{P}(\mathcal{W})} \lambda \mathbb{E}_{W \sim Q} \left[R_{\rm test}(W, Z) - R_{\rm train}(W, Z) \right] - {\rm D}_{\rm KL}(Q || P) - \frac{\lambda^2 B^2 C_{m,u}(m+u)}{8mu} } \right] \leq 1.
\end{equation}
Then
\begin{equation}
\begin{aligned}
    & \mathbb{E}_{Z} \left[e^{\mathop{\rm sup}_{Q \in \mathcal{P}(\mathcal{W}), \lambda \in \Lambda} \lambda \mathbb{E}_{W \sim Q} \left[R_{\rm test}(W, Z) - R_{\rm train}(W, Z) \right] - {\rm D}_{\rm KL}(Q || P) - \frac{\lambda^2 B^2 C_{m,u}(m+u)}{8mu} } \right] \\
    = & \mathbb{E}_{Z} \left[ \mathop{\rm sup}_{\lambda \in \Lambda} e^{\mathop{\rm sup}_{Q \in \mathcal{P}(\mathcal{W})} \lambda \mathbb{E}_{W \sim Q} \left[R_{\rm test}(W, Z) - R_{\rm train}(W, Z) \right] - {\rm D}_{\rm KL}(Q || P) - \frac{\lambda^2 B^2 C_{m,u}(m+u)}{8mu} } \right] \\
    \leq & \sum_{\lambda \in \Lambda} \mathbb{E}_{Z} \left[e^{\mathop{\rm sup}_{Q \in \mathcal{P}(\mathcal{W}), \lambda \in \Lambda} \lambda \mathbb{E}_{W \sim Q} \left[R_{\rm test}(W, Z) - R_{\rm train}(W, Z) \right] - {\rm D}_{\rm KL}(Q || P) - \frac{\lambda^2 B^2 C_{m,u}(m+u)}{8mu} } \right] = \vert \Lambda \vert. 
\end{aligned}
\end{equation}
By Markov's inequality, for any $\lambda > 0$ and $0 < \delta < 1$:
\begin{footnotesize}
\begin{equation*}
\begin{aligned}
    & \mathbb{P}_Z \left( \mathop{\rm sup}_{\substack{Q \in \mathcal{P}(\mathcal{W}) \\ \lambda \in \Lambda}} \lambda \mathbb{E}_{W \sim Q} \left[R_{\rm test}(W, Z) - R_{\rm train}(W, Z) \right] - {\rm D}_{\rm KL}(Q || P) - \frac{\lambda^2B^2C_{m,u}(m+u)}{8mu} \geq \log (\vert \Lambda \vert/\delta) \right) \\
    = & \mathbb{P}_Z \left( \exp \left\{ \mathop{\rm sup}_{\substack{Q \in \mathcal{P}(\mathcal{W}) \\ \lambda \in \Lambda}} \lambda \mathbb{E}_{W \sim Q} \left[R_{\rm test}(W, Z) - R_{\rm train}(W, Z) \right] - {\rm D}_{\rm KL}(Q || P) - \frac{\lambda^2B^2C_{m,u}(m+u)}{8mu} \right\}  \geq \frac{\vert \Lambda \vert}{\delta} \right) \\
    \leq & \frac{\delta}{\vert \Lambda \vert} \mathbb{E}_{Z} \left[ \exp \left\{ \mathop{\rm sup}_{\substack{Q \in \mathcal{P}(\mathcal{W}) \\ \lambda \in \Lambda}} \lambda \mathbb{E}_{W \sim Q} \left[R_{\rm test}(W, Z) - R_{\rm train}(W, Z) \right] - {\rm D}_{\rm KL}(Q || P) - \frac{\lambda^2B^2C_{m,u}(m+u)}{8mu} \right\} \right] \\
    \leq & \delta.
\end{aligned}
\end{equation*}
\end{footnotesize}%
This finishes the proof of the first inequality. For the second inequality, let $\Lambda \triangleq \left\{ e^i, i\in \mathbb{N} \right\} \cap [1,mu/(m+u)]$, we have
\begin{equation}
\begin{aligned}
    & \mathop{\rm inf}_{\lambda \in \Lambda} \left\{ \frac{\lambda C_{m,u}(m+u)}{8mu} + \frac{{\rm D}_{\rm KL}(Q || P) + \log (\vert \Lambda \vert/\delta)}{\lambda} \right\} \\
    \leq & \mathop{\rm inf}_{\lambda \in \left[1, \frac{mu}{m+u} \right]} \left\{ \frac{e^{\lfloor \log \lambda \rfloor} C_{m,u}(m+u)}{8mu} + \frac{1}{e^{\lfloor \log \lambda \rfloor}}\left( {\rm D}_{\rm KL}(Q || P) + \log \left( \frac{1}{\delta}\log \left( \frac{mu}{m+u} \right) \right) \right) \right\}\\
    \leq & \mathop{\rm inf}_{\lambda \in \left[1, \frac{mu}{m+u} \right]} \left\{ \frac{\lambda C_{m,u}(m+u)}{8mu} + \frac{e}{\lambda}\left( {\rm D}_{\rm KL}(Q || P) + \log \left( \frac{1}{\delta}\log \left( \frac{mu}{m+u} \right) \right) \right) \right\},
\end{aligned}
\end{equation}
where the second line is due to the fact that $\vert \Lambda \vert \leq \log \left( \frac{mu}{m+u} \right)$, and the third line is obtained by the inequality $\log (\lambda) -1 \leq \lfloor \log \lambda \rfloor \leq \log \lambda$. This finishes the proof.

\section{Proof of Corollary~\ref{cor2}}
\noindent
{\bf Proof}. This proof is motivated by the proof of Theorem~2 of \cite{Foret2021}. The core idea is that, for a given posterior distribution $Q$, we aim to properly select the optimal prior distribution $P^*$ such that the KL divergence term ${\rm D}_{\rm KL}(Q || P)$ is minimized. However, this solution is not applicable because the prior $P$ obtained by this approach will depend on $Z$. However, $P$ is required to be chosen before observing $Z$. The approach to address this issue is to construct a predefined set of prior distribution $\{P_j\}_{j\in \mathbb{N}_+}$, and then establish a high probability guarantee for each measure in this set. After that, we can establish a high probability guarantee for the optimal prior $P^*$ by using union bound inequality.

Formally, denote by $W \in \mathbb{R}^d$ the parameter returned by the learning algorithm, we define the posterior distribution as $Q \triangleq \mathcal{N}(W, \sigma^2 \mathbf{I}_d)$. Let $c$ be a constant that depends on $m,u,d,\sigma$, whose value will be specified later. The probability measure in the predefined set $\{P_j\}_{j\in \mathbb{N}_+}$ is defined as $P_j = \mathcal{N}(O, \sigma^2_j \mathbf{I}_d)$, where $\sigma^2_j = ce^{(1-j)/d}$. For any $j \in \mathbb{N}_+$, by calculating the KL divergence term, we have
\begin{equation*}
    {\rm D}_{\rm KL}(Q || P_j) = \frac{1}{2} \left[ \frac{d\sigma^2 + \left\Vert W \right\Vert_2^2}{\sigma^2_j} - d + d \log \left( \frac{\sigma^2_j}{\sigma^2} \right) \right],
\end{equation*}
which implies that
\begin{equation*}
    \mathop{\rm argmin}_{\sigma_j} {\rm D}_{\rm KL}(Q || P) = \sqrt{ \frac{d\sigma^2 + \Vert W \Vert^2_2}{d}} .
\end{equation*}
Therefore, we can define the optimal prior distribution as $P^* = \mathcal{N}(O, \sigma^2_{j^*} \mathbf{I})$ with
\begin{equation*}
    j^* = \left\lfloor 1 - d\log \left( \frac{d\sigma^2 + \left\Vert W \right\Vert^2_2}{d c} \right) \right\rfloor.
\end{equation*}
One can verify that the following inequalities hold
\begin{equation}\label{optj}
    -d\log \left( \frac{d\sigma^2 + \left\Vert W \right\Vert^2_2}{dc} \right) \leq j^* \leq 1 - d\log \left( \frac{d\sigma^2 + \left\Vert W \right\Vert^2_2}{dc} \right).
\end{equation}
Combining Eq.~(\ref{optj}) with the equality $\sigma^2_{j^*} = ce^{(1-j^*)/d}$ yields
\begin{equation}
    \sigma^2 + \frac{\Vert W \Vert^2_2}{d} \leq \sigma^2_{j^*} \leq e^{{1}/{d}} \left(\sigma^2 + \frac{\Vert W \Vert^2_2}{d} \right).
\end{equation}
Then we have
\begin{equation}
\begin{aligned}
    {\rm D}_{\rm KL}(Q || P^*) = & \frac{1}{2} \left[ \frac{d\sigma^2 + \left\Vert W \right\Vert_2^2}{\sigma^2_{j^*}} - d + d \log \left( \frac{\sigma^2_{j^*}}{\sigma^2} \right) \right] \\
    \leq & \frac{1}{2} \left[ \frac{d(d\sigma^2 + \Vert W \Vert^2_2)}{d\sigma^2 + \Vert W \Vert^2_2} - d + d \log \left( \frac{e^{{1}/{d}} \left(d\sigma^2 + \Vert W \Vert^2_2 \right)}{d\sigma^2} \right) \right] \\
    = & \frac{d}{2} \log \left( \frac{e^{1/d} \left(d\sigma^2 + \Vert W \Vert^2_2 \right)}{d\sigma^2} \right)  = \frac{1}{2} \left[ 1 + d \log \left( 1 + \frac{\Vert W \Vert^2_2}{d\sigma^2} \right) \right].
\end{aligned}
\end{equation}
For any $\lambda > 0$ and $j\in \mathbb{N}_+$, denote by $A_j$ the event that 
\begin{small}
\begin{equation*}
    A_j \triangleq \left\{ \left\vert \mathbb{E}_{W \sim Q} \left[ R_{\rm test}(W, Z) - R_{\rm train}(W, Z) \right] \right\vert \geq \frac{\lambda C_{m,u}(m+u)}{8mu} + \frac{{\rm D}_{\rm KL}(Q || P_j) + \log (1/\delta_j)}{\lambda} \right\}.
\end{equation*}
\end{small}
Let $\delta_j = \frac{6\delta}{\pi^2 j^2}$, by Theorem~\ref{thm4}, for any posterior distribution $Q$ we have
\begin{equation}
    \mathbb{P} \left\{ A_{j^*} \right\} \leq \mathbb{P} \Bigg\{ \mathop{\bigcup}_{j=1}^{\infty} A_{j} \Bigg\} \leq \sum_{j=1}^{\infty} \mathbb{P} \left\{ A_{j} \right\} = \sum_{j=1}^{\infty} \delta_j = \sum_{j=1}^{\infty} \frac{6\delta}{\pi^2 j^2} = \delta.
\end{equation}
Therefore, with probability at least $1-\delta$, for any $\lambda > 0$:
\begin{equation}\label{pac3}
\begin{aligned}
    & \mathbb{E}_{W \sim Q} \left[ R_{\rm test}(W, Z) - R_{\rm train}(W, Z) \right] \\
    = & \mathbb{E}_{{\bm \epsilon} \sim \mathcal{N}({\bm 0}, \sigma^2 \mathbf{I}_d)} \left[ R_{\rm test}(W+\bm{\epsilon}, Z) - R_{\rm train}(W+\bm{\epsilon}, Z) \right] \\
    \leq & \frac{\lambda C_{m,u}(m+u)}{8mu} + \frac{{\rm D}_{\rm KL}(Q || P_{j^*})+\log(1/\delta_j)}{\lambda} \\
    \leq & \frac{\lambda C_{m,u}(m+u)}{8mu} +\frac{1}{\lambda} \left( \frac{1}{2} \left[ 1 + d \log \left( 1 + \frac{\Vert W \Vert^2_2}{d\sigma^2} \right) \right] + \log \left( \frac{1}{6\delta} \right) + 2\log \left( \pi j^* \right) \right) .
\end{aligned}
\end{equation}
Since $\bm{\epsilon} \sim \mathcal{N}({\bm 0}, \sigma^2 \mathbf{I}_d)$, Lemma~1 of \cite{Beatrice2000} shows that $\mathbb{P} ( \Vert \bm{\epsilon} \Vert^2_2 \geq d\sigma^2 + 2\sigma^2\sqrt{dt} + 2t\sigma^2 ) \leq e^{-t}$. Let $\widetilde{C}_{m,u} \triangleq \sqrt{2\log (mu/(m+u))/d}$, with probability at least $1-(m+u)/mu$ we have
\begin{equation}\label{rho}
\begin{aligned}
    \Vert \bm{\epsilon} \Vert^2_2 \leq & d\sigma^2 + 2\sigma^2\sqrt{d\log(mu/(m+u))} + 2\sigma^2\log(mu/(m+u)) \\
    \leq & d\sigma^2 + 2\sigma^2\sqrt{2d\log(mu/(m+u))} + 2\sigma^2\log(mu/(m+u)) \\
    = & \sigma^2 d \left( 1 + \sqrt{\frac{2\log(mu/(m+u))}{d}} \right)^2 = \sigma^2 d \left( 1 + \widetilde{C}_{m,u} \right)^2 = \rho^2.
\end{aligned}
\end{equation}
Combining the above ingredients together, with probability at least $1-\delta$, for any $\lambda > 0$:
\begin{equation}\label{pac4}
\begin{aligned}
    & R_{\rm test}(W, Z) \\
    \leq & \mathbb{E}_{{\bm \epsilon} \sim \mathcal{N}({\bm 0}, \sigma^2 \mathbf{I}_d)} \left[ R_{\rm test}(W+\bm{\epsilon}, Z) \right] \\
    = & \mathbb{P}\left(\Vert \bm{\epsilon} \Vert^2_2 \leq \rho^2\right) \mathbb{E}_{{\bm \epsilon} \sim \mathcal{N}({\bm 0}, \sigma^2 \mathbf{I}_d)} \left[ R_{\rm test}(W+\bm{\epsilon}, Z) |\Vert \bm{\epsilon} \Vert^2_2 \leq \rho^2 \right] \\
    & + \mathbb{P} \left(\Vert \bm{\epsilon} \Vert^2_2 > \rho^2 \right) \mathbb{E}_{{\bm \epsilon} \sim \mathcal{N}({\bm 0}, \sigma^2 \mathbf{I}_d)} \left[ R_{\rm test}(W+\bm{\epsilon}, Z) |\Vert \bm{\epsilon} \Vert^2_2 > \rho^2 \right] \\
    \leq & \mathbb{E}_{{\bm \epsilon} \sim \mathcal{N}({\bm 0}, \sigma^2 \mathbf{I}_d)} \left[ R_{\rm test}(W+\bm{\epsilon}, Z) |\Vert \bm{\epsilon} \Vert^2_2 \leq \rho^2 \right] + \frac{m+u}{mu} \\
    = & \mathbb{E}_{{\bm \epsilon} \sim \mathcal{N}({\bm 0}, \sigma^2 \mathbf{I}_d)} \left[ R_{\rm test}(W+\bm{\epsilon}, Z) - R_{\rm train}(W+\bm{\epsilon}, Z) + R_{\rm train}(W+\bm{\epsilon}, Z) |\Vert \bm{\epsilon} \Vert^2_2 \leq \rho^2 \right] + \frac{m+u}{mu} \\
    \leq & \max_{\Vert \bm \epsilon \Vert_2 \leq \rho} R_{\rm train}(W+\bm{\epsilon}, Z) + \frac{m+u}{mu} + \frac{\lambda C_{m,u}(m+u)}{8mu} \\
    & + \frac{1}{\lambda} \left( \frac{1}{2} \left[ 1 + d \log \left( 1 + \frac{\Vert W \Vert^2_2}{d\sigma^2} \right) \right] + \log \left( \frac{1}{6\delta} \right) + 2\log \left( \pi j^* \right) \right) \\
    \leq & \max_{\Vert \bm \epsilon \Vert_2 \leq \rho} R_{\rm train}(W+\bm{\epsilon}, Z) + \frac{(\lambda C_{m,u}+8)(m+u)}{8mu}\\
    & + \frac{1}{\lambda} \left( \frac{1}{2} \left[ 1 + d \log \left( 1 + \frac{\Vert W \Vert^2_2}{\rho^2} \left( 1 + \widetilde{C}_{m,u} \right)^2 \right) \right] + \log \left( \frac{1}{6\delta} \right) + 2\log \left( \pi j^* \right) \right).
\end{aligned}
\end{equation}
Here we use the assumption to obtain the second line. The third and the forth line are due to law of total expectation and the fact that $R_{\rm test}(w, z) \leq 1$ for any $w$ and $z$. The remaining step is to specify the value of $c$. Note that if
\begin{equation}
    \Vert W \Vert^2_2 \geq \sigma^2d \left( e^{\frac{4mu}{(m+u)d}} - 1\right) = \frac{\rho^2}{(1+\widetilde{C}_{m,u})^2} \left( e^{\frac{4mu}{(m+u)d}} - 1\right),
\end{equation}
the slack term in Eq.~(\ref{pac4}) will exceed $1$ and the inequality holds trivially. Therefore, we only need to consider the case that
\begin{equation}
    \Vert W \Vert^2_2 < \frac{\rho^2}{(1+\widetilde{C}_{m,u})^2} \left( e^{\frac{4mu}{(m+u)d}} - 1\right) = d\sigma^2 \left( e^{\frac{4mu}{(m+u)d}} - 1\right),
\end{equation}
which implies that
\begin{equation}
    \sigma^2 + \frac{\left\Vert W \right\Vert^2_2}{d} < \sigma^2 e^{\frac{4mu}{(m+u)d}} \triangleq c.
\end{equation}
Here we have used Eq.~(\ref{rho}) since we only need to consider the case that $\Vert \bm{\epsilon} \Vert^2_2 \leq \rho^2$. One can verify that $j^*$ is an valid integer under this definition. Note that
\begin{equation}\label{pac5}
\begin{aligned}
    \log (j^*) & \leq \log \left( 1 + d\log \left( \frac{dc}{d\sigma^2 + \left\Vert W \right\Vert^2_2} \right) \right) \\
    & \leq \log \left( 1 + d\log \left( \frac{c}{\sigma^2} \right) \right) \\
    & = \log \left( 1 + \frac{4mu}{(m+u)} \right) \leq \log \left(\frac{6mu}{m+u} \right).
\end{aligned}
\end{equation}
Plugging Eq.~(\ref{pac5}) into Eq.~(\ref{pac4}), with probability at least $1-\delta$ over the randomness of $Z$, for any $\lambda > 0$:
\begin{small}
\begin{equation*}
\begin{aligned}
    R_{\rm test}(W, Z) \leq & \max_{\Vert \bm \epsilon \Vert_2 \leq \rho} R_{\rm train}(W+\bm{\epsilon}, Z) + \frac{(\lambda C_{m,u}+8)(m+u)}{8mu} \\
    & + \frac{1}{\lambda} \left( \frac{1}{2} \left[ 1 + d \log \left( 1 + \frac{\Vert W \Vert^2_2}{\rho^2} \left( 1 + \widetilde{C}_{m,u} \right)^2 \right) \right] + \log \left( \frac{1}{6\delta} \right) + 2\log \left( \frac{6\pi mu}{m+u} \right) \right).
\end{aligned}
\end{equation*}
\end{small}%
This finishes the proof.

\section{Proof of Corollary~\ref{cor3} and Theorem~\ref{thm_catoni}}
\noindent
{\bf Proof}. We start from Eq.~(\ref{pac_1}) and rewrite it as an inequality under the random sampling setting. Notice that here $Z$ represents the randomness of selecting training labels after the dataset $S$ is observed. For any $\lambda > 0$ and any exchangeable prior $P$:
\begin{equation}\label{new_pac}
\begin{aligned}
    & \mathbb{E}_{S} \left[ e^{\mathop{\rm sup}_{Q \in \mathcal{P}(\mathcal{W})} \lambda \mathbb{E}_{W\sim Q} \left[R_{\rm test}(W, S) - R_{\rm train}(W, S) \right] - {\rm D}_{\rm KL}(Q || P)} \right] \\
    = & \mathbb{E}_{S} \mathbb{E}_{W\sim P} \left[e^{\lambda(R_{\rm test}(W, S) - R_{\rm train}(W, S))} \right] \\
    = & \int_w \mathbb{E}_S \left[ e^{ \frac{\lambda}{u}\sum_{i=m+1}^{m+u} \ell(w,S_i) - \frac{\lambda}{m} \sum_{i=1}^{m} \ell(w,S_i)} \right] {\,\mathrm{d}}P{(w)} \\
    = & \int_w \mathbb{E}_S \mathbb{E}_Z \left[ e^{ \frac{\lambda}{u}\sum_{i=m+1}^{m+u} \ell(w,S_{Z_i}) - \frac{\lambda}{m} \sum_{i=1}^{m} \ell(w,S_{Z_i})} \right] {\,\mathrm{d}}P{(w)} \\
    \leq & e^{\frac{\lambda^2 B^2 C_{m,u}(m+u)}{8mu}}.
\end{aligned}
\end{equation}
By Markov's inequality, for any $0<\delta<1 $ and $\lambda > 0$, with probability $1-\delta$ over the randomness of $Z$ and $S$:
\begin{equation*}
\begin{aligned}
    & \mathop{\rm sup}_{Q \in \mathcal{P}(\mathcal{W})} \mathbb{E}_{W\sim Q} \left[R_{\rm test}(W, S) - R_{\rm train}(W, S) \right] \leq \frac{{\rm D}_{\rm KL}(Q || P)+\log(1/\delta)}{\lambda} + \frac{\lambda B^2 C_{m,u}(m+u)}{8mu}.
\end{aligned}
\end{equation*}
By Jensen's inequality, Eq.~(\ref{new_pac}) implies that for any $\lambda > 0$:
\begin{equation*}
    \mathop{\rm sup}_{Q \in \mathcal{P}(\mathcal{W})} \mathbb{E}_{W\sim Q} \mathbb{E}_{S} \left[\lambda\left[R_{\rm test}(W, S) - R_{\rm train}(W, S) \right] - {\rm D}_{\rm KL}(Q || P) \right] \leq \frac{\lambda^2 B^2 C_{m,u}(m+u)}{8mu}.
\end{equation*}
Denote by $S'$ the independent copy of $S$. Let $Q = P_{W|S}$, $P = \mathbb{E}_{S'}\left[ P_{W|S'} \right]$, we have
\begin{equation*}
\begin{aligned}
    \mathbb{E}_{W,S} \left[R_{\rm test}(W, S) - R_{\rm train}(W, S) \right] \leq \frac{\mathbb{E}_{S}\left[{\rm D}_{\rm KL} \left(P_{W|S} || \mathbb{E}_{S'} \left[ P_{W|S'} \right]\right)\right]}{\lambda} + \frac{\lambda B^2 C_{m,u}(m+u)}{8mu}.
\end{aligned}
\end{equation*}
Using the fact that $\mathbb{E}_{S}\left[{\rm D}_{\rm KL} \left(P_{W|S} || \mathbb{E}_{S'} \left[ P_{W|S'} \right]\right)\right] = I(W;S)$, optimizing $\lambda$ on the r.h.s. yields the following information-theoretic bound,
\begin{equation}
    \mathbb{E}_{W,S} \left[R_{\rm test}(W, S) - R_{\rm train}(W, S) \right] \leq \sqrt{\frac{B^2 C_{m,u}I(W;S)(m+u)}{2mu}}.
\end{equation}
For the self-consistency of this paper, we provide the proof of Catoni's result (\citealp[Theorem~3.1.2]{Catoni2007}). From now on we assume that $\ell(\cdot)$ is the zero-one loss and $u=km$. For any $w \in \mathcal{W}$ and $\lambda > 0$:
\begin{equation}
\begin{aligned}
    \log \left( \mathbb{E}_{U} \left[ e^{-\frac{\lambda}{m}\sum_{i=1}^m \ell \left(w,s_{i+mU_{i,0}} \right)} \right] \right) = & \log \left( \mathbb{E}_{U} \left[ \prod_{i=1}^m e^{-\frac{\lambda}{m} \ell \left(w,s_{i+mU_{i,0}} \right)} \right] \right) \\
    = & \log \left( \prod_{i=1}^m \mathbb{E}_{U_i} \left[ e^{-\frac{\lambda}{m} \ell \left(w,s_{i+mU_{i,0}} \right)} \right] \right) \\
    = & \sum_{i=1}^m \log \left( \mathbb{E}_{U_i} \left[ e^{-\frac{\lambda}{m} \ell \left(w,s_{i+mU_{i,0}} \right)} \right] \right) \\
    \leq & m \log \left( \frac{1}{m}\sum_{i=1}^m \mathbb{E}_{U_i} \left[ e^{-\frac{\lambda}{m} \ell \left(w,s_{i+mU_{i,0}} \right)} \right] \right).
\end{aligned}
\end{equation}
Since $\ell(\cdot)$ is the zero-one loss, it can only take values in $\{0,1\}$, which implies that 
\begin{equation}
    e^{-\frac{\lambda}{m} \ell \left(w,s_{i+mU_{i,0}} \right)} = 1 - \ell \left(w,s_{i+mU_{i,0}} \right) + e^{-\frac{\lambda}{m}} \ell \left(w,s_{i+mU_{i,0}} \right), i\in [m].    
\end{equation}
Recall that $\Phi_a(p) = -a^{-1}\log \left(1-[1-e^{-a}]p \right)$ with $a\in \mathbb{R}$ and $0<p<1$, we have
\begin{equation}
\begin{aligned}
    & \frac{1}{m}\sum_{i=1}^m \mathbb{E}_{U_i} \left[ e^{-\frac{\lambda}{m} \ell \left(w,s_{i+mU_{i,0}} \right)} \right] \\
    = & \frac{1}{m}\sum_{i=1}^m \mathbb{E}_{U_i} \left[ 1 - \ell \left(w,s_{i+mU_{i,0}} \right) + e^{-\frac{\lambda}{m}} \ell \left(w,s_{i+mU_{i,0}} \right) \right] \\
    = & 1 - \frac{1}{m}\sum_{i=1}^m\mathbb{E}_{U_i} \left[\ell \left(w,s_{i+mU_{i,0}} \right) \right] + \frac{e^{-\frac{\lambda}{m}}}{m}\sum_{i=1}^m \mathbb{E}_{U_i} \left[\ell \left(w,s_{i+mU_{i,0}} \right)\right] \\
    = & 1 - \bar{R}(w,s) + e^{-\frac{\lambda}{m}}\bar{R}(w,s),
\end{aligned}
\end{equation}
where $\bar{R}(w,s) \triangleq \frac{1}{(k+1)m} \sum_{i=1}^{(k+1)m} \ell(w,s_i)$. Combining the above results and taking expectation on both side over $S$ gives
\begin{equation}
    \mathbb{E}_S \left[ e^{\lambda \Phi_{\lambda/m}(\bar{R}(w,S))} \mathbb{E}_U \left[ e^{-\frac{\lambda}{m}\sum_{i=1}^m \ell \left(w,S_{i+mU_{i,0}} \right)} \right] \right] \leq 1.
\end{equation}
For any $\lambda > 0$ and any partially exchangeable prior $P$, by Lemma~\ref{lemma} we have:
\begin{equation}
\begin{aligned}
    & \mathbb{E}_{S} \left[ e^{\mathop{\rm sup}_{Q \in \mathcal{P}(\mathcal{W})} \mathbb{E}_{W\sim Q} \left[\lambda \left( \Phi_{\lambda/m}(\bar{R}(W,S)) - R_{\text{train}}(W,S) \right) \right] - {\rm D}_{\rm KL}(Q || P)} \right] \\
    = & \mathbb{E}_{S} \mathbb{E}_{W\sim P} \left[e^{\lambda \left( \Phi_{\lambda/m}(\bar{R}(W,S)) - R_{\text{train}}(W,S) \right)} \right] \\
    = & \int_w \mathbb{E}_{S} \left[ e^{\lambda \Phi_{\lambda/m}(\bar{R}(w,S))} e^{-\frac{\lambda}{m}\sum_{i=1}^m \ell(w,S_i)} \right] {\,\mathrm{d}}P{(w)} \\
    = & \int_w \mathbb{E}_{S} \mathbb{E}_U \left[ e^{\lambda \Phi_{\lambda/m}(\bar{R}(w,S))} e^{-\frac{\lambda}{m}\sum_{i=1}^m \ell(w,S_i)} \right] {\,\mathrm{d}}P{(w)} \\
    = & \int_w \mathbb{E}_{S} \left[ e^{\lambda \Phi_{\lambda/m}(\bar{R}(w,S))} \mathbb{E}_U \left[ e^{-\frac{\lambda}{m}\sum_{i=1}^m \ell(w,S_{i+mU_{i,0}})} \right] \right] {\,\mathrm{d}}P{(w)} \leq 1.
\end{aligned}
\end{equation}
By Markov's inequality, for any $0<\delta<1$ and $\lambda > 0$, with probability $1-\delta$ over the randomness of $S$:
\begin{equation}
    \mathop{\rm sup}_{Q \in \mathcal{P}(\mathcal{W})} \mathbb{E}_{W\sim Q} \left[ \Phi_{\lambda/m}(\bar{R}(W,S)) - R_{\text{train}}(W,S) \right]  \leq \frac{{\rm D}_{\rm KL}(Q || P) + \log (1/\delta)}{\lambda}.
\end{equation}
For any fixed posterior $Q$, the fact that $\Phi_{\lambda/m}(\cdot)$ is a convex function gives
\begin{equation}
    \Phi_{\lambda/m} \left(\mathbb{E}_{W\sim Q}\left[\bar{R}(W,S)\right] \right) \leq \mathbb{E}_{W\sim Q}\left[R_{\text{train}}(W,S)\right] + \frac{{\rm D}_{\rm KL}(Q || P) + \log (1/\delta)}{\lambda},
\end{equation}
which implies that with probability at least $0<\delta<1 $ over the randomness of $S$:
\begin{equation}
\begin{aligned}
    & \mathbb{E}_{W\sim Q}\left[R_{\text{test}}(W,S)\right] \\
    \leq & \frac{(k+1)\left(1-\exp \left(-\frac{\lambda \mathbb{E}_{W\sim Q}\left[R_{\text{train}}(W,S)\right] + {\rm D}_{\rm KL}(Q || P) + \log (1/\delta)}{m} \right)\right)}{k(1-e^{-\lambda/m})} - \frac{1}{k}\mathbb{E}_{W\sim Q}\left[R_{\text{train}}(W,S)\right].
\end{aligned}
\end{equation}
This finishes the proof.

\section{Proof of Theorem~\ref{bound_gap}}
\noindent
{\bf Proof}. Assume that the parameter $w\in \mathbb{R}^d$. Denote by
\begin{equation*}
    \bm{H}(w, Z) \triangleq \frac{1}{u} \sum_{i=m+1}^{m+u} \frac{\partial^2 \ell(w,s_{Z_i})}{\partial w^2} - \frac{1}{m} \sum_{i=1}^{m} \frac{\partial^2 \ell(w,s_{Z_i})}{\partial w^2} \triangleq \bm{H}_{\rm test}(w, Z) - \bm{H}_{\rm train}(w, Z)
\end{equation*}
a function that maps $n$ random variables $Z_1,\ldots,Z_n$ to a self-joint matrix, namely, the Hessian of the transductive generalization gap $\frac{\partial^2 \mathcal{E}({w},Z)}{\partial w^2}$. We remark that $\bm{H}(w, Z) \in \mathbb{R}^{d \times d}$ and $\mathbb{E}[\bm{H}(w, Z)]=\bm{O}$. We firstly establish an upper bound for the moment-generating function $\mathbb{E}_{Z} \left[ e^{\theta {\rm Tr}(\bm{H}(w, Z))} \right]$ by the matrix martingale technique, where $w$ is a fixed realization of $W$. To this end, we construct the following Doob’s martingale difference sequences
\begin{equation}
    \xi_i \triangleq \mathbb{E} [ \bm{H}(w, Z) | Z_1, \ldots, Z_i ] - \mathbb{E} [ \bm{H}(w, Z) | Z_1, \ldots, Z_{i-1} ], i \in [n].
\end{equation}
With this definition, one can verify that $\bm{H}(w, Z) - \mathbb{E}[\bm{H}(w, Z)] = \sum_{i=1}^n \xi_i$. Notice that $\xi_i$ is a function of $Z_1,\ldots,Z_i$. Define
\begin{equation*}
\begin{aligned}
   \xi^{\rm inf}_i & \triangleq \mathop{\rm inf}_z \left\Vert \mathbb{E}[\bm{H}(w, Z)|Z_1,\ldots,Z_{i-1},Z_i=z] - \mathbb{E}[\bm{H}(w, Z)|Z_1,\ldots,Z_{i-1}] \right\Vert, \\
   \xi^{\rm sup}_i & \triangleq \mathop{\rm sup}_z \left\Vert \mathbb{E}[\bm{H}(w, Z)|Z_1,\ldots,Z_{i-1},Z_i=z] - \mathbb{E}[\bm{H}(w, Z)|Z_1,\ldots,Z_{i-1}] \right\Vert, \\
\end{aligned}
\end{equation*}
we have $\xi^{\rm inf}_i \leq \Vert \xi_i \Vert \leq \xi^{\rm sup}_i$. One can find that
\begin{equation*}
\begin{aligned}
    \xi^{\rm sup}_i - \xi^{\rm inf}_i = & \mathop{\rm sup}_z \left\Vert \mathbb{E}[\bm{H}(w, Z)|Z_1,\ldots,Z_{i-1},Z_i=z] - \mathbb{E}[\bm{H}(w, Z)|Z_1,\ldots,Z_{i-1}] \right\Vert \\
    & - \mathop{\rm inf}_z \left\Vert \mathbb{E}[\bm{H}(w, Z)|Z_1,\ldots,Z_{i-1},Z_i=z] - \mathbb{E}[\bm{H}(w, Z)|Z_1,\ldots,Z_{i-1}] \right\Vert \\
    = & \mathop{\rm sup}_{z,\tilde{z}} \Big\{ \left\Vert \mathbb{E}[\bm{H}(w, Z)|Z_1,\ldots,Z_{i-1},Z_i=z] - \mathbb{E}[\bm{H}(w, Z)|Z_1,\ldots,Z_{i-1}] \right\Vert \\
    & - \left\Vert \mathbb{E}[\bm{H}(w, Z)|Z_1,\ldots,Z_{i-1},Z_i=\tilde{z}] - \mathbb{E}[\bm{H}(w, Z)|Z_1,\ldots,Z_{i-1}] \right\Vert \Big\} \\
    \leq & \mathop{\rm sup}_{z,\tilde{z}} \left\{ \left\Vert \mathbb{E}[\bm{H}(w, Z)|Z_1,\ldots,Z_{i-1},Z_i=z] - \mathbb{E}[\bm{H}(w, Z)|Z_1,\ldots,Z_{i-1},Z_i=\tilde{z}] \right\Vert \right\}.
\end{aligned}
\end{equation*}
After $Z_1,\ldots,Z_m$ are given, the values of $Z_{m+1},\ldots,Z_{n}$ do not affect the value of $\bm{H}(w, Z)$. Thus, $\xi^{\rm sup}_i - \xi^{\rm inf}_i = 0$ holds for $i = m+1, \ldots, n$. Now we discuss the case that $i\in [m]$. Similar to the proof in Appendix~\ref{pf1}, we consider a realization $z_{j}$ of $Z_{j}$ with $j\in [i-1]$. Denote by $z_i$ and $\widetilde{z}_i$ the realizations of $Z_i$ and $\widetilde{Z}_i$, respectively. Let $Z_{i+1:n}=(z_1,\ldots,z_{i-1},z_i,Z_{i+1},\ldots,Z_n)$ be the sequence where $Z_{i+1},\ldots,Z_n$ are obtained by sampling without replacement from $[n]\backslash \{z_1,\ldots,z_{i-1},z_i\}$, and $\widetilde{Z}_{i+1:n} = (z_1,\ldots,z_{i-1},\tilde{z}_i,\widetilde{Z}_{i+1},\ldots,\widetilde{Z}_n)$ be the sequence where $\widetilde{Z}_{i+1},\ldots,\widetilde{Z}_n$ are obtained by sampling without replacement from $[n]\backslash \{z_1,\ldots,z_{i-1},\tilde{z}_i\}$. Now we need to compute the maximum value of the expectation $\mathbb{E}[\bm{H}(w,Z_{i+1:n}) - \bm{H}(w,\widetilde{Z}_{i+1:n})]$ over any possible values $z_1,\ldots,z_{i-1},z_i,\widetilde{z}_i$. To this end, it is sufficient to consider two cases: (1) $z_i$ appears at positions $i+1$ to $m$ of $\widetilde{Z}_{i+1:n}$, and (2) $z_i$ appears at positions $m+1$ to $n$ of $\widetilde{Z}_{i+1:n}$. For case~(1), the expectation is equal to zero, since for each realization of $Z_{i+1:n}$, we can always find a corresponding and unique realization of $\widetilde{Z}_{i+1:n}$ such that they are equal to each other. For case~(2), for each realization of $Z_{i+1:n}$, we can always find a corresponding and unique realization of $\widetilde{Z}_{i+1:n}$ such that $Z$ and $\widetilde{Z}$ only differs at the $i$-th position. Thus, the maximum value of the expectation is given by
\begin{equation}
    \frac{m+u}{mu}\left\Vert \frac{\partial^2 \ell(w,s_{\widetilde{z}_i})}{\partial w^2} - \frac{\partial^2 \ell(w,s_{z_i})}{\partial w^2} \right\Vert \leq \frac{2(m+u)}{mu}\mathop{\rm sup}_{z} \left\Vert \frac{\partial^2 \ell(w,z)}{\partial w^2} \right\Vert = \frac{2(m+u)B_H}{mu}.
\end{equation}
The last step is to compute the probability of $z_i$ appearing at positions $m+1$ to $n$ of $\widetilde{Z}_{i+1:n}$. To this end, we need to sample $m-i$ elements among the rest $n-i-1$ elements (that is, the set $[n] \backslash \{z_1,\ldots,z_{i},\widetilde{z}_i\}$), and then apply permutation on them. Thus, the probability is given by $\frac{u!(m-i)!C_{n-i-1}^{m-i}}{(n-i)!}$. Put all the above ingredients together, we obtain 
\begin{equation}
    \xi^{\rm sup}_i - \xi^{\rm inf}_i = \begin{cases}
         \frac{u!(m-i)!C_{n-i-1}^{m-i}}{(n-i)!} \cdot \frac{2(m+u)B_H}{mu} = \frac{2(m+u)B_H}{m(m+u-i)}, & i = 1,\ldots,m, \\
         0, & i = m+1, \ldots, n.
        \end{cases}
\end{equation}
Denote by
\begin{equation}
    \mathbf{A}_i \triangleq \begin{cases}
         \frac{2(m+u)B_H}{m(m+u-i)} \mathbf{I}_d, & i = 1,\ldots,m, \\
         \bm{O}, & i = m+1, \ldots, n.
        \end{cases}
\end{equation}
Since $\lambda_{\rm max} \left( \xi^2_i \right) = \left\Vert \xi_i \right\Vert^2 \leq (\xi^{\rm sup}_i - \xi^{\rm inf}_i)^2$, we conclude that for $i \in [n]$, $\mathbf{A}^2_i - \xi^2_i$ is a semi-positive definite matrix. Denote by $\varepsilon$ the standard Rademacher variable, that is, $\mathbb{P}(\varepsilon=1)=\mathbb{P}(\varepsilon=-1)=\frac{1}{2}$, which is independent to $Z$. For any $\theta \in \mathbb{R}$ we have
\begin{equation}
\begin{aligned}
   & \mathbb{E} \left[ {\rm Tr}\left( \exp \left\{ \theta \sum_{i=1}^n \xi_i \right\} \right) \right] = \mathbb{E} \left[ \mathbb{E} \left[ {\rm Tr}\left( \exp \left\{ \theta \sum_{i=1}^{n-1} \xi_i + \theta \xi_n \right\} \right) \Bigg| Z_1, \ldots, Z_{n-1} \right] \right] \\
   \leq & \mathbb{E} \left[ \mathbb{E} \left[ {\rm Tr}\left( \exp \left\{ \theta \sum_{i=1}^{n-1} \xi_i + 2\varepsilon \theta \xi_n \right\} \right) \Bigg| Z_1, \ldots, Z_{n-1} \right] \right] \\
   \leq & \mathbb{E} \left[ {\rm Tr}\left( \exp \left\{ \theta \sum_{i=1}^{n-1} \xi_i + \log \mathbb{E} \left[ \exp \left\{ 2\varepsilon \theta \xi_n \right\} | Z_1, \ldots, Z_{n-1}\right] \right\} \right) \right] \\
   \leq & \mathbb{E} \left[ {\rm Tr}\left( \exp \left\{ \theta \sum_{i=1}^{n-1} \xi_i + 2\theta^2 \mathbf{A}^2_n \right\} \right) \right],
\end{aligned}
\end{equation}
where the first line is due to the tower property of conditional expectation. The second, the third, and the forth line is obtained by Lemma~7.6, Corollary~3.3 and Lemma~7.7 of \cite{Tropp2012}, respectively. By iteration we obtain
\begin{equation*}
\begin{aligned}
    & \mathbb{E} \left[ {\rm Tr}\left( \exp \left\{ \theta \sum_{i=1}^n \xi_i \right\} \right) \right] \leq {\rm Tr}\left( \exp \left\{ 2\theta^2 \sum_{i=1}^{n} \mathbf{A}^2_i \right\} \right) \\
    = & {\rm Tr}\left( \exp \left\{ \frac{8\theta^2B_H^2(m+u)^2}{m^2} \left(\sum_{i=1}^{m} \frac{1}{(m+u-i)^2} \right) \mathbf{I}_d \right\} \right) \\
    = & {\rm Tr}\left( \sum_{k=0}^\infty \frac{1}{k!} \cdot \left[ \frac{8\theta^2B_H^2(m+u)^2}{m^2} \left(\sum_{i=1}^{m} \frac{1}{(m+u-i)^2} \right) \mathbf{I}_d \right]^k \right) \\
    = & \sum_{k=0}^\infty \frac{d}{k!} \cdot \left[ \frac{8\theta^2B_H^2(m+u)^2}{m^2} \left(\sum_{i=1}^{m} \frac{1}{(m+u-i)^2} \right) \right]^k \\
    = & d \exp \left\{ \frac{8\theta^2B_H^2(m+u)^2}{m^2} \left(\sum_{i=1}^{m} \frac{1}{(m+u-i)^2} \right) \right\} \leq d \exp \left\{ \frac{8d^2\theta^2B_H^2(m+u)^2}{m(u-1/2)(m+u-1/2)} \right\}.
\end{aligned}
\end{equation*}
Due to the symmetry of $m$ and $u$, we have
\begin{equation}
    \mathbb{E} \left[ {\rm Tr}\left( \exp \left\{ \theta \sum_{i=1}^n \xi_i \right\} \right) \right] \leq d \exp \left\{ \frac{8\theta^2B_H^2(m+u)^2}{u(m-1/2)(m+u-1/2)} \right\}.
\end{equation}
Then the final bound is obtained by taking the smaller one of these two bounds,
\begin{equation}
\begin{aligned}
    \mathbb{E} \left[ {\rm Tr}\left( \exp \left\{ \theta \sum_{i=1}^n \xi_i \right\} \right) \right] & \leq d \exp \left\{ \frac{8\theta^2B_H^2(m+u)^2}{mu(m+u-1/2)} \cdot \frac{2\max(m,u)}{2\max(m,u)-1} \right\} \\
    & = d \exp \left\{ \frac{8\theta^2B_H^2C_{m,u}(m+u)}{mu} \right\}.
\end{aligned}
\end{equation}
By noticing that ${\rm Tr}( \bm{H}(w,z) ) \leq d \lambda_{\rm max}\left( \bm{H}(w,z) \right)$ that holds for any $w$ and $z$, for any $\theta > 0$ we have
\begin{equation}\label{part2_1}
\begin{aligned}
    & \mathbb{E}_{Z} \left[ e^{ \theta{\rm Tr}(\bm{H}(w,Z))} \right] \leq \mathbb{E}_{Z} \left[ e^{ d\theta \lambda_{\rm max}(\bm{H}(w,Z))} \right] \\
    = & \mathbb{E}_{Z} \left[ \lambda_{\rm max}( e^{ d \theta \bm{H}(w,Z) }) \right] \leq \mathbb{E}_{Z} \left[ {\rm Tr} \left( e^{ d \theta \bm{H}(w,Z) } \right) \right] \\
    = & \mathbb{E} \left[ {\rm Tr}\left( \exp \left\{ d \theta \sum_{i=1}^n \xi_i \right\} \right) \right] \leq d \exp \left\{ \frac{8\theta^2B_H^2C_{m,u}(m+u)}{mu} \right\}.
\end{aligned}
\end{equation}
By Lemma~\ref{lemma}, for any $\theta > 0$:
\begin{equation}\label{part2_2}
\begin{aligned}
    & \mathbb{E}_{Z} \mathbb{E}_{W\sim P} \left[ e^{\theta{\rm Tr}(\bm{H}(W,Z))} \right] = \mathbb{E}_{Z} \left[ e^{\mathop{\rm sup}_{Q \in \mathcal{P}(\mathcal{W})} \theta\mathbb{E}_{W \sim Q} \left[{\rm Tr}(\bm{H}(W,Z)) \right] - {\rm D}_{\rm KL}(Q || P)} \right] \\
    \geq & e^{\mathbb{E}_{Z} \left[ \mathop{\rm sup}_{Q \in \mathcal{P}(\mathcal{W})} \theta\mathbb{E}_{W \sim Q} \left[{\rm Tr}(\bm{H}(W,Z)) \right] - {\rm D}_{\rm KL}(Q || P) \right] } \\
    \geq & e^{ \mathop{\rm sup}_{Q \in \mathcal{P}(\mathcal{W})} \mathbb{E}_{Z} \left[\mathbb{E}_{W \sim Q} \left[\theta {\rm Tr}(\bm{H}(W,Z)) \right] - {\rm D}_{\rm KL}(Q || P) \right] }.
\end{aligned}
\end{equation}
Combining Eq.~(\ref{part2_1}) with Eq.~(\ref{part2_2}), for any $\theta > 0$ we have
\begin{equation*}
\begin{aligned}
    & \mathop{\rm sup}_{Q \in \mathcal{P}(\mathcal{W})} \mathbb{E}_{Z} \left[\mathbb{E}_{W \sim Q} \left[\theta {\rm Tr}(\bm{H}(W,Z)) \right] - {\rm D}_{\rm KL}(Q || P) \right] \leq \frac{8d^2 \theta^2 B^2_H C_{m,u}(m+u)}{mu} + \log(d).
\end{aligned}
\end{equation*}
Now we put $Q=P_{W_T|Z}$ and $P=\mathbb{E}[P_{W_T|Z}]=P_{W_T}$ and get
\begin{equation}
\begin{aligned}
    \theta \mathbb{E}_{W_T,Z} \left[{\rm Tr}(\bm{H}(W,Z)) \right] \leq I(W_T;Z) + \frac{8d^2\theta^2 B^2_H C_{m,u} (m+u)}{mu} + \log(d),
\end{aligned}
\end{equation}
which implies that
\begin{equation}\label{part2_3}
    \left\vert \mathbb{E}_{W_T,Z} \left[{\rm Tr}(\bm{H}(W_T,Z)) \right] \right\vert \leq \sqrt{\frac{32d^2 B^2_H C_{m,u} (I(W_T;Z)+\log (d)) (m+u)}{mu}}.
\end{equation}
By Taylor’s expansion on $R(w,z)$ w.r.t. $w$ and using $N_{1:T} \sim \mathcal{N}(0,\sum_{t=1}^T \sigma^2_t) \in \mathbb{R}^d$, we have
\begin{equation}
\begin{aligned}
    & R_{\rm train}({w}_T + N_{1:T}, z) - R_{\rm train}(w_T, z) \\
    = & \sum_{i=1}^{d} \left(N_{1:T}\right)_i \Bigg(\frac{\partial R_{\rm train}(W,Z)}{\partial w} \bigg|_{w=w_T}\Bigg)_i + \sum_{i,j=1}^{d} \left(N_{1:T}\right)_i \left(N_{1:T}\right)_j \Bigg(\frac{\partial^2 R_{\rm train}(W,Z)}{\partial w^2} \bigg|_{w=w_T}\Bigg)_{ij} \\
    & + \sum_{i,j,k=1}^{d} \left(N_{1:T}\right)_i \left(N_{1:T}\right)_j \left(N_{1:T}\right)_k \Bigg(\frac{\partial^3 R_{\rm train}(W,Z)}{\partial w^3} \bigg|_{w=w_T}\Bigg)_{ijk} + \mathcal{O}\Bigg(\bigg(\sum_{t=1}^T \sigma^2_t\bigg)^2\Bigg).
\end{aligned}
\end{equation}
Taking expectation on both side over $N_{1:T}$, we have
\begin{equation}
\begin{aligned}
    & \mathbb{E}_{N_{1:T}} \left[ R_{\rm train}({w}_T + N_{1:T}, z) - R_{\rm train}(w_T, z) \right] \\
    = & \sum_{i=1}^{d} \mathbb{E} \left[ \left(N^2_{1:T} \right)_i \right] \Bigg(\frac{\partial^2 R_{\rm train}(W,Z)}{\partial w^2} \bigg|_{w=w_T} \Bigg)_{ii} \\
    = & {\rm Tr}\left(\mathbb{E}[N_{1:T}]\mathbb{E}[N_{1:T}^\top] \bm{H}_{\rm train}(w_T,z) \right) + \mathcal{O}\Bigg(\bigg(\sum_{t=1}^T \sigma^2_t\bigg)^2\Bigg),
\end{aligned}
\end{equation}
which implies that
\begin{equation}
\begin{aligned}
    & \mathbb{E}_{Z,W_T,N_{1:T}} \left[ R_{\rm train}({W}_T+N_{1:T}, Z) - R_{\rm train}(W_T, Z) \right] \\
    = & {\rm Tr}\left( \mathbb{E}[N_{1:T}]\mathbb{E}[N_{1:T}^\top] \left(\mathbb{E}_{Z,W_T}\left[\bm{H}_{\rm train}(W_T,Z)\right] \right) \right) + \mathcal{O}\Bigg(\bigg(\sum_{t=1}^T \sigma^2_t\bigg)^2\Bigg).
\end{aligned}
\end{equation}
Similarly we have
\begin{equation}
\begin{aligned}
    & \mathbb{E}_{Z,W_T,N_{1:T}} \left[ R_{\rm test}({W}_T+N_{1:T}, Z) - R_{\rm test}(W_T, Z) \right] \\
    = & {\rm Tr}\left( \mathbb{E}[N_{1:T}]\mathbb{E}[N_{1:T}^\top] \left(\mathbb{E}_{Z,W_T}\left[\bm{H}_{\rm test}(W_T,Z)\right] \right) \right) + \mathcal{O}\Bigg(\bigg(\sum_{t=1}^T \sigma^2_t\bigg)^2\Bigg).
\end{aligned}
\end{equation}
Therefore,
\begin{small}
\begin{equation*}
\begin{aligned}
    & \mathbb{E}_{Z,W_T,N_{1:T}} \left[\left[ R_{\rm train}(W_T+N_{1:T}, Z) - R_{\rm train}(W_T, Z) \right] - \left[ R_{\rm test}(W_T+N_{1:T}, Z) - R_{\rm test}(W_T, Z) \right] \right] \\
    = & {\rm Tr}\left( \mathbb{E}[N_{1:T}]\mathbb{E}[N_{1:T}^\top]\left(\mathbb{E}_{Z,W_T}[\bm{H}_{\rm train}(W_T,Z) - \bm{H}_{\rm test}(W_T,Z)] \right) \right) + \mathcal{O}\Bigg(\bigg(\sum_{t=1}^T \sigma^2_t\bigg)^2\Bigg) \\
    = & \left(\sum_{t=1}^T \sigma^2_t \right) {\rm Tr} \left( \mathbb{E}_{Z,W_T}[\bm{H}_{\rm train}(W_T,Z) - \bm{H}_{\rm test}(W_T,Z)] \right) + \mathcal{O}\Bigg(\bigg(\sum_{t=1}^T \sigma^2_t\bigg)^2\Bigg) \\
    \leq & \left(\sum_{t=1}^T \sigma^2_t \right) \left\vert \mathbb{E}_{W_T,Z} \left[{\rm Tr}(\bm{H}(W,Z)) \right] \right\vert + \mathcal{O}\Bigg(\bigg(\sum_{t=1}^T \sigma^2_t\bigg)^2\Bigg) \\
    \leq & \left(\sum_{t=1}^T \sigma^2_t \right) \sqrt{\frac{32 d^2 B^2_H C_{m,u} (I(W_T;Z)+\log (d)) (m+u)}{mu}} + \mathcal{O}\Bigg(\bigg(\sum_{t=1}^T \sigma^2_t\bigg)^2\Bigg).
\end{aligned}
\end{equation*}
\end{small}
This finishes the proof.

\section{Proof of Theorem~\ref{thm5}}\label{proof5}
\noindent
{\bf Proof}. This proof is inspired by the work of \cite{Neu2021,Wang2022}.
We use $\widetilde{W}_T $ as the abbreviation of $W_T+N_{1:T}$.
Following the work of \cite{Wang2022}, the mutual information term is decomposed by
\begin{equation*}
\begin{aligned}
    I(Z;\widetilde{W}_T) = & I \left( Z;\widetilde{W}_{T-1} - \frac{\eta}{\sqrt{V_T}+\epsilon} \odot g(W_{T-1}, B_T(Z)) + N_T \right) \\
    \leq & I \left( Z;\widetilde{W}_{T-1}, - \frac{\eta}{\sqrt{V_T}+\epsilon} \odot g(W_{T-1}, B_T(Z)) + N_T \right) \\
    = & I(Z;\widetilde{W}_{T-1}) + I\left(- \frac{\eta}{\sqrt{V_T}+\epsilon} \odot g(W_{T-1}, B_T(Z)) + N_T;Z \bigg|\widetilde{W}_{T-1}\right).
\end{aligned}
\end{equation*}
By iteration we obtain
\begin{equation}\label{information}
\begin{aligned}
   I(Z;\widetilde{W}_T) \leq & \sum_{t=1}^T I\left(-\frac{\eta}{\sqrt{V_t}+\epsilon} \odot g(W_{t-1}, B_t(Z)) + N_t;Z \bigg|\widetilde{W}_{t-1}\right) \\
   = & \sum_{t=2}^T I\left(-\frac{\eta}{\sqrt{V_t}+\epsilon} \odot g(\widetilde{W}_{t-1} - N_{1:t-1}, B_t(Z)) + N_t;Z \bigg|\widetilde{W}_{t-1}\right) \\
   & + I\left(-\frac{\eta}{\sqrt{V_1}+\epsilon} \odot g(\widetilde{W}_{0}, B_t(Z)) + N_1;Z \bigg|\widetilde{W}_{0}\right).
\end{aligned}
\end{equation}
Then we need to provide an upper bound for the conditional mutual information. Let $V,X,U$ be random variables that are independent of $N \sim \mathcal{N}(\bm{0}, \mathbf{I}_d)$ and $\Psi$ be a function of random variables $U,V,X,Y$. Denote by $h(\cdot)$ the differential entropy, then
\begin{equation}\label{entropy}
\begin{aligned}
    & I(\Psi(V,y-U,X)+\sigma N;X|Y=y) \\
    = & h(\Psi(V,y-U,X)+\sigma N|Y=y) - h(\Psi(V,y-U,X)+\sigma N|X,Y=y).
\end{aligned}
\end{equation}
For the first term in Eq.~(\ref{entropy}), by Theorem~2.7 of \cite{Polyanskiy2022} we have
\begin{equation}\label{entropy1}
\begin{aligned}
    & h(\Psi(V,y-U,X)+\sigma N|Y=y) \\
    \leq & \frac{d}{2} \log \left( \frac{2\pi e \mathbb{E} \left[ \Vert \Psi(V,y-U,X) + \sigma N \Vert_2^2 | Y=y \right]}{d} \right) \\
    = & \frac{d}{2} \log \left( \frac{2\pi e \left( \mathbb{E} \left[ \Vert \Psi(V,y-U,X) \Vert_2^2 | Y=y \right] + \sigma^2 \mathbb{E} \left[ \Vert N \Vert_2^2 \right] \right)}{d} \right) \\
    = & \frac{d}{2} \log \left( \frac{ 2\pi e \left( \mathbb{E} \left[ \Vert \Psi(V,y-U,X) \Vert^2_2 | Y=y \right] + d\sigma^2 \right) }{d} \right).
\end{aligned}
\end{equation}
For the second term in Eq.~(\ref{entropy}), we have
\begin{equation}\label{entropy2}
\begin{aligned}
    & h(\Psi(V,y-U,X) + \sigma N|X,Y=y) \geq h(\Psi(V,y-U,X)+\sigma N|U,V,X,Y=y)  \\
    = & h(\sigma N) = \frac{d}{2}\log \left(2\pi e \sigma^2 \right).
\end{aligned}
\end{equation}
For $t = 2,\ldots, T$, let $\sigma = \sigma_t$, $V=W^{[t-2]} \triangleq (W_0, \ldots, W_{t-2})$, $X=Z$, $Y=\widetilde{W}_{t-1}$, $U = N_{1:t-1}$ and 
\begin{equation*}
\begin{aligned}
     & \Psi(V,y-U,X) = \Psi(W^{[t-2]},\widetilde{w}_{t-1}-N_{1:t-1},Z) \\
     = & -\frac{\eta}{\sqrt{V_t(W^{[t-2]},\widetilde{w}_{t-1}-N_{1:t-1})}+\epsilon} \odot g(\widetilde{w}_{t-1}-N_{1:t-1}, B_t(Z)).
\end{aligned}
\end{equation*}
Plugging Eqs.~(\ref{entropy1},\ref{entropy2}) into Eq.~(\ref{entropy}), for $t=2,\ldots,T$ we have
\begin{equation*}
\begin{aligned}
   & I\left(\Psi(W^{[t-2]},\widetilde{w}_{t-1}-N_{1:t-1},Z)+N_t;Z \big|\widetilde{W}_{t-1}=\widetilde{w}_{t-1} \right) \\
   \leq & \frac{d}{2}\log \left( \frac{1}{d\sigma_t^2} \mathbb{E} \left[ \big\Vert \Psi(W^{[t-2]},\widetilde{w}_{t-1}-N_{1:t-1},Z) \big\Vert^2_2 \Big|\widetilde{W}_{t-1}=\widetilde{w}_{t-1} \right] + 1 \right),
\end{aligned}
\end{equation*}
which implies that
\begin{equation}\label{term}
\begin{aligned}
    & I\left(\Psi(W^{[t-2]},\widetilde{w}_{t-1}-N_{1:t-1},Z)+N_t;Z \big|\widetilde{W}_{t-1}\right) \\
    \leq & \int_{\widetilde{w}_{t-1}} \frac{d}{2}\log \left( \frac{1}{d\sigma_t^2} \mathbb{E} \left[ \big\Vert \Psi(W^{[t-2]},\widetilde{w}_{t-1}-N_{1:t-1},Z) \big\Vert^2_2 \Big|\widetilde{W}_{t-1}=\widetilde{w}_{t-1} \right] + 1 \right) \dif P_{\widetilde{W}_{t-1}}(\widetilde{w}_{t-1}) \\
    \leq & \frac{d}{2}\log \left( \frac{1}{d\sigma_t^2} \mathbb{E} \left[ \Big\Vert \Psi(W^{[t-2]},\widetilde{W}_{t-1}-N_{1:t-1},Z) \Big\Vert^2_2 \right] + 1 \right) \\
    = & \frac{d}{2}\log \left( \frac{1}{d\sigma_t^2} \mathbb{E} \left[ \Big\Vert \Psi(W^{[t-1]},Z) \Big\Vert^2_2 \right] + 1 \right).
\end{aligned}
\end{equation}
By the same way, we have 
\begin{equation}\label{term2}
\resizebox{0.95\hsize}{!}{$
\begin{aligned}
    & I\left(-\frac{\eta}{\sqrt{V_1}+\epsilon} \odot g(\widetilde{W}_{0}, B_t(Z)) + N_1;Z \bigg|\widetilde{W}_{0}\right) \leq \frac{d}{2}\log \left( \frac{1}{d\sigma_1^2} \mathbb{E} \left[ \big\Vert \Psi(W_0,Z) \big\Vert^2_2 \Big|\widetilde{W}_{0}=\widetilde{w}_{0} \right] + 1 \right).
\end{aligned}$}
\end{equation}
Plugging Eqs.~(\ref{term},\ref{term2}) into Eq.~(\ref{information}), we have
\begin{equation}\label{information2}
\begin{aligned}
    I(Z;\widetilde{W}_T) \leq & \sum_{t=1}^T \frac{d}{2}\log \left( \frac{1}{d\sigma_t^2} \mathbb{E} \left[ \Big\Vert \Psi(W^{[t-2]},{W}_{t-1},Z) \Big\Vert^2_2 \right] + 1 \right) \\
    = & \sum_{t=1}^T \frac{d}{2}\log \left( \frac{1}{d\sigma_t^2} \mathbb{E} \left[ \bigg\Vert \frac{\eta}{\sqrt{V_T(W^{[t-1]})}+\epsilon} \odot g(W_{t-1}, B_t(Z)) \bigg\Vert^2_2 \right] + 1 \right) .
\end{aligned}
\end{equation}
Now we discuss how to extend this result to Adam optimization algorithm. For $t\in [T]$, the update rule of Adam is
\begin{equation*}
\begin{aligned}
    & M_t = \beta_1 M_{t-1} + (1-\beta_1) g(W_{t-1}, B_t(Z)), \\
    & V_t = \beta_2 V_{t-1} + (1-\beta_2) g(W_{t-1}, Z) \odot g(W_{t-1}, B_t(Z)), \\
    & \hat{V}_t = \frac{V_t}{1-\beta^t_2}, \hat{M}_t = \frac{M_t}{1-\beta^t_1}, W_t = W_{t-1} - \frac{\eta }{\sqrt{\hat{V}_t}+\epsilon} \odot \hat{M}_t,
\end{aligned}
\end{equation*}
where $\beta_1$, $\beta_2$, $\eta$ and $\epsilon$ are hyperparameters. Define
\begin{equation*}
    \Psi(W^{[t-1]},Z) \triangleq - \sum_{\tau = 0}^{t-1} \frac{\eta(1-\beta_1)\beta^{t-\tau-1}_1}{\sqrt{\hat{V}_t}+\epsilon} \odot g(W_{\tau},B_{\tau+1}(Z)),
\end{equation*}
we have $W_t = W_{t-1} + \Psi(W^{[t-1]},Z)$. Similarly, we construct the weight process as
\begin{equation*}
    \widetilde{W}_0 = W_0, \widetilde{W}_{t} = \widetilde{W}_{t-1} + \Psi(W^{[t-1]},Z) + N_t, t\in [T].
\end{equation*}
Following the above technique, one can find that
\begin{equation*}
\begin{aligned}
    I(Z;\widetilde{W}_T) \leq & \sum_{t=1}^T \frac{d}{2}\log \left( \frac{1}{d\sigma_t^2} \mathbb{E} \left[ \Big\Vert \Psi(W^{[t-2]},{W}_{t-1},Z) \Big\Vert^2_2 \right] + 1 \right) \\
    = & \sum_{t=1}^T \frac{d}{2}\log \left( \frac{1}{d\sigma_t^2} \mathbb{E} \left[ \bigg\Vert \sum_{\tau = 0}^{t-1}  \frac{\eta(1-\beta_1)\beta^{t-\tau-1}_1}{\sqrt{\hat{V}_t}+\epsilon} \odot g(W_{\tau},B_{\tau+1}(Z)) \bigg\Vert^2_2 \right] + 1 \right).
\end{aligned}
\end{equation*}
This completes the proof.

\section{Proof of Proposition~\ref{pro2}}\label{ssl_pro}
\noindent
{\bf Proof}. Denote by $g(F_i,U_i,\widetilde{Z}, S) \triangleq r(F_{i,U_i}, Y_{\widetilde{Z}_{i,U_i}}) - r(F_{i,1-U_i}, Y_{\widetilde{Z}_{i,1-U_i}})$ the function of $(F_i,U_i,\widetilde{Z},S)$. Recall that $F_{i,U_i} = f_W(X_{\widetilde{Z}_{i,U_i}})$ and $F_{i,1-U_i} = f_W(X_{\widetilde{Z}_{i,1-U_i}})$. Let $f_i$, $\widetilde{z}$ and $s$ be the fixed realizations of $F_i$, $\widetilde{Z}$ and $S$, respectively. For any $\lambda \in \mathbb{R}$, by Hoeffding's inequality \citep{Hoeffding1963},
\begin{equation}
    \mathbb{E}_{U_i} \left[ \exp \left\{ \lambda g(f_i, U_i, \widetilde{z}, s) \right\} \right] \leq \exp \left\{ \frac{\lambda^2 B^2}{2} \right\}, i\in [m].
\end{equation}
Let $U'_i$ be the independent copy of $U_i$, by Lemma~\ref{lemma},
\begin{equation}
\begin{aligned}
   & I(F_i;U_i|\widetilde{Z}=\widetilde{z},S=s) \\
   \geq & \lambda \mathbb{E}_{F_i,U_i|\widetilde{Z}=\widetilde{z},S=s} \left[ g(F_i, U_i, \widetilde{z}, s) \right] - \log \mathbb{E}_{F_i,U'_i|\widetilde{Z}=\widetilde{z},S=s} \left[ \exp \left\{ \lambda g(F_i, U'_i,\widetilde{z}, s) \right\} \right] \\
   \geq & \lambda \mathbb{E}_{F_i,U_i|\widetilde{Z}=\widetilde{z},S=s} \left[ g(F_i, U_i,\widetilde{z}) \right] - \frac{\lambda^2B^2}{2},
\end{aligned}
\end{equation}
which implies that
\begin{equation}
    \left\vert \mathbb{E}_{F_i,U_i|\widetilde{Z}=\widetilde{z},S=s} \left[ g(F_i, U_i,\widetilde{z},s) \right] \right\vert \leq B\sqrt{2I^{\widetilde{z},s}(F_i;U_i)}.
\end{equation}
Taking expectation over the joint distribution of $\widetilde{Z}$ and $S$ on both side yields
\begin{equation}
    \mathbb{E}_{\widetilde{Z},S}\left\vert \mathbb{E}_{F_i,U_i|\widetilde{Z},S} \left[ g(F_i, U_i,\widetilde{Z}) \right] \right\vert \leq B\mathbb{E}_{\widetilde{Z},S}\sqrt{2I^{\widetilde{Z},S}(F_i;U_i)}.
\end{equation}
Then we have
\begin{equation}\label{equ1}
\begin{aligned}
    & \left\vert \mathbb{E}_{W,S} \left[ R_{\text{test}}(W,S) - R_{\text{train}}(W,S) \right] \right\vert \\
    = & \left\vert \mathbb{E}_{W,S,\widetilde{Z},U} \left[ \frac{1}{m}\sum_{i=m+1}^{2m} \ell \left( W,S_{\mathscr{Z}_i(\widetilde{Z},U)} \right) - \frac{1}{m}\sum_{i=1}^m \ell \left( W,S_{\mathscr{Z}_i(\widetilde{Z},U)} \right) \right] \right\vert \\
    = & \left\vert \frac{1}{m}\sum_{i=1}^{m} \mathbb{E}_{\widetilde{Z},S}\mathbb{E}_{W,U_i|\widetilde{Z},S} \left[  \ell \left( W,S_{\widetilde{Z}_{i,1-U_i}} \right) - \ell \left( W,S_{\widetilde{Z}_{i,U_i}} \right) \right] \right\vert \\
    \leq & \frac{1}{m}\sum_{i=1}^{m} \mathbb{E}_{\widetilde{Z},S} \left\vert \mathbb{E}_{W,U_i|\widetilde{Z},S} \left[  \ell \left( W,S_{\widetilde{Z}_{i,1-U_i}} \right) - \ell \left( W,S_{\widetilde{Z}_{i,U_i}} \right) \right] \right\vert \\
    = & \frac{1}{m} \sum_{i=1}^m \mathbb{E}_{\widetilde{Z},S} \left\vert \mathbb{E}_{W,U_i|\widetilde{Z},S} \left[ r \big(f_W(X_{\widetilde{Z}_{i,1-U_i}}), Y_{\widetilde{Z}_{i,1-U_i}} \big) - r \big(f_W(X_{\widetilde{Z}_{i,U_i}}), Y_{\widetilde{Z}_{i,U_i}} \big) \right] \right\vert \\
    = & \frac{1}{m} \sum_{i=1}^m \mathbb{E}_{\widetilde{Z},S} \left\vert \mathbb{E}_{F_i,U_i|\widetilde{Z},S} \left[ r (F_{i,1-U_i}, Y_{\widetilde{Z}_{i,1-U_i}} ) - r (F_{i,U_i}, Y_{\widetilde{Z}_{i,U_i}} \big) \right] \right\vert \\
    = & \frac{1}{m} \sum_{i=1}^m \mathbb{E}_{\widetilde{Z},S} \left\vert \mathbb{E}_{F_i,U_i|\widetilde{Z},S} \left[ g(F_i, U_i,\widetilde{Z},S) \right] \right\vert \leq \frac{B}{m} \sum_{i=1}^m \mathbb{E}_{\widetilde{Z},S} \sqrt{2I^{\widetilde{Z},S}(F_i;U_i)}.
\end{aligned}
\end{equation}
Denote by $g(L_i, U_i) = L_{i,1-U_i} - L_{i,U_i}$ and $g(\Delta_i, U_i) \triangleq (-1)^{U_i}\Delta_i$, following the above procedure we have
\begin{align}
    & \left\vert \mathbb{E}_{W,S} \left[ R_{\text{test}}(W,S) - R_{\text{train}}(W,S) \right] \right\vert \leq \frac{B}{m} \sum_{i=1}^m \mathbb{E}_{\widetilde{Z},S} \sqrt{2I^{\widetilde{Z},S}(L_i;U_i)}, \label{eql1} \\
    & \left\vert \mathbb{E}_{W,S} \left[ R_{\text{test}}(W,S) - R_{\text{train}}(W,S) \right] \right\vert \leq \frac{B}{m} \sum_{i=1}^m \mathbb{E}_{\widetilde{Z},S} \sqrt{2I^{\widetilde{Z},S}(\Delta_i;U_i)}.
\end{align}
For the cases that $u=km,k\in \mathbb{N}_+$, define
$F_{i} \triangleq (f_W(X_{\widetilde{Z}_{i,0}}), \ldots, f_W(X_{\widetilde{Z}_{i,k}})), i\in [m]$ and $L_{i,:} \triangleq (\ell(W,S_{\widetilde{Z}_{i,0}}),\ldots, \ell(W,S_{\widetilde{Z}_{i,k}})), i\in [m]$ be the prediction of model and the sequence of loss values. Denote by $g(F_i,U_i,\widetilde{Z},S) \triangleq \frac{1}{k}\sum_{j=1}^k r \big( F_{i,U_{i,j}}, Y_{\widetilde{Z}_{i,U_{i,j}}} \big) - r \big( F_{i,U_{i,0}}, Y_{\widetilde{Z}_{i,U_{i,0}}} \big)$ and $g(F_i,U_i,\widetilde{Z},S) \triangleq \frac{1}{k}\sum_{j=1}^k L_{i,U_{i,j}}, - L_{i,U_{i,0}}$ the function of $(F_i,U_i,\widetilde{Z},S)$ and $(L_i,U_i,\widetilde{Z},S)$, respectively. Following the above process one can verify that Eqs.~(\ref{equ1},\ref{eql1}) still hold under the cases that $u=km,k\in \mathbb{N}_+$. This finishes the proof.

\section{Experimental Details}\label{exp_detail}

We first discuss how to estimate the expected transductive generalization gaps and the derived bounds. Notice that computing their accurate values is not applicable since we need to run the algorithm on $(m+u)!$ splits in total. Therefore we use Monte Carlo simulation to estimate these expectations based on finite samples. For semi-supervised learning, the sampling process is as follows: (\romannumeral1) randomly draw $t_1$ full data points $S$ by each time sampling $m+u$ images from the raw images set, (\romannumeral2) randomly draw $t_2$ transductive supersamples $\widetilde{Z}$ based on Definition~\ref{def2}, (\romannumeral3) randomly draw $t_3$ selector sequence $U$ and obtain the training and test samples set according to Section~\ref{cmi_sec}. 
Now we discuss the estimation of the transductive generalization gap and the upper bounds established in Proposition~\ref{pro2}. Take Eq.~(\ref{semi_bound1}) as an example, for each $(s,\widetilde{z})$ we use the mean value over $t_3$ samples of $U$ to estimate the conditional term expectation term $\frac{1}{m}\sum_{i=1}^m \mathbb{E}_{F_i,U_i|s,\widetilde{z}} \left[g(F_i,U_i,s,\widetilde{z})\right]$. After that, we use $t_1$ samples of $S$ and $t_2$ samples of $\widetilde{Z}$ to estimate the expected generalization gap, whose mean and standard deviation are shown in Figure~\ref{ssl}. Similarly, we use a plug-in estimator \citep{Paninski2003} to estimate the disentangled mutual information $I^{s,\widetilde{z}}(F_i;U_i)$ over the $t_3$ samples of $U$. Then the upper bounds in Proposition~\ref{pro2} are estimated by the $t_1$ samples of $S$ and $t_2$ samples of $\widetilde{Z}$, whose mean and standard deviation are shown in Figure~\ref{ssl}. For transductive graph learning, the estimation process generally follows that of semi-supervised learning. The only difference is that we do not need to consider the sampling of $S$. Concretely, the sampling process is only composed of (\romannumeral2) and (\romannumeral3). Accordingly, we use $t_2$ samples of $\widetilde{Z}$ to estimate the expected bounds and the conditional mutual information. The results are shown in Figure~\ref{graph1} and Figure~\ref{graph2}. 

\begin{table}
\centering
\resizebox{\textwidth}{!}{
\begin{tabular}{ccl}
  \toprule
  Layer Type & Parameter  \\
  \midrule
  Conv & $16$ filters, $3\times 3$ kernels, stride $1$, padding $1$, BatchNormalization, ReLU \\
  Conv & $32$ filters, $3\times 3$ kernels, stride $1$, padding $1$, BatchNormalization, ReLU\\
  Conv & $32$ filters, $3\times 3$ kernels, stride $1$, padding $1$ \\
  ConvShortcut & $32$ filters, $1\times 1$ kernels, stride $1$, BatchNormalization, ReLU \\
  Conv$\times 6$ & $32$ filters, $3\times 3$ kernels, stride $1$, padding $1$, BatchNormalization, ReLU \\
  Conv & $64$ filters, $3\times 3$ kernels, stride $1$, padding $1$, BatchNormalization, ReLU\\
  Conv & $64$ filters, $3\times 3$ kernels, stride $1$, padding $1$ \\
  ConvShortcut & $64$ filters, $1\times 1$ kernels, stride $1$, BatchNormalization, ReLU \\
  Conv$\times 6$ & $64$ filters, $3\times 3$ kernels, stride $1$, padding $1$, BatchNormalization, ReLU \\
  Conv & $128$ filters, $3\times 3$ kernels, stride $1$, padding $1$, BatchNormalization, ReLU\\
  Conv & $128$ filters, $3\times 3$ kernels, stride $1$, padding $1$ \\
  ConvShortcut & $128$ filters, $1\times 1$ kernels, stride $1$, BatchNormalization, ReLU \\
  Conv$\times 6$ & $128$ filters, $3\times 3$ kernels, stride $1$, padding $1$, BatchNormalization, ReLU \\
  FC & $10$ units, linear activation \\
  \bottomrule
\end{tabular}}
\caption{The architecture of the convolutional neural network used for CIFAR-10. $\times 6$ means repeating the layer for $6$ times.}
\label{tab1}
\end{table}

Now we detail the settings of network architecture and hyperparameters. For the semi-supervised learning task, the network architectures on MNIST and CIFAR-$10$ are presented in Table~1 of \cite{Harutyunyan2021} and Table~\ref{tab1}, respectively. On both these two datasets, we set $t_1=t_2=2$ and $t_3=50$. For the transductive graph learning task, the architecture of GAT and GPR-GNN follows the ones used by \cite{Chien2021adaptive}. We set $t_2=5$ and $t_3=50$ on all datasets. We adopt the code released by \cite{Chien2021adaptive} to generate the cSBMs datasets with $n \in \{500, 1000, 2000\}$ and $\phi \in \{-0.5, 0.5\}$. Recall that the number of training data points is defined by $m \triangleq \frac{n}{k+1}$. To ensure that $m \in \mathbb{N}_+$, we set $k=1$ for Cora and Actor, and $k=2$ for CiteSeer and Chameleon.

\vskip 0.2in
\bibliography{main}

\end{document}